\documentclass[11pt]{article}

\usepackage{appendix}
\usepackage{setspace}\spacing{1.3}
\usepackage{palatino}
\usepackage{tocloft}
\usepackage[margin=1in, headheight=14pt]{geometry} %
\usepackage{graphics}
\usepackage{graphicx}
\usepackage{gensymb} %
\usepackage{titlesec}
\usepackage[natbibapa]{apacite}
\usepackage{xcolor}
\usepackage{array}%
\usepackage{bibentry} %
\usepackage{subfig} %
\usepackage{booktabs}%
\usepackage{tikz} %
\usepackage{ctable}%
\usepackage[countmax]{subfloat} %
\usepackage{enumitem} %
\usepackage{multirow} %
\usepackage{wrapfig} %
\usepackage{float}  %
\usepackage{hyperref} %
\usepackage{textcomp}
\usepackage{amsmath}
\usepackage{amssymb}
\usepackage{wasysym}

\usepackage[font=footnotesize]{caption} %
\usepackage{pifont} %
\usepackage{enumitem}
\usepackage{url}
\usepackage{breakurl} %
\usepackage[affil-it]{authblk}
\usepackage{fancyhdr}
\usepackage{placeins}
\usepackage{ifthen}

\graphicspath{
{figures/}
{../figures/}
{../../figures/}
{../../../figures/}
}

\hypersetup{
    colorlinks,%
    citecolor=black,%
    filecolor=black,%
    linkcolor=black,%
    urlcolor=black
    }

\setlist{leftmargin=*} 
\setlist[1]{labelindent=\parindent} %
\setlist{topsep = 0cm,partopsep = 1pt, parsep = 1pt} %

\titleformat{\section}
{\bfseries \Large}{\thesection}{0.5em}{}

\titleformat{\subsection}
{\bfseries \large}{\thesubsection}{0.5em}{}

\titleformat{\subsubsection}
{\bfseries}{\thesubsubsection}{0.5em}{}

\titlespacing{\section}{0cm}{0.3cm}{0.1cm}
\titlespacing{\subsection}{0cm}{0.3cm}{0.1cm}
\titlespacing{\subsubsection}{0cm}{0.3cm}{0.1cm}

\newcolumntype{M}{>{\centering\arraybackslash}m{\dimexpr.2\linewidth-2\tabcolsep}}

\newcommand{\specialcell}[2][c]{%
  \begin{tabular}[#1]{@{}c@{}}#2\end{tabular}}

\DeclareMathOperator*{\E}{\mathbb{E}}
\DeclareMathOperator*{\Do}{Do}
\DeclareMathOperator*{\pa}{pa}

\newcommand{\ww}{\mathbf{w}}
\newcommand{\eee}{\mathbf{c}}
\newcommand{\dd}{\mathbf{d}}
\newcommand{\cald}{\mathcal{D}}
\newcommand{\calc}{\mathcal{C}}
\newcommand{\call}{\mathcal{L}}
\renewcommand{\lll}{\mathbf{l}}

\begin{document}

\begin{center} 
{\LARGE \textbf{Formalizing Neurath's Ship: Approximate Algorithms for Online Causal Learning}}
\linebreak
\linebreak
{\large Neil R. Bramley\\
Peter Dayan\\
Thomas L. Griffiths\\
David A. Lagnado}

%
\end{center}

\vspace{2cm}
\noindent Cite as:\\
\hangindent=0.7cm Bramley, N. R., Dayan, P., Griffiths, T. L. \& Lagnado, D. A. (2017).  Formalizing Neurath's ship: Approximate algorithms for online causal learning. \emph{Psychological Review}, Vol 124 (3), 301-338. \url{http://dx.doi.org/10.1037/rev0000061}
\vspace{2cm}

\noindent
Author note:\\
Neil R. Bramley, Department of Experimental Psychology, University College London; Peter Dayan, Gatsby Computational Neuroscience Unit, University College London; Thomas L. Griffiths, Department of Psychology, University of California Berkeley; David A. Lagnado, Department of Experimental Psychology, University College London.

This research was supported in part by an Economic and Social Research Council UK grant (RES 062330004).

Correspondence concerning this article should be addressed to: 201, 26 Bedford Way, University College London, London, UK,WC1H 0DS, Email: \href{mailto:neil.bramley@ucl.ac.uk}{neil.bramley@ucl.ac.uk} .

Experiment 2 previously appeared in a conference paper presented at The 37\textsuperscript{th} Annual Conference of The Cognitive Science Society \citep{bramley2015staying}.  A preprint of the current manuscript was made available on arXiv on September 17th 2016 here: \url{https://arxiv.org/abs/1609.04212}.

\newpage

\abstract{\noindent Higher-level cognition depends on the ability to learn models of the world. We can characterize this at the computational level as a structure-learning problem with the goal of best identifying the prevailing causal relationships among a set of relata.  However, the computational cost of performing exact Bayesian inference over causal models grows rapidly as the number of relata increases.  This implies that the cognitive processes underlying causal learning must be substantially approximate.  A powerful class of approximations that focuses on the sequential absorption of successive inputs is captured by the \emph{Neurath's ship} metaphor in philosophy of science, where theory change is cast as a stochastic and gradual process shaped as much by people's limited willingness to abandon their current theory when considering alternatives as by the ground truth they hope to approach. Inspired by this metaphor and by algorithms for approximating Bayesian inference in machine learning, we propose an algorithmic-level model of causal structure learning under which learners represent only a single global hypothesis that they update locally as they gather evidence.  We propose a related scheme for understanding how, under these limitations, learners choose informative interventions that manipulate the causal system to help elucidate its workings.  We find support for our approach in the analysis of three experiments.

\newpage

By adulthood, a normal person will have developed a sophisticated and
structured understanding of the world.  The ``blooming buzzing
confusion''\citep[p462]{james1890principles} of moment-to-moment sensory
experience will have given way to a more coherent dance of objects and
forces, relata and causal relationships.  Such representations enable
humans to exploit their physical and social environments in flexible and
inherently model-based ways
\citep{dolan2013goals,sloman2005causal,griffiths2007two}.  An important
question, therefore is how people learn appropriate causal relationships
from the data they gather by observing and manipulating the world. Much
recent work on causal learning has used Pearl's
\citeyearpar{pearl2000causality} \emph{causal Bayesian network}
framework to demonstrate that people make broadly normative causal
inferences based on cues like observed contingencies or the outcomes of
interventions, which are manipulations or tests of the system
\citep[e.g.][]{lagnado2004advantage, lagnado2006time, lagnado2007cues,
  griffiths2009theory,holyoak2011causal}. Related work has begun to
explore how people engage in ``active learning'' -- selecting
interventions on variables in systems of interest in order to be effective
at reducing uncertainty about the true causal model \citep{bramley2015fcs,
  steyvers2003intervention, coenen2015strategies,sobel2006importance}.

Models of human causal learning based on Bayesian networks have tended to
focus on what Marr \citeyearpar{marr1982vision} called the {\em
  computational} level. This means that they consider the abstract
computational problem being solved and its ideal solution rather than the
actual cognitive processes involved in reaching that solution -- Marr's
{\em algorithmic} level. In practice the demands of computing and storing
the quantities required for exactly solving the problem of causal learning
are intractable for any non-trivial world and plausibly-bounded learner.
Even a small number of potential relata permit massive numbers of patterns
of causal relationships.  Moreover, real learning contexts involve noisy
(unreliable) relationships and the threat of exogenous interference,
further compounding the complexity of normative inference.  Navigating
this space of possibilities optimally would require maintaining
probability distributions across many models and updating all these
probabilities whenever integrating new evidence. This evidence might in
turn be gathered piecemeal over a lifetime of experience. Doing so
efficiently would require choosing maximally informative interventions, a
task which poses even greater computational challenges: consideration and
weighting of all possible outcomes, under all possible models for all
possible interventions \citep{nyberg2006intervention,murphy2001active}.

In order to understand better the cognitive processes involved in
learning causal relationships, we present a detailed exploration of how
people, with their limited processing resources, represent and reason
about causal structure. We begin by surveying existing proposals in the
literature.  We then draw on the literature on algorithms for
approximating probabilistic inference in computer science using these to
construct a new model.  We show that our new model captures the
behavioral patterns using a scalable and cognitively plausible algorithm
and explains why aggregate behavior appears noisily normative in the
face of individual heterogeneity.

Many existing experiments on human causal learning involve small numbers
of possible structures, semi-deterministic relationships and limited
choices or opportunities to intervene. These constraints limit the
computational demands on learners, and thus the need for heuristics or
approximations. Further, in most existing studies, subjects make causal
judgments only at the end of a period of learning, limiting what we can
learn about how their beliefs evolved as they observed more evidence,
and how this relates to intervention choice dynamics. One exception is
\cite{bramley2015fcs}, which explored online causal learning in
scenarios where participants' judgments about an underlying causal
structure were repeatedly elicited over a sequence of interventional
tests. Another is \cite{bramley2015staying}, which built on this
paradigm.  Both papers explained participants' judgments with accounts that
are not completely satisfying algorithmically, lacking cognitively
plausible or scalable procedures that could capture the ways in which
judgments and intervention choices deviated from the rational norms.
Here, we develop the algorithmic level account and demonstrate that it outperforms or equals competitors in modelling the data from both previous papers and a new experiment.  

The resulting class of algorithms embodies an old idea about theory
change known as the Duhem--Quine thesis \citep{duhem1991aim}.  The idea
can illustrated by a simile, originally attributed to Otto Van Neurath
\citeyearpar{neurath1983protocol} but popularized by Quine, who
writes:\\ 

``\emph{We} [theorists] \emph{are like sailors who on the open sea
  must reconstruct their ship but are never able to start afresh from the
  bottom. Where a beam is taken away a new one must at once be put there,
  and for this the rest of the ship is used as support.  In this way, by
  using the old beams and driftwood the ship can be shaped entirely anew,
  but only by gradual reconstruction.}''
\citeyearpar[][p3]{quine1969word}\\

\noindent The \emph{Neurath's ship} metaphor describes the piecemeal growth and evolution of scientific theories over the course of history.  In the metaphor, the theorist (sailor) is cast as relying on their existing theory (ship) to stay afloat, without the privilege of a dry-dock in which to make major improvements. Unable to step back and consider all possible alternatives, the theorist is limited to building on the existing theory, making a series of small changes with the goal of improving the fit.

We argue that people are in a similar position when it comes to their
beliefs about the causal structure of the world.  We propose that a
learner normally maintains only a single hypothesis about the global
causal model, rather than a distribution over all possibilities. They
update their hypothesis by making local changes (e.g. adding, removing
and reversing individual connections, nodes or subgraphs) while
depending on the rest of the model as a basis.  We show
that by doing this, the learner can end up with a relatively accurate
causal model without ever representing the whole hypothesis space or
storing all the old evidence, but that their causal beliefs will exhibit
a particular pattern of sequential dependence.  We provide a related
account of bounded intervention selection, based on the idea that
learners adapt to their own learning limitations when choosing what
evidence to gather next, attempting to resolve local rather than global
uncertainty.  Together, our \emph{Neurath's ship} model and
local-uncertainty-based schema for intervention selection provide a step
towards an explanation of how people might achieve a \emph{resource
  rational} \citep{griffiths2015rational,simon1982models} trade-off
between accuracy and the cognitive costs of maintaining an accurate
causal model of the world.

The paper is organized as follows.  We first formalize causal model
inference at the computational level.  We then highlight the ways in
which past experiments have shown human learning to diverge from the
predictions of this idealized account, using these to motivate two causal judgment heuristics proposed in the literature: \emph{simple endorsement} \citep{bramley2015fcs,fernbach2009causal} and \emph{win-stay, lose-sample} \citep{bonawitz2014win} before developing out own \emph{Neurath's ship} framework for belief change and active
learning.

We next show that participants' overall patterns of judgments and intervention choices are in line with the predictions of our framework across a variety of problems varying in terms of the complexity and noise in the true generative model, and whether the participants' are trained or must infer the noise.

We then compare models at the individual level, showing that all three causal-judgment proposals substantially outperform baseline and computational level competitors.  While our Neurath's ship provides the best overall fit, we find considerable diversity of strategies across participants. In particular, we find that the \emph{simple endorsement} heuristic emerges as a strong competitor.  We provide additional details about the formal framework and model specification in the Appendix.  Also, where indicated, additional figures are provided in Supplementary materials at \url{http://www.ucl.ac.uk/lagnado-lab/el/ns_sup}.

\section*{A computational-level framework for active structure learning}\label{section:ideal_observer}

Before presenting our theoretical framework, we lay out our
computational-level analysis of the problem of structure learning. This
can be broken down into three interrelated elements: (1) representing 
causal models (2) performing inference over possible models given
evidence (observations and the outcomes of interventions), and (3)
selecting interventions to gather more evidence and support this
inference.  We introduce the three elements here, providing more detail
where indicated in Appendix~\ref{app:a}.

\subsection*{Representation}

We use a standard representation for causal models -- the parameterized
directed acyclic graph \citep[][see
Figure~\ref{fig:cbn}a]{pearl2000causality}.  Nodes represent variables
(i.e. the component parts of a causal system); arrows represent causal
connections; and parameters encode the combined influence of parents
(the source of an arrow) on children (the arrow's
target)\footnote{Following standard graph nomenclature, we will often
  refer to the space between a pair of nodes in a model as an ``edge'',
  so that an acyclic causal model defines each edge as taking one of
  three states: forward $\rightarrow$, backward $\leftarrow$, or
  inactive $\varnothing$.}.  Such graphs can represent continuous
variables and any form of causal relationship; but here we focus on
systems of binary $\{0=\mathrm{absent},1=\mathrm{present}\}$ variables
and and assume generative connections -- meaning we assume that the
presence of a cause will always \emph{raise} the probability that the
effect is also present.\footnote{It is worth noting that these graphs
  cannot naturally represent cyclic or reciprocal relationships.
  However, there are various ways to extend the formalism as we discuss
  in General Discussion, and our theory is not fundamentally tied to a
  particular representation.}

We also adopt Cheng's Power PC \citeyearpar{cheng1997from} convention
for parameterization, which provides a simple way to capture how
probabilistic causal influences combine.  This assumes that causes have
independent chances of producing their effects, meaning the probability
that a variable takes the value $1$ is a noisy-OR combination of the
\emph{power} or strength $w_S$ of any active causes of it in the model
(we assume that this value is the same for all connections), together
with that of an omnipresent background strength $w_B$ encapsulating the
influence of any causes exogenous to the model.  We write
$\ww=\{w_S,w_B\}$. The probability that variable $x$ takes the value 1
is thus
\begin{equation}
P(x=1|pa(x),\ww)=1-(1-w_B)(1-w_S)^{\sum_{y\in
    pa(x)}y} 
\label{noisyor} 
\end{equation}
where $pa(x)$ denotes the parents of variable $x$ in the causal model
(see Figure~\ref{fig:cbn}a for an example).  For convenience, we
assume $\ww$ is the same for all components.\footnote{We also restrict
  ourselves to cases without any latent variable, although we note that
  imputing the presence of hidden variables is another important and
  computationally-challenging component of causal inference
  \citep{buchanan2010edge, kushnir2010inferring}.}

\subsection*{Inference} 

Each causal model $m$ over variables $X$ with strength and background
parameters $\ww$, assigns a probability to each datum ${\dd}=\{x\ldots
z\}$, propagating information from the variables that are fixed through
intervention $\eee$, to the others (see Figure~\ref{fig:cbn}b). The
space of all possible interventions $\mathcal{C}$ is made up of all
possible combinations of fixed and unfixed variables, and for each
intervention $\eee$ the possible data $\cald_\eee$ is made up of all
combinations of absent/present on the unfixed variables.  We use Pearl's
$\Do[.]$ operator \citep{pearl2000causality} to denote what is fixed on
a given test.  For instance, $\Do[x\!=\!1,y\!=\!0]$ means a variable $x$
has been fixed ``on'' and variable $y$ has been fixed ``off'', with all
other variables free to vary\footnote{We include the pure observation
  $Do[\varnothing]$ in $\mathcal{C}$.}.  Interventions allow a learner
to override the normal flow of causal influence in a system, initiating
activity at some components and blocking potential influences between
others.  This means they can provide information about the presence and
direction of influences between variables that is typically unavailable
from purely observational data \citep[see][ for a more detailed
introduction]{pearl2000causality,bramley2015fcs}, without additional
cues such as temporal information \citep{bramley2014order}. For
instance, in Figure~\ref{fig:cbn}b, we fix $y$ to 1 and leave $x$ and
$z$ free ($\eee=\Do[y\!=\!1]$).  Under the $x\rightarrow y\rightarrow z$
model we would then expect $x$ to activate with probability $w_B$ and
$z$ with a probability of $1-(1-w_B)(1-w_S)$.

In total, the probability of datum $\dd$, given intervention $\eee$, is
just the product of the probability of each variable that was not
intervened upon, given the states of its parents in the model 
\begin{equation}
  P({\dd}|m, \ww,\eee) = \prod\nolimits_{x\in (X\notin \eee)}
  P(x|\{{\dd},\eee\}_{pa(x)},\ww). \label{likelihood}
\end{equation}
where $\{{\dd},\eee\}_{pa(x)}$ indicates that those parents might either be observed (part of $\dd$) or fixed by the intervention (part of $\eee$).

In fully Bayesian inference, the true model is considered to be a
random variable $M$.  Our prior belief $P(M)$ is then an assignment of
probabilities, adding up to 1 across possible models $m\in M$ in the set
of models $\mathcal{M}$.  When we observe some data $D=\{{\dd}^i\}$,
associated with interventions $C=\{\eee^i\}$, we can update these
beliefs with Bayes theorem by multiplying our prior by the probability
of the observed data under each model and dividing by the weighted
average probability of those data across all the possible models:
\begin{equation}
 P(m|D, \ww;C)=\frac{P(D|m,\ww;C)P(m)}{\sum_{m' \in M}P(D|m',\ww;C)P(m')} .
\end{equation}
We will typically treat
the data as being independent and identically distributed, so
$P(D|m,\ww;C)=\prod_i P({\dd}^i|m,\ww;\eee^i)$.

If the data arrive sequentially (as
$D^t=\{{\dd}^1,\ldots,{\dd}^t\}$; and similarly for the
interventions), we can either store them and update at the end, or
update our beliefs sequentially, taking the posterior
$P(M|D^{t-1}, \ww;C^{t-1})$ at timestep $t-1$ as the new ``prior'' for
datum ${\dd}^t$.  If we are also unsure about the parameters of the
true model (i.e. $w_B$ and $w_S$) we have to treat them as random
variables too and average over our uncertainty about them to compute a
marginal posterior over models $M$ (see
Appendix~\ref{app:a}).

\begin{figure}[t]
   \centering
   \includegraphics[width = 0.8\columnwidth]{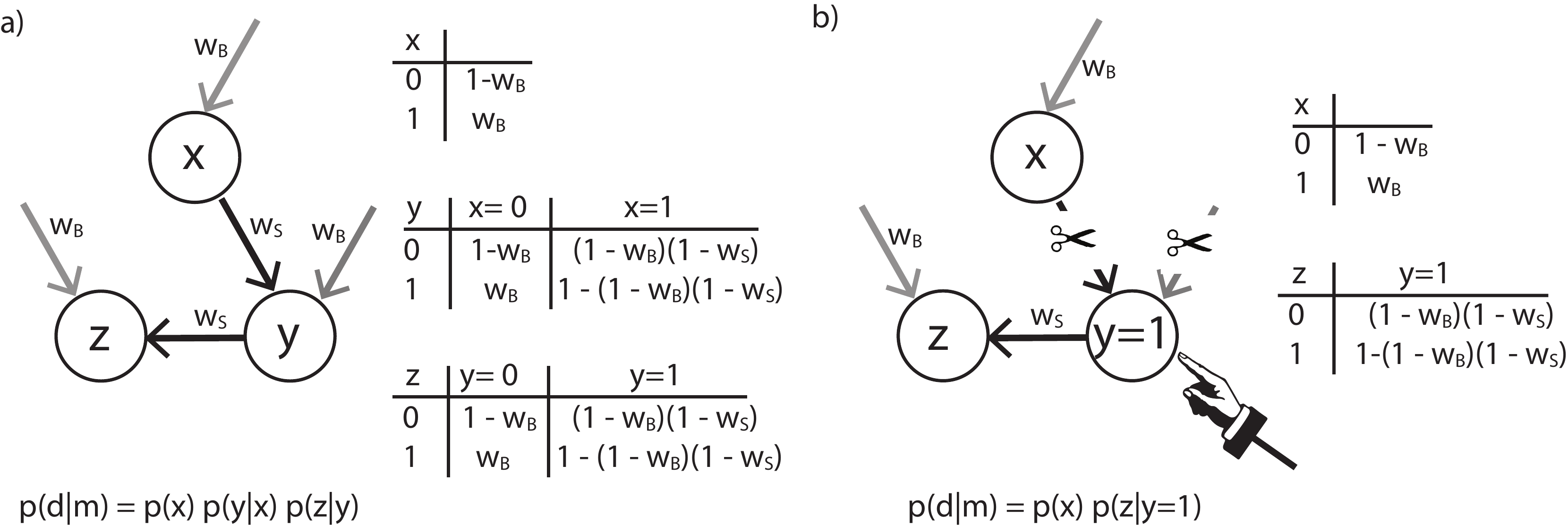}
   \caption{Causal model representation. a) An example causal Bayesian network, parametrized with strength
     $w_S$ and base rate $w_B$. The tables give the probability of each
     variable taking the value 1 conditional on its parents in the
     model and the omnipresent background noise rate $w_B$.  b)
     Visualization of intervention $\Do[y\!=\!1]$. Setting $y$ to 1
     renders it independent of its normal causes as indicated by the
     scissors symbols.}
   \label{fig:cbn}
\end{figure}

\subsection*{Choosing interventions}\label{section:rational_active_learning}

It is clear that different interventions yield different outcomes, which in turn have different probabilities under different models.  This means that which interventions are valuable for identifying the true model depends strongly on the hypothesis space and prior.  For instance fixing $y$ to 1 ($\Do[y\!=\!1]$) is (probabilistically) diagnostic if you are primarily unsure whether $x$ causes $z$ because $p(z|\Do[y\!=\!1])$ differs depending whether $pa(z)$ includes $x$.  However, it is not diagnostic if you are primarily unsure whether $x$ causes $y$ because y will take the value 1 the same regardless of whether $\pa(x)$ includes $y$.

The value of an intervention can be quantified relative to a notion of
uncertainty.  We can define the value of an intervention as the expected
reduction in uncertainty about the true model after seeing its
outcome.\footnote{Strictly this is greedy rather than optimal because
  planning several steps ahead can result in a different intervention
  being favored.  However, planning ahead was shown to make little
  difference for the similar problems explored in Bramley et al
  \citeyearpar{bramley2015fcs}.} To calculate this
expectation, we must average, prospectively, over the different possible
outcomes ${\dd}^\prime \in \mathcal{D}_{\eee}$ (recalling
$\mathcal{D}_{\eee}$ is the space of possible outcomes of intervention
$\eee$) weighted by their marginal likelihoods under the prior.  For a
greedily optimal sequence of interventions $\eee^1,\ldots,\eee^t$, we
take $P(M|D^{t-1},\ww;C^{t-1})$ as our prior each time. The most
valuable intervention $\eee^t$ at a given time point is then
\begin{equation}
\arg\max_{\eee\in \mathcal{C}}\E_{{\dd}^\prime \in \mathcal{D}_{\eee}}
\left[\Delta H(M|{\dd}^{\prime}, D^{t-1},\ww;C^{t-1},\eee)\right] ,
\label{eq:info_gain}
\end{equation}
where $\E[.]_{{\dd}^\prime \in \mathcal{D}_{\eee}}$ denotes the average over outcomes ${\dd}'$ and $\Delta H(.)$ denotes
reduction in uncertainty.  We use Shannon entropy
\citeyearpar{shannon1951prediction} to measure uncertainty (see
Appendix~\ref{app:a}). Shannon is just one of a broad family of possible entropy measures \citep{nielsen2011closed}.  However, it is one that has proved at least as long-run successful as a number of variants when applied as a greedy strategy for choosing interventions \citep{bramley2014should} or asking binary questions \citep{nelson2005finding}.

\section*{Behavioral patterns and existing explanations}

Unfortunately, both inference and choosing interventions scale so poorly
in the number of variables, they are fundamentally intractable for any
plausibly bounded learner
\citep{cooper1990computational,van2014rational}.  The number of possible
graphs grows rapidly with the number of variables they relate (3-, 4-
and 5-variable problems have 25, 543 and 29281 respectively).  Active intervention selection adds extra complexity because there are many possible interventions (3-,4- and 5- variable problems permit 27, 81 and 243 patterns of fixed ``on'', fixed ``off'' and free components), each of which might yield many outcomes (up to 8, 16 and 32 respectively, depending how many variables are left free to vary). All combinations of potential model, intervention and outcome should be averaged over in order to select the most valuable intervention.  This implies that people must find a considerably more economical way to approximate model inference while maintaining satisfactory accuracy.

It is therefore not surprising that behavioral learning patterns in existing studies exhibit marked divergence from the predictions of idealized Bayesian learning.  Participants' model judgments are typically robustly better than chance, yet poor when compared directly against an idealized Bayesian learner \citep{lagnado2002learning,lagnado2004advantage,bramley2014order,fernbach2009causal}.  Likewise, adults and even children have been shown to select interventions that are robustly more informative than chance, but much less efficient than idealized active learning \citep{bramley2015fcs,gureckis2009battleship,markant2012does,coenen2015strategies,lucas2014children,mccormack2016children,steyvers2003intervention}. 

More revealing than mere performance are the ways in which
participants' judgments diverge from these rational norms
\citep{anderson1990adaptive}.  \cite{bramley2015fcs} found that
participants' judgments in a sequential active causal learning task
resembled probability matching when lumped together, but that
individuals' trajectories were not well captured by simply adding
decision noise to the Bayesian predictions.  Individuals' sequences of
judgments were much too sequentially dependent, or ``sticky'', compared
to the Bayesian predictions, tending to remain the same or similar over
multiple elicitations as the objectively most likely structure shifted.
At the same time, when participants did change their judgments, they
tended to do so in ways that were consistent with the most recently
gathered evidence, neglecting evidence gathered earlier in
learning.  The result was a dual pattern of recency in terms of
judgments' consistency with the evidence, and stickiness in terms of
consistency with the previous judgments.  Bramley et al found that they
could capture these patterns with the addition of two parameters to
the Bayesian model.  The first was a forgetting parameter, encoding
trial-by-trial leakage of information from the posterior as it became
the prior for the next test.  The second was a conservatism parameter,
encoding a non-normatively high probability assigned to the latest
causal hypothesis.  While the resulting model captured participants
choices, it still made the implausible assumption that learners maintained
weighted probabilistic beliefs across the whole hypothesis space and
performed efficient active learning with respect to these.

As with Bramley et al, \cite{bonawitz2014win} found that children and adults' online structure judgments exhibited sequential dependence.  To account for this they proposed an account of how causal learners might rationally reduce the computational effort of continually reconsidering their model.  In their  ``win-stay, lose-sample'' scheme they suggest that learners maintain a single structural hypothesis, only resampling a new hypothesis from the posterior when they see something surprising under their current model, concretely, with a probability that increases as the most recent observation becomes less probable. This scheme guarantees that the learner's latest hypothesis is a sample from the posterior distribution at every point, but does not the require them to resample with every new trial.  While it captures the intuitive idea that people will tend to stick with a hypothesis until it fails to perform, ``win-stay, lose-sample'' still requires the learner to store all the past evidence to use when resampling, and does not provide a recipe for how the samples are drawn.\footnote{The authors mention that MCMC could be used to draw these samples without representing the full posterior.}

Another approach to understanding deviations between people's causal
judgments and rational norms comes from the idea that people construct
causal models in a modular or piecewise way.  For example,
\cite{waldmann2008causal} propose a \emph{minimal rational model} under
which learners infer the relationships between each pair of variables
separately without worrying about the dependencies between them, ending
up with a modular causal model that allows for good local inferences but
which leads to so-called ``Markov violations'' in more complex
inferences where participants fail to respect the conditional
dependencies and independences implied by the global model
\citep{rehder2014independence}.  They show that this minimal model is
sufficient to capture participants judgment patterns in two case
studies.  Building on this idea of locality, \cite{fernbach2009causal}
asked participants to make judgments following observation of several
preselected interventions.  They found that participants were
particularly bad at inferring chains, often inferring spurious
additional links from the root to the sink node (e.g. $x\rightarrow z$
as well as $x\rightarrow y$ and $y\rightarrow z$), a pattern also
observed in \cite{bramley2015fcs}.  Fernbach and Sloman proposed that this
was a consequence of participants inferring causal relationships through
local rather than global computations.  In the example, the
interventions on $x$ would normally lead to activations of $z$ due to
the indirect connection via $y$.  If learners attended only to $x$ and
$z$ there would be the appearance of a direct relationship.  They found
that they could better model participants by assuming they inferred each
causal link separately while ignoring the the rest of the model.
Embodying this principle, \citep{bramley2015fcs} proposed a \emph{simple
  endorsement} heuristic for online causal learning that would tend to
add direct edges to a model between intervened-on variables and any
variables that activated as a result, removing edges going to any
variables that didn't activate.  By doing this after each new piece of
evidence, the model exhibited recency as the older edges would tend to
be overwritten by newly inferred ones, as well as as capturing the
pattern of adding unnecessary direct connections in causal chains.  The
model did a good job of predicting participants' patterns but was
outperformed by the Bayesian model bounded with forgetting and
conservatism.  Additionally, like any heuristic, \emph{simple
  endorsement}'s success is conditional on its match to the situation.
For instance, \emph{simple endorsement} does badly in cases where there
are many chains -- meaning that the outcome of many interventions are
indirect, and also if the true $w_B$ is high.

Going beyond causal learning, sequential effects are ubiquitous in
cognition.  In some instances they can be rational; for instance moderate recency is rational in a changing world \citep{julier1997new}.  Regardless, there are a plethora of non
Bayesian models that can reproduce various sequential effects
\citep{decarlo1992intertrial,treisman1984theory,gilden2001cognitive}.  A
common class of these is based on the idea of adjusting an estimate
part way toward new evidence
\citep[e.g.][]{petrov2005dynamics,einhorn1986judging,rescorla1972theory}.
Updating point estimates means that a learner need not keep all the
evidence in memory but can instead make use of the location of the point(s) as a proxy for what was learned in the past.  
\cite{bramley2015staying} propose a model inspired by these ideas, that %
maintains a single hypothesis, but simultaneously attempts to minimize edits along with the number of variables' latest states that the current model fails to explain. The result is a model where the current belief acts as an anchor and the learner tends to try to explain the latest evidence by making the minimal number of changes to it.  Again, this model provided a good fit with participants' judgments, but did not provide a procedure for how participants were able to search the hypothesis space for the causal structure that minimized these constraints.

In summary, a number of ideas and models have been proposed in the
causal and active learning literatures.  By design, they all do a
good job of capturing patterns in human causal judgments.  However, it
is not clear that any of these proposals provide a general purpose,
scalable explanation for human success in learning a complex causal
world-model.  Some (e.g \emph{win-stay, lose-sample}) capture behavioral
patterns within the normative framework, but do not provide a scalable
algorithm.  Others (e.g. \emph{simple endorsement}) provide simple
scalable heuristics but may not generalize beyond the tasks they were
designed for, nor explain human successes in harder problems.  In the
next section we take inspiration from methods for approximate inference
in machine learning to construct a general purpose algorithm for
incremental structure change that satisfies both these desiderata.

\section*{Algorithms for causal learning with limited resources}\label{section:intro_approx}

We now turn to algorithms in machine learning that make approximate learning efficient in otherwise
intractable circumstances.  Additionally, research in these fields on
active learning and optimal experiment design has identified a range of
reasonable heuristics for selecting queries when the full expected
information calculation of (Equation~\ref{eq:info_gain}) is
intractable. We will take inspiration from some of these ideas to give a
formal basis to the intuitions behind the \emph{Neurath's ship}
metaphor.  We will then use this formal model to generate predictions that
we will compare to participants' behavior in several experiments.

\subsection*{Approximating with a few hypotheses}

One common approximation, for situations where a posterior cannot be
evaluated in closed form, is to maintain a manageable number of individual
hypotheses, or ``particles'' \citep{liu1998sequential}, with weights
corresponding to their relative likelihoods. The ensemble of particles
then acts as an approximation to the desired distribution.  Sophisticated
reweighting and resampling schemes can then filter the ensemble as data
are observed, approximating Bayesian inference.

These ``particle filtering'' methods have been used to explain how
humans and other animals might approximate the solutions to complex
problems of probabilistic inference. In associative learning
\citep{courville2007rat}, categorization \citep{sanborn2010rational} and
binary decision making \citep{vul2009one}, it has been proposed that
people's beliefs actually behave most like a single particle, capturing
why individuals often exhibit fluctuating and sub-optimal judgment while
maintaining a connection to Bayesian inference, particularly at the
population level.

\subsection*{Sequential local search}\label{section:intro_local_search}

The idea that people's causal theories are like particles requires they
also have some procedure for sampling or adapting these theories as
evidence is observed.  Another class of useful machine learning methods
involves generating sequences of hypotheses, each linked to the next via a
form of possibly stochastic transition mechanism. Two members of this
class are particularly popular in the present context: Markov Chain Monte
Carlo (MCMC) sampling, which asymptotically approximates the posterior
distribution; and (stochastic) hill climbing, which merely tries to find
hypotheses that have high posterior probabilities. 

MCMC algorithms involve stochastic transitions with samples that are typically easy to generate. Under various conditions, this implies that the sequences of (dependent) sample hypotheses form a Markov chain with a stationary distribution that is the full, intended, posterior distribution \citep{metropolis1953equation}. The samples will appear to ``walk'' randomly around space of possibilities, tending to visit more probable hypotheses more frequently. If samples are extracted from the sequence after a sufficiently long initial, so-called burn-in, period, and sufficiently far apart (to reduce the effect of dependence), they can provide a good approximation to the true posterior distribution. There are typically many different classes of Markov chain transitions that share the same stationary distribution, but differ in the properties of burn-in and subsampling.

The stochasticity inherent in MCMC algorithms implies that the sequence
sometimes makes a transition from a more probable to a less probable
hypothesis -- this is necessary to sample multi-modal posterior
distributions. A more radical heuristic is only to allow transitions to
more probable hypotheses --- this is called ``hill-climbing'',
attempting to find, and then stick at, the best hypothesis
\citep{tsamardinos2006max}. This is typically faster than a full MCMC
algorithm to find a good hypothesis, but is prone to become stuck in a
local optimum, where the current hypothesis is more likely than all its
neighbors, but less likely than some other more distant hypothesis.

Applied to causal structure inference, we might in either case consider transitions that change at most a single edge in the model \citep{cooper1992bayesian,goudie2011efficient}.  A simple case is Gibbs sampling \citep{geman1984stochastic}, starting with some structural hypothesis and repeatedly selecting an edge (randomly or systematically) and re-sampling it (either adding, removing or reversing) conditional on state of the other edges.  This means that a learner can search for a new hypothesis by making local changes to their current hypothesis, reconsidering each of the edges in turn, conditioning on the state of the others without ever enumerating all the possibilities.  By constructing a short chain of such ``rethinks'' a learner can easily update a singular hypothesis without starting from scratch.  The longer the chain, the less dependent or ``local'' the new hypothesis will be to the starting point.

The idea that stochastic local search plays an important role in cognition has some precedent \citep{sanborn2010rational,gershman2012multistability}.  For instance, \cite{abbott2012human} propose a random local search model of memory retrieval and \cite{ullman2012theory} propose an MCMC search model for capturing how children search large combinatorial theory spaces when learning intuitive physical theories like taxonomy and magnetism.  The idea that people might update their judgments by something like MCMC sampling is also explored by Lieder, Griffiths and Goodman \citeyearpar{lieder2012burn,liederanchor}.  They argue that under reasonable assumptions about the costs of resampling and need for accuracy, it can be rational to update one's beliefs by constructing short chains where the the updated judgment retains some dependence on its starting state, arguing that this might explain \emph{anchoring} effects \citep{kahneman1982judgment}. 

In addition to computational savings, updating beliefs by local search
can be desirable for statistical reasons.  If the learner has forgotten
some of the evidence they have seen, the location of their previous
hypothesis acts like a very approximate version of a sufficient
statistic for the forgotten information.  This can make it advantageous
to the learner to strike a good balance between editing their model to
account better for the data they can remember, and staying close to
their previous model to retain the connection to the data they have
forgotten \citep{bramley2015fcs}.

\section*{Neurath's ship: An algorithmic-level model of sequential belief change}\label{section:model_gibbs}

The previous section summarized two ideas derived from computer science
and statistics that provide a potential solution to the computational
challenges of causal learning: maintaining only a single hypothesis at a
time, and exploring new hypotheses using local search based on
sampling. In this section, we formalize these ideas to define a class of
models of causal learning inspired by the metaphor of Neurath's ship. We
start by treating interventions as given, and only focus on inference. We
then consider the nature of the interventions.

Concretely, we propose that causal learners maintaining only a single causal model (a single particle), $b^{t-1}$ and a collection of recent evidence and interventions $\cald_r^{t-1}$ and $\calc_r^{t-1}$ at time $t-1$.  They then make inferences by:

\begin{enumerate}
\item Observing the latest evidence ${\dd}^t$ and $\eee^t$ and adding it to the collection to make $\cald_r^{t}$ and $\calc_r^{t}$.
\item Then, searching for local improvements to $b^{t-1}$ by sequentially reconsidering edges
  $E_{ij}\in\{1: i\rightarrow j,~0: i \nleftrightarrow j, ~-1: i
  \leftarrow j\}$ (adding, subtracting or reorienting them) conditional
  on the current state of the edges in the rest of their model
  $E_{\setminus ij}$ -- e.g. with probability $ P(E_{ij}|E_{\setminus
    ij}, \cald_r^{t},\calc_r^t,\ww)$.
  \item After searching for $k$ steps, stopping and taking the latest
  version of their model as their new belief $b^t$.  If $b^t$ differs from $b^{t-1}$ the evidence is
  forgotten ($\cald_r^t$ and $\calc_r^t$ become $\{\}$), and they begin collecting evidence again.
\end{enumerate}
A detailed specification of this process is given in
Appendix~\ref{app:a}.

Starting with any hypothesis and repeatedly resampling edges conditional
on the others is a form of Gibbs sampling \citep{goudie2011efficient}.
Further, the learner can make use of the data they have forgotten by
starting the search with their current belief $b^{t-1}$, since these
data are represented to some degree in the location of $b^{t-1}$.
Resampling using the recent data $P(M|\cald_r^t,\calc_r^t,\ww)$ allows the
learner to adjust their beliefs to encapsulate better the data they have
just seen, and let this evidence fall out of memory once it has
been incorporated into the model. %

\subsection*{Resampling, hill climbing or random change}

Following the procedure outlined above, the learner's search steps would
constitute dependent samples from the posterior over structures given $\cald^t_r$.  However, it is also plausible that learners
will try to hill-climb rather than sample, preferring to move to more
probable local models more strongly than would be predicted by Gibbs
sampling.  In order to explore this, we will consider generalizations of
of the update equation allowing transitions to be governed by powers of
the conditional edge probability (i.e.  $P^\omega(E_{ij}=e| E_{\setminus
  ij},\cald^t_r,\calc^t_r,\ww)$), yielding stronger or weaker preference for
the most likely state of $E_{ij}$ depending whether $\omega>1$ or $<1$.
By setting $\omega$ to zero, we would get a model that does not learn
but just moves randomly between hypotheses, tending to remain local and
by setting it to infinity we would get a model that always moved to the
most likely state for the edge.

\subsection*{Search length}

It is reasonable to assume that the number of search steps $k$ that a
learner performs will be variable, but that their \emph{capacity} to
search will be relatively stable.  Therefore, we assume that
for each update, the learner searches for $k$ steps, where $k$ is drawn
from a Poisson distribution with mean $\lambda\in[0,\infty]$.

The value of $\lambda$ thus determines how sequentially dependent a
learner's sequences of beliefs are.  A large $\lambda$ codifies a tendency
to move beliefs a long way to account for the latest data $\cald_r^t$ at the
expense of the older data -- retained only in
the location of the previous belief $b^{t-1}$ -- while a moderate
$\lambda$ captures a reasonable trade-off between starting state and new
evidence, and a small $\lambda$ captures \emph{conservatism},
i.e. failure to shift beliefs enough to account for the latest data.\footnote{Note that we later cap $k$ at 50 when estimating our model having established that search lengths beyond these bounds made negligible difference to predictions.}

\subsection*{Putting these together}

By representing the transition probabilities from model $i$ to model
$j$, for a particular setting of hill climbing parameter $\omega$ and
data $\cald_r^t$, with a transition matrix $R^\omega_t$, we can thus make
probabilistic predictions about a learner's new belief $b^t\in B^t$.\footnote{We define this matrix formally in Appendix~\ref{app:a}.  Note
  that we assume transitions that would create a loop in the overall
  model get a probability of zero.  This assumption could be dropped for
  learning dynamic Bayesian networks but is necessary for working with
  directed acyclic graphs.}
The probabilities depend on the previous belief $b^{t-1}$ and their
average search length $\lambda$.  By averaging over different search
lengths with their probability controlled by $\lambda$, and taking the
requisite row of the resulting transition matrix we get the following
equation
\begin{equation}
P(b^t=m|\cald_r^t, \calc_r^t,b^{t-1}, \omega, \lambda)={\sum\limits_{0}^\infty}
\frac{\lambda^k e^{-\lambda}}{k!} [(R^\omega_t)^k]_{b^{t-1}m} 
\label{eq:model_gibbs}
\end{equation}
Note that this equation describes the probability of a Neurath's
ship style search terminating in a given new location. The learner
themselves need only follow the four steps described above, {\it
  sampling\/} particular edges and search length rather than averaging
over the possible values of these quantities.  See
Appendix~\ref{app:a} for more details and
Figure~\ref{fig:neurath_gibbs_example} for an example.

\begin{figure}[t]
   \centering
   \includegraphics[width = \columnwidth]{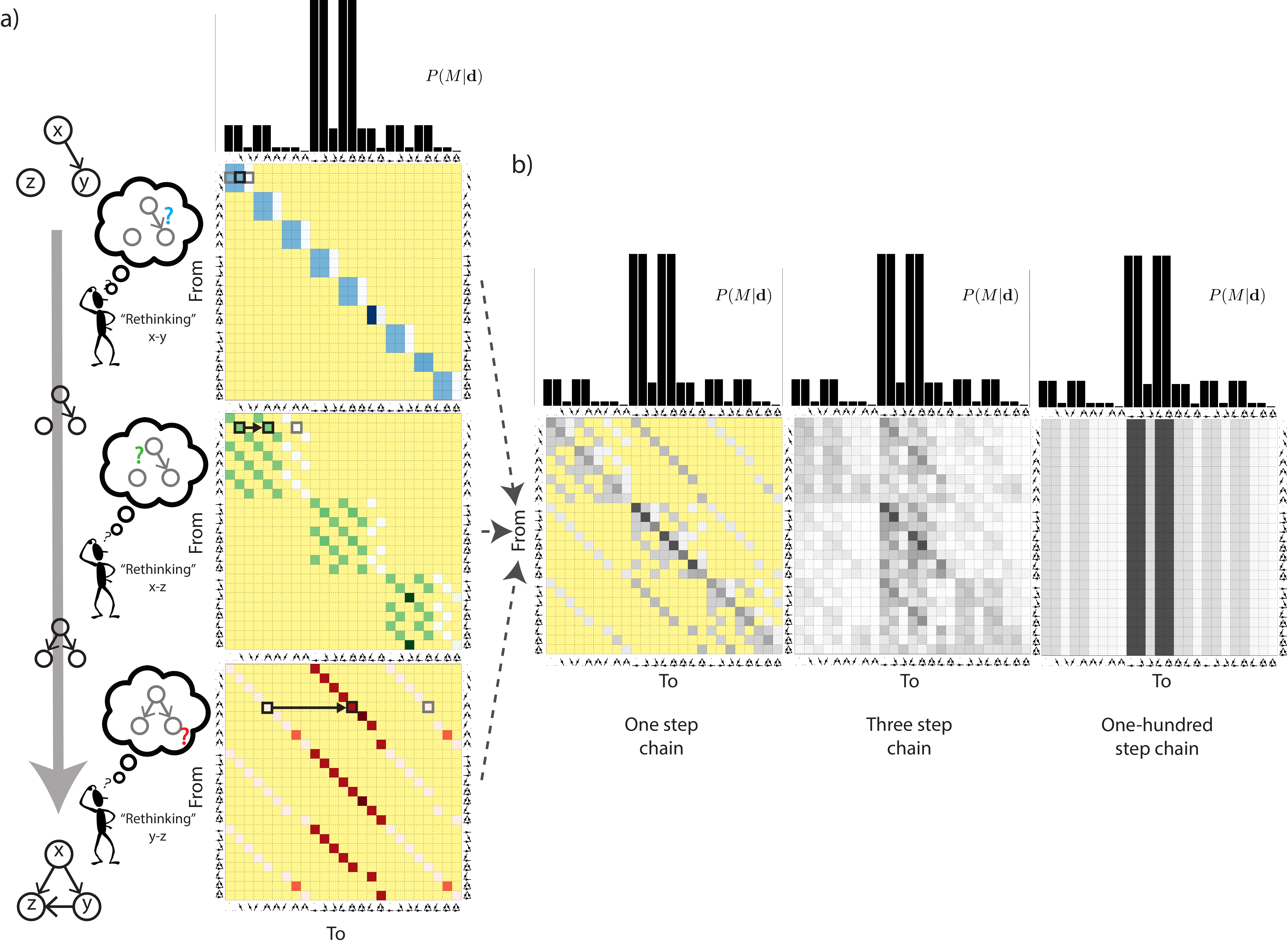}
   \caption{An illustration of NS model of causal belief updating. a) An example search path: The learner starts out with a singly connected model at the top ($x\rightarrow y$ connection only).  They update their beliefs by resampling one edge at a time $e\in\{\rightarrow, \nleftrightarrow, \leftarrow\}$.  Each entry $i,j$ in the matrices gives the probability of moving from model in the row $i$ to the model in the column $j$ when resampling the edge marked with the colored question mark.  Lighter shades of the requisite color indicate low transition probability, darker shades indicate greater transition probability; yellow is used to indicate zero probabilities.  Here the learner stops after resampling each edge once, moving from $b^{t-1}$ of $[x\rightarrow y]$ to $b^t$ of $[ x\rightarrow y, x\rightarrow z, y\rightarrow z]$. b) Assuming the edge to resample is chosen at random, we can average over the different possible edge choices to derive a 1-step Markov chain transition matrix $R_t^\omega$ encompassing all the possibilities.  By raising this matrix to higher powers we get the probability of different end points for searches of that length. If the chain is short (small $k$) the final state depends heavily on the starting state (left) but for longer chains (large $k$), the starting state becomes less important, getting increasingly close to independent sampling from the desired distribution (right).}
   \label{fig:neurath_gibbs_example}
\end{figure}

\section*{Selecting interventions on Neurath's ship: A local uncertainty schema}\label{section:local_uncertainty_schema}

In situations where a posterior is already hard to evaluate, calculating
the globally most informative intervention -- finding the intervention
$\eee^t$ that maximizes Equation~\ref{eq:info_gain} -- will almost always
be infeasible. Therefore, a variety of heuristics have been developed that
allow tests to be selected that are more useful than random selection, but
do not require the full expected information gain be computed
\citep{settles2012active}.  These tend to rely on the learners' current,
rather than expected, uncertainty (e.g. \emph{uncertainty sampling}
which chooses based on outcome uncertainty under the prior) or the
predictions under just a few favored hypotheses (e.g. \emph{query by
  committee}) as a substitute for the full expectancy calculation.  The
former relies on maintaining a complete prior distribution, making the
latter a more natural partner to the \emph{Neurath's ship} framework.

We have proposed a model of structure inference under which learners are
 only able to consider a small set of of alternatives at a time, and only
 able to generate alternatives that are ``local'' in some dimension.
 Locally driven intervention selection is a natural partner to this for at
 least two reasons: (1) Under the constraints of the \emph{Neurath's ship}
 framework, learners would not be able to work with the prospective
 distributions required to estimate global expected informativeness, but
 could potentially estimate expected informativeness with respect to a
 sufficiently narrow sets of alternatives. (2) Evidence optimized to
 distinguishing local possibilities (focused on one edge at a time for
 instance) might better support sequential local belief updates (of the
 kind emphasized in our framework) than the globally most informative
 evidence \citep{patil2014optimal}. In line with this, we propose one way in which learners might select robustly informative interventions by attempting only to distinguish a few ``local'' possibilities at a time, requiring only ``local'' uncertainty estimates to target the possibilities on which to focus \citep{markant2015self}.

The idea that learners will focus on distinguishing only a few alternatives at a time requires specifying how they choose which of the many possible subsets of the full hypothesis space to target with a particular test. Queries that optimally reduce expected uncertainty about one local aspect of a problem are liable to differ from those that promise high global uncertainty reduction.  For example, Figure~\ref{fig:example_local_interventions}b shows two trials taken from our experiments, and shows that the expected values of each of a range of different intervention choices (shown in Figure~\ref{fig:example_local_interventions}a) are very different depending whether the learner is focused on resolving global uncertainty all at once, or on resolving some specific ``local'' aspect of it.  This illustrates the idea that a learner might choose a test that is optimally informative with respect to a modest range of options that they have in mind at the time (e.g. models that differ just in terms of the state of $E_{xz}$) yet appear sporadically inefficient from the perspective of greedy global uncertainty reduction. Furthermore, by licensing quite different intervention preferences, they allow us to diagnose individual and trial-by-trial differences in focus preference.

In the current work, we will consider three possible
varieties of focus, one motivated by the \emph{Neurath's ship} framework
(\emph{edge} focus) and two inspired by existing ideas about bounded
search and discovery in the literature (\emph{effects} focus and
\emph{confirmation} focus).  While these are by no means exhaustive they
represent a reasonable starting point.

\subsection*{The two stages of the schema}

\renewcommand{\lll}{\mathbf{l}}

The idea that learners focus on resolving local rather than global
uncertainty results in a metaproblem of choosing what to focus on next,
making intervention choice a two stage process. We write $L$ for the set of all possible foci $\lll$, and $\call\subset L$ for the subset of
possibilities that the learner will consider at a time, such as the 
the state of a particular edge or the effects of a particular
variable. The procedure is:
\begin{itemize}[align=left]
\item[\textbf{Stage 1}] Selecting a local focus $\lll^t\in \call$ 
\item [\textbf{Stage 2}] Selecting an informative test $\eee^t$ with respect to the chosen focus $\lll^t$
\end{itemize}

Different learners might differ in the types of questions they consider,
meaning that $\call$ might contain different varieties and
combinations of local focuses.  We first formalize the two stages of the
schema, and then propose three varieties of local focus that learners
might consider in their option set $\call$ that differ in terms of
which and how many alternatives they include.

As mentioned above, we assume that the learner has some way of
estimating their current local confidence.  We will assume confidence
here is approximately the inverse of uncertainty, so assume for
simplicity that learners can calculate uncertainty from the evidence
they have gathered since last changing their model in the form of the
entropy $H(\lll|\cald^t_r,\ww;\calc^t_r)$ for all $\lll\in \call$ (the
assumption we examine in the discussion). They then choose
(\textbf{Stage 1}) the locale where these data imply the least certainty
\begin{equation}
\lll^{t}=\arg\max_{\lll\in \call}H(\lll|b^{t-1},\cald^t_r,\ww;\calc^t_r)
\label{eq:focus_value}
\end{equation}
However, in carrying out \textbf{Stage 2} we make the radical assumption
that learners do not use $P(\lll^{t}|\cald_r^t,b^{t-1},\ww;\calc_r^t)$, but
rather, consistent with the method of inference itself, only consider
the potential next datum $\dd'$. This means that the intervention
$\eee^t$ itself is chosen to maximize the expected information about
$\lll^t$, ignoring pre-existing evidence, and using what amounts to a
uniform prior.  Specifically, we assume that $\eee^t$ is chosen as
\begin{equation}
  \eee^t = \arg\max_{\eee\in \mathcal{C}}\E_{{\dd}\in
  \mathcal{D}_{\eee}}\left[\Delta
  H(\lll^t|{\dd},\ww,b^{t-1};\eee)\right]
\label{eq:choice_value}
\end{equation}
where we detail the term in the expectation below for the three types of
focuses.  

Assuming real learners will exhibit some decision noise, we can model both choice of focus and choice of intervention relative to a focus as soft \citep{luce1959choice} rather than strict maximization giving focus probabilities
\begin{equation}
P(\lll^{t}|\cald_r^t,b^{t-1},\ww;\calc_r^t)=\frac{\exp(H(\lll^t|\cald_r^t,b^{t-1},\ww;\calc_r^t)\rho)}{\sum_{\lll\in \call}\exp(H(\lll|\cald_r^t,b^{t-1},\ww;\calc_r^t)\rho)}
\label{eq:stage1} 
\end{equation}
governed by some inverse temperature parameter $\rho$, and choice probabilities
\begin{equation}
  P(\eee^t|\lll,\ww,b^{t-1}) = \frac{\exp(\E_{{\dd}'\in \mathcal{D}_{\eee}}\left[\Delta H(\lll|{\dd'},\ww,b^{t-1};\eee^t)\right]\eta) }{\sum_{c\in\mathcal{C}}\exp(\E_{{\dd}'\in \mathcal{D}_{\eee}}\left[\Delta H(\lll|{\dd}',\ww,b^{t-1};\eee)\right]\eta) }
  \label{eq:stage2}
\end{equation}
governed by inverse temperature $\eta$.

\subsection*{Three varieties of local focus}

\subsubsection*{Edges}

An obvious choice, given the \emph{Neurath's ship} framework, would be for learners to try to distinguish alternatives that differ in terms of a single edge (Figure~\ref{fig:example_local_interventions}a), i.e. those they would consider during a single update step.  

For a chosen edge $E_{xy}$ we can then consider a learner's goal to be to maximise their expectation of

\begin{equation}
\Delta H(E_{xy}|E^{t-1}_{\setminus xy}, {\dd}, \ww;\eee)
\label{eq:edge}
\end{equation}
(see Appendix~\ref{app:a} for the full local entropy equations).  Note that Equation~\ref{eq:edge} is a refinement of Equation~\ref{eq:choice_value} for the case of focusing on an edge, from $b^{t-1}$ the learner need only condition on the other edges $E^{t-1}_{\setminus xy}$.  This goal results in a preference for fixing one of the nodes of the target edge ``on'', leaving the other free, and depending on the other connections in $b^{t-1}$, either favors fixing the other variables ``off'' or is indifferent about whether they are ``on'', ``off'' or ``free'' (Figure~\ref{fig:example_local_interventions}b).  For an edge focused local learner, the set of possible focuses includes all the edges $\call\in\forall_{i< j\in N}E_{ij}$.

\subsubsection*{Effects}

A commonly proposed heuristic for efficient search in the deterministic domains is to ask about the dimension that best divides the hypothesis space, eliminating the greatest possible number of options on average.  This is variously known as ``constraint-seeking'' \citep{ruggeri2014learning} or ``the split half heuristic'' \citep{nelson2014children}.  In the case of identifying the true deterministic ($w_S=1$ and $w_B=0$) causal model on $N$ variables through interventions it turns out that the best split is achieved by querying the effects of a randomly chosen variable, essentially asking: ``What does $x$ do?'' (Figure~\ref{fig:example_local_interventions}a)\footnote{This is also the most globally informative type of test relative to a uniform prior in all of the noise conditions we consider in the current paper}.  Formally we might think of this question as asking: which other variables (if any) are descendants of variable $x$ in the true model?  This a broader focus than querying the state of a single edge, but considerably simpler question than the global ``which is the right causal model?'' because the possibilities just include the different combinations of the other variables as effects (e.g. \emph{neither}, \emph{either} or \emph{both} of $y$ and $z$ are descendants of $x$ in a 3-variable model) rather than the superexponential number of model possibilities\footnote{The number of directed acyclic graphs on N nodes, $|\mathcal{M}|_N$, can be computed with the recurrence relation $|\mathcal{M}|_N= \sum_{k\in N}(-1)^{k-1}\binom{N}{2}2^{k(N-k)}|\mathcal{M}|_{N-1}$ \citep[see][]{robinson1977counting}}.

Relative to a chosen variable $x$, we can write an \emph{effect focus}
goal as maximizing the expectation of
\begin{equation}
\Delta H(\mathrm{De}(x)|{\dd},\ww;\eee)
\label{eq:effects}
\end{equation}
where $\mathrm{De}(x)$ is is the set of $x$'s direct or indirect descendants.  This focus does not depend on $b^{t-1}$.  This goal results in a preference for fixing the target node ``on'' (e.g. $\Do[x\!=1\!]$) and leaving the rest of the variables free to vary (Figure~\ref{fig:example_local_interventions}b).  For an effect focused local learner, the set of possible focuses includes all the nodes $\call\in\forall_{i\in X}\mathrm{De}(X\{i\})$.

\subsubsection*{Confirmation}

Another form of local test, is to seek evidence that would confirm or refute the current hypothesis, against a single alternative ``null'' hypothesis.  Confirmatory evidence gathering is a ubiquitous psychological phenomenon \citep{nickerson1998confirmation,klayman1989hypothesis}. Although confirmation seeking is widely touted as a bias, it can also be shown to be optimal, for example under deterministic or sparse hypotheses spaces or peaked priors \citep{navarro2011hypothesis, austerweil2011seeking}.

Accordingly, Coenen et al.~\citeyearpar{coenen2015strategies} propose that causal learners adopt a ``positive test strategy'' when distinguishing causal models.  They define this as a preference to ``turn on'' a parent component of one's hypothesis -- observing whether the activity propagates to the other variables in the way that this hypothesis predicts.  They find that people often intervene on suspected parent components, even when this is uninformative, and do so more often under time pressure.  In Coenen et al's tasks, the goal was always to distinguish between two hypotheses, so their model assumed people would sum over the number of descendants each variable had under each hypotheses and turn on the component that had the most descendants on average.  However, this does not generalize to the current, unrestricted, context where all variables have the same number of descendants if you average over the whole hypothesis space. However, Steyvers et al \citeyearpar{steyvers2003intervention} propose a related \emph{rational test model} that selects interventions with a goal of distinguishing a single current hypothesis from a null hypothesis that there is no causal connection.

Following Steyvers et al.~\citeyearpar{steyvers2003intervention}, for a \emph{confirmatory} focus we consider interventions expected to best reduce uncertainty between the learner's current hypothesis $b^{t-1}$ and a null $b^0$ in which there are no connections (Figure~\ref{fig:example_local_interventions}a).

\begin{equation}
\Delta H(\{b^t, b^0\}|b^{t-1},{\dd},\ww;\eee)
\label{eq:confirmatory}
\end{equation}
This goal results in a preference for fixing on the root node(s) of the
target hypothesis (Figure~\ref{fig:example_local_interventions}c ii, noting the confirmation focus favours $\Do[x\!=\!1, y\!=\!1]$ here).  The effectiveness of confirmatory focused
testing depends on the level of noise and the prior, becoming
increasingly useful later once the model being tested has sufficiently
high prior probability.  For a confirmation focused learner there is always just a single local focus.

\subsection*{Implications of the schema}

The local uncertainty schema implies that intervention choice depends on
two separable stages.  Thus it accommodates the idea that a learner
might be poor at choosing what to focus on but good at selecting an
informative intervention relative to their chosen focus.  It also allows
that we might understand differences in learners' intervention choices
as consequences of the types of local focus they are inclined or able to
focus on.  Learners cognizant of the limitations in their ability to
incorporate new evidence might choose to focus their intervention on
narrower questions (i.e. learning about a single edge at a time) while
others might focus too broadly and fail to learn effectively.  In the current work we will fit behavior assuming that learners choose between these local focuses, using their patterns to diagnose which local focuses they include in their option set $\call$, which of these they choose on a given test $l^t$ and finally how these choices relate to their final performance.

\begin{figure}[t]
   \centering
   \includegraphics[width = \columnwidth]{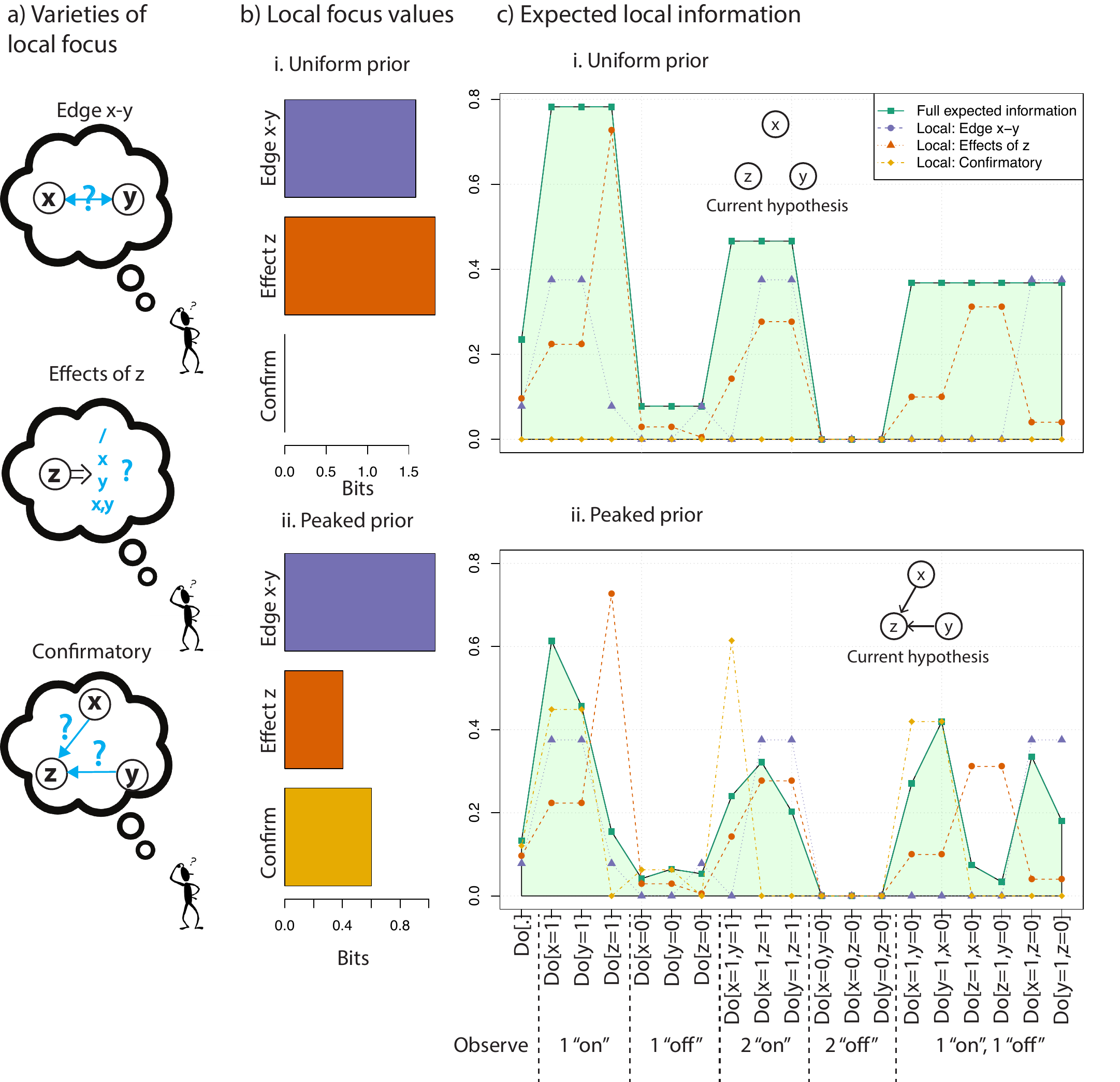}
   \caption{An illustrative example of local focused uncertainty minimization a) Three possible ``local'' focuses.  b) The value of these choices of focus according to their current uncertainty Equation~\ref{eq:focus_value} (i.) at the start of learning  and (ii.) after several tests have been performed.  Note that uncertainty is measured with Shannon entropy based on the local possibilities and $\cald_r^t$ and that confirmation is undefined at the start of learning where both current and null hypothesis are that there are no connections in the model.  c) Expected value of 19 different interventions assuming: global expected information gain from the true prior (green squares, and shaded), effects of $z$ focus (red circles), the relationship between $x$ and $y$ (blue triangles) and confirming $b^{t-1}$ (yellow diamonds), assuming a uniform prior over the requisite possibilities and a known $w_S$ and $w_B$ of .85 and .15.
}
   \label{fig:example_local_interventions}
\end{figure}

\section*{Comparing model predictions to experiments}
 
The \emph{Neurath's ship} framework we have introduced has two distinct
signatures. Making only local edits from a single hypothesis results in
sequential dependence.  Making these edits by local resampling leads to
aggregate behavior that can range between probability matching and hill
climbing--which can give better short term gains but with a tendency to
get stuck in local optima.  Two of the other heuristics also lead to sequential dependence.
\emph{Win-stay lose-sample} predicts all-or-none dependence whereby
learners' judgments will either stay the same or jump to a new location
that depends only on the posterior. \emph{Simple endorsement} also predicts recency, although is
distinguished by its failure to separate direct from indirect effects
of interventions, leading to a different pattern of structural change. 

In terms of interventions, if participants are locally focused, we expect their hypotheses to deviate from optimal predictions in ways that can be accommodated by our local uncertainty schema, i.e. selecting interventions that are more likely to be targeted toward local rather than global uncertainty. If learners do not maintain the full posterior, we expect their intervention distributions to be relatively insensitive to the evidence that has already been seen, while still being locally informative. If people disproportionately focus on identifying \emph{effects}, we expect to see relatively unconstrained interventions with one variable fixed ``on'' at a time.  If people focus on individual \emph{edges} we expect more constraining interventions with more variables fixed ``off''.  If \emph{confirmatory} tests are employed, we expect to see more interventions on putative parents than on child nodes.

We first compare the predictions of our framework to existing data from \cite{bramley2015fcs}.  We then report on three new experiments designed to further test the specific predictions of our framework.

\subsection*{\cite{bramley2015fcs}}

In \cite{bramley2015fcs}, participants interacted with five probabilistic causal systems involving 3 variables (see Figure~\ref{fig:weighted_judgments_e1}a), repeatedly selecting interventions (or tests) to perform in which any number of the variables are either fixed ``on'' or ``off'', while the remainder are left free to vary. The tests people chose, along with the parameters $\ww$ of the true underlying causal model, jointly determined the data they saw.  In this experiment $w_S$ was always .8 and $w_B$ was always .1.  After each test, participants registered their best guess about the underlying structure. They were incentivised to report their best guess about the structure, through receipt of a bonus for each causal relation (or non-relation) correctly registered at the end.  There were three conditions:  \emph{no information} (N=79) was run first. After discovering that a significant minority of participants performed at chance, condition  \emph{information} (N=30), added a button that participants could hover over and remind themselves of the key instructions during the task (the noise, strengths, the goal) and condition \emph{information + summary} (N=30) additionally provided a visual summary of all previous tests and their outcomes.\footnote{In the paper this was reported as two experiments, the second with two between-subjects conditions.  They share identical structure they were subsequently analyzed together.  Therefore we do the same here, reporting as a single experiment with three between-subjects conditions.}  Participants could draw cyclic causal models if they wanted (e.g. $x\rightarrow y\rightarrow z \rightarrow x$)  and  were not forced to select something for every edge from the start but instead could leave some or all of the edges as ``?''.  Once a relationship was selected they could not return to ``?''.  The task is available online at \url{http://www.ucl.ac.uk/lagnado-lab/el/ns_sup}.

\subsubsection*{Comparing judgment patterns}

We compared participants' performance in \cite{bramley2015fcs} to that of several simulated learners.  \emph{Posterior} draws a new sample from the posterior for each judgment.  \emph{Random} simply draws a random graph on each judgment.  \emph{Neurath's ship} follows the procedure detailed in the previous section, beginning with its previous judgment ($b^{t-1}$, or an unconnected model at t=1) and reconsidering one edge at a time based on the evidence gathered since its last change $\cald_r^t$ for a small number of steps after observing each outcome. We illustrate this with a simulation with a short mean search length $\lambda$ of 1.5 and behavior $\omega$ of 10 corresponding moderate hill climbing. \emph{Win-stay, lose-sample} sticks with the previous judgment with probability $1-P(D^t|b^{t-1}\ww;C^{t})$ or alternatively samples from the full posterior.  The \emph{simple endorser} always adds edges from any intervened-upon variables to any activated variables on each trial, and removes them from any intervened-upon variables to any non-activated variables, overwriting any edges going in the opposing direction.  Participants' final accuracy in \cite{bramley2015fcs} was closest to the \emph{Neurath's ship} as is clear in Figure~\ref{fig:ns_fcs}a and b.  That the Neurath's ship simulation unperformed participants in condition \emph{information + summary} is to be expected since these participants were given a full record of past tests while \emph{Neurath's ship} uses only the recent data.

Additionally, participants' online judgments exhibited sequential dependence.  This can be seen in Figure \ref{fig:ns_fcs}b comparing the distribution of edits (bars) to the markedly larger shifts we would expect to see assuming random or Bayesian posterior sampling on these trials (black full and dotted lines).  The overall pattern of edit distances from judgment to judgment is commensurate with those produced by the \emph{Neurath's ship} procedure (red line), but also, here by \emph{win-stay, lose-sample} (blue line) and \emph{simple endorser} (green line) simulations.

\subsubsection*{Comparing intervention patterns}

To compare intervention choices to global and locally driven intervention selection, we simulated the task with the same number of simulations as participants, stochastically generating the outcomes of the simulations' intervention choices according to the true model and true $\ww$ (which the participants knew).  Simulated efficient active learners would perfectly track the posterior and always select the greediest intervention (as in Equation~\ref{eq:info_gain}).

 We also compared participants' interventions to those of several other simulated learners, each restricted to one of the three types of local focus introduced in Section~4 (`edge'. `effects' or `confirmation').\footnote{We assumed these tests were chosen based on a uniform prior over the options considered.  We used the latest most probable judgment $\text{argmax} p(M|D^{t-1},\ww)$ in place of a current hypothesis $b^{t-1}$ for edge focused and confirmatory testing so as not to presuppose a particular belief update rule in assessing intervention selection.}   When one of the simulated learners did not generate a unique best intervention, it would sample uniformly from the joint-best interventions according to that criterion. The results of the simulations %
 are visualized in Figure~\ref{fig:ns_fcs}c and d.

Participants' intervention choices in \cite{bramley2015fcs} were clearly more informative than random selection but less so than ideal active learning.  This is evident in Figure~\ref{fig:ns_fcs}c comparing participants (bars) to simulations of ideal active learning (black circles) and random intervening (black squares), and in Figure~\ref{fig:ns_fcs}d comparing the participants (red lines) to the ideal active learning (pink lines) and random intervening (blue lines) simulations.  Furthermore, the informativeness of participants' interventions is in the range of the simulations of any of the three local foci (yellow, green and blue lines).

As we see in Figure~\ref{fig:ns_fcs}d, idealized active learning favored fixing one variable on at a time ($\Do[x\!=\!1]$, $\Do[y\!=\!1]$ etc, hereafter called ``one-on'' interventions) for the majority of tests.  It always chose ``one-on'' for the first few tests but would sometimes select controlled (e.g. $\Do[x\!=\!1, y\!=\!0]$) tests on later tests when the remaining uncertainty was predominantly between direct and indirect causal pathways as in between chain, fork and fully connected structures.

Locally driven testing had different signatures depending on the focus.  The \emph{edge} focused simulation would fix the component at one end of their edge of interest ``on'' and leave the component at the other end ``free''.  What it did with the third component depended on its latest judgment about the network.  If, according to $b^{t-1}$, another component was a cause of the component that was left free-to-vary, the simulation favored fixing it ``off''.  Otherwise, it did not distinguish between ``on'', ``off'' or ``free'' choosing one of these at random.  The resulting pattern is a spread across ``one-on'', ``two-on'' and ``one-on, one-off'' tests with a bias toward controlled ``one-on, one-off'' tests.  The effects focused learner always favored ``one-on'' interventions.  The \emph{confirmation} focused tester would generally fix components with children in $b^{t-1}$ on, and leave components with parents in $b^{t-1}$ free.  This led to the choice of a mixture of ``one-on'' and ``two-on'' interventions.%

Like the \emph{ideal} or the \emph{effects} focused simulations,
participants in \cite{bramley2015fcs} strongly favored ``one-on'' tests.   Consistent with
\emph{confirmatory} testing, components with at least one child
according to the latest hypothesis $b^{t-1}$ were more
likely to be fixed ``on'' than components believed to have no children
(60\% compared to 56\% of the time $t(24568)=3.2,
p=.001$).\footnote{We ran the same number of simulated learners as participants in each experiment and condition to facilitate statistical comparison.}  Participants' intervention selections were markedly less
dynamic across trials than those of the efficient learner.  For example,
the proportion of single (e.g. $\do[x\!=\!1]$) interventions decreased
only fractionally on later tests, dropping from 78\% to 73\% from the first to the last test.

\begin{figure}[t]
  \centering
   \includegraphics[width = 0.75\columnwidth]{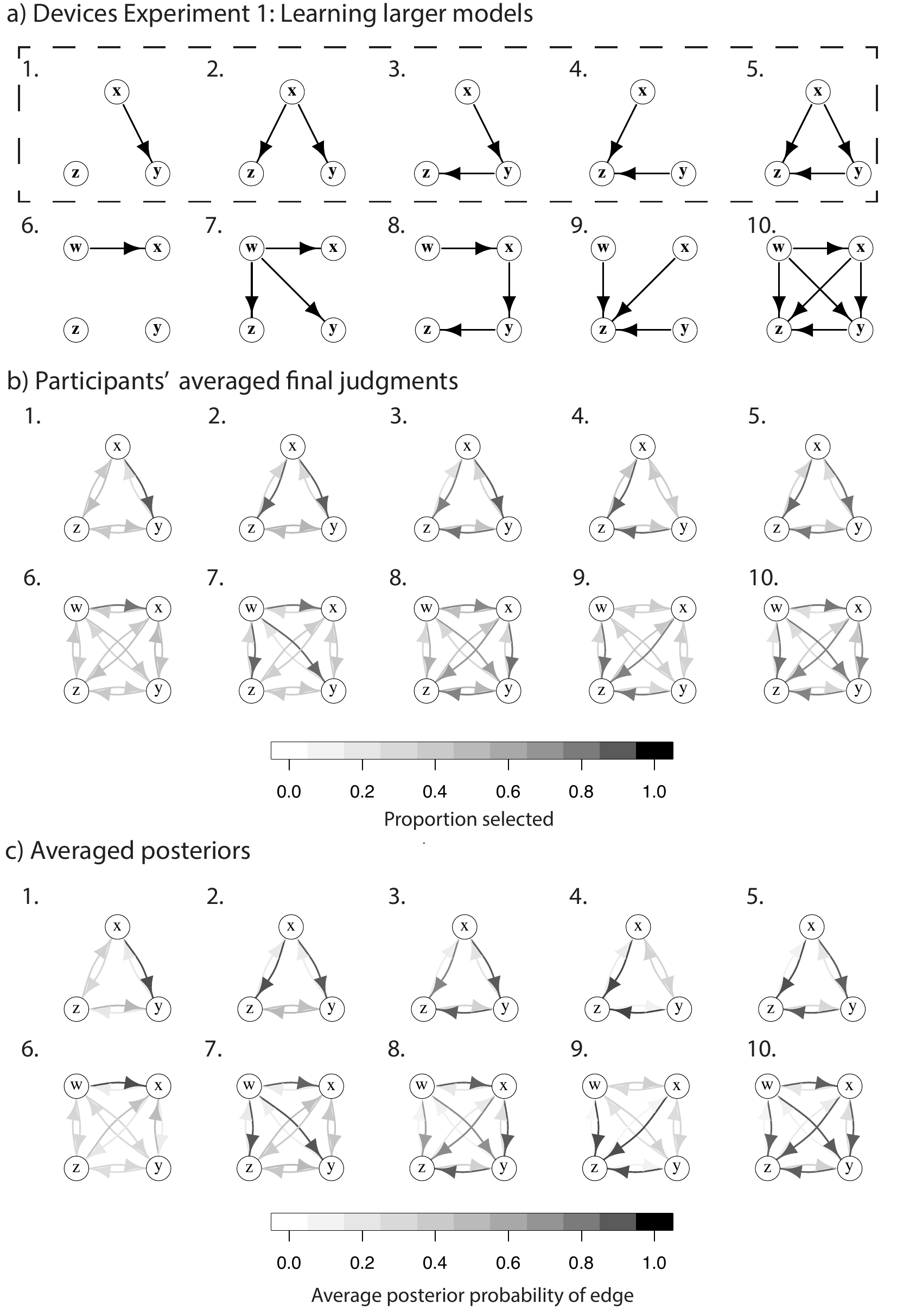}
   \caption{The true models from Experiment 1: Learning larger models, and visualisation of averaged judgments and posteriors.  a) The problems
     faced by participants.  Dashed box indicates those that also appeared in \cite{bramley2015fcs}.   b) Averaged final judgments by
     participants.  Darker arrows indicate that a larger proportion of
     participants marked this edge in their final model.  c)
     Bayes-optimal final marginal probability of each edge in
     $P(M|D^T,E^T,\ww)$, averaged over participants' data.}
   \label{fig:weighted_judgments_e1}
\end{figure}

\begin{figure}[t]
  \centering
   \includegraphics[width = .8\columnwidth]{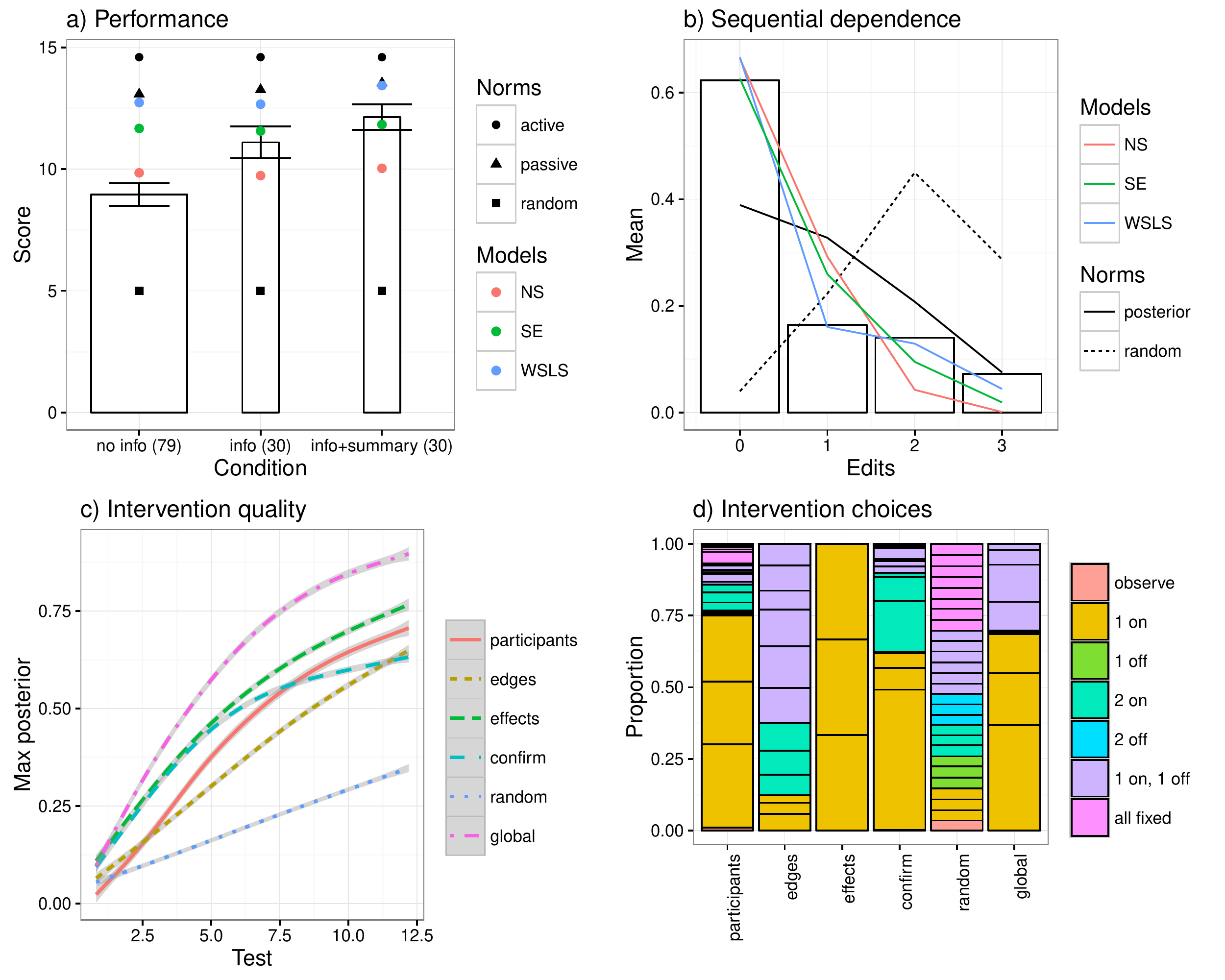}
   \caption{\cite{bramley2015fcs}; performance and interventions. a) Accuracy by condition.   Bars show participant accuracy by condition, and points compare with the models, bar widths visualize the number of participants per condition.  b) Sequential dependence.  The number of edits made by participants between successive judgments, bars give proportion of participants updates with different numbers of edits, lines compare with the models.  c) Quality of participants' and simulated learners'
     intervention choices measured by the probability that an ideal learner would
     guess the correct model given the information generated.  The plot
     shows values smoothed with R's {\tt gam} function and the gray
     regions give 99\% confidence intervals.   The proportion of
     interventions of different types chosen by participants as compared
     to simulated learners.  observe = $\Do[\varnothing]$, 1 on =
     e.g. $\Do[x\!=\!1]$, 1 off =  e.g. $\Do[x\!=\!0]$ and so on. All
     fixed = e.g. $\Do[x\!=\!0,y\!=\!1,z\!=\!0]$.}
   \label{fig:ns_fcs}
\end{figure}

\subsection*{Motivating the new experiments}

In analyzing  \cite{bramley2015fcs}, we found patterns of judgments and interventions broadly consistent with our framework.  However, the conclusions we can draw from this data alone are somewhat limited.  Firstly, the problems participants faced did not strongly delineate our \emph{Neurath's ship} proposal from other proposed approximations, namely the approximate \emph{win-stay, lose-sample} or the heuristic \emph{simple endorsement} which also predicted similar patterns of accuracy and sequential dependence. 

Similarly, in terms of interventions, participants' strong preference for ``one-on'' interventions was consistent with local effect-focused testing.  However, ``one-on'' interventions were also the globally most informative choices for the majority of participants' trials, especially early during learning.  Thus, we cannot be confident what participants focused on when selecting their interventions.

Methodologically also, several aspects of \cite{bramley2015fcs} are suboptimal for testing our framework.  Participants were allowed to leave edges unspecified until the the last test and could also draw cyclic models, both of which complicated our analyses.  Furthermore, participants had 12 tests on each problem, allowing an idealized learner to approach certainty given the high $w_S$ and low $w_B$, and for a significant minority of people to perform at ceiling.  These choices limit the incentive for participants to be efficient with their interventions. Additionally, participants were only incentivised to be accurate with their final judgment, meaning we cannot be confident that intermediate judgments always represented their best and latest guess about the model. Finally, participants were not forced to update all their edges after each test, meaning that lazy responding could be confused with genuine sequential dependence of beliefs.

Next, we report on two new experiments that build on the paradigm from \cite{bramley2015fcs}, making methodological improvements, while also exploring harder more revealing problems, and eliciting additional measures, all with the goal of better distinguishing our framework from competitors.

Experiment 1 explores learning in more complex problems than in \cite{bramley2015fcs}, with more variables and a range of strengths $w_S$ and levels of background noise $w_B$, and fewer interventions per problem.  The increased complexity and noise provides more space and stronger motivation for the use of approximations and heuristics.  Furthermore, the broader range of possible structures and intervention choices increases the discriminability of our framework from alternatives such as \emph{win-stay, lose-sample} and \emph{simple endorsement}, while the shorter problems avoid ceiling effects and ensure participants choose interventions carefully.  To ensure participants register their best and latest belief at every time point, we also incentivize participants through their accuracy at random time points during learning.  To eliminate the possibility of lazy responding biasing results in favor of \emph{Neurath's ship}, we force participants to mark all edges anew after every test without a record of their previous judgment as a guide.

Experiment 2 inherits the methodological improvements, compares two elicitation procedures, and also takes several additional steps.  In the previous studies, participants were pretrained on strength $w_S$ and background noise $w_B$. This will not generally be true; learners will normally have to take into account their uncertainty about these sources of noise during inference.  Therefore, Experiment 2 focuses on cases where participants are not pretrained on $\ww$.  Additionally, our framework makes predictions about participants' problem representation that go beyond how it should manifest in final structure judgments and intervention choices. Specifically, our local intervention schema proposes that people focus on subparts of the overall problem during learning, switching between these by comparing their current local uncertainty.  Experiment 2 probes these assumptions by asking learners for confidence judgments about the edges in the model during learning, and eliciting free explanations of what interventions are supposed to be testing.  When we go on to fit our framework to individuals in the final section of the paper, we are able to code up these free responses in terms of the hypotheses they refer to and compare them to the focuses predicted by our local uncertainty schema.

\section*{Experiment 1: Learning larger causal models}\label{section:exp_2}

Our first new experiment looks at learning in harder problems with a range of $w_S$ and $w_B$ and a mixture of 3- and 4-variable problems, asking whether we now see a clearer signature of \emph{Neurath's ship}, \emph{simple endorsement} or \emph{win-stay,lose-sample} style local updating or of local focus during interventions selection.

\subsection*{Methods}

\subsubsection*{Participants}
120 participants (68 male, mean$\pm$SD age $33\pm9$) were recruited from Amazon Mechanical Turk\footnote{Mechanical Turk (\url{http://www.mturk.com/}) is a web based platform for crowd-sourcing short tasks widely used in psychology research.  It offers an well validated \citep{hauser2015attentive,crump2013evaluating,buhrmester2011amazon,gosling2004should, mason2012conducting} subject pool, diverse in age and background, suitable for high-level cognition tasks.}, split randomly so that 30 performed in each of 4 conditions.  They were paid \$1.50 and received a bonus of 10c per correctly identified connection on a randomly chosen test for each problem ($\max= \$6.00$, mean$\pm$SD $\$3.7\pm 0.65$). The task took an average of $44\pm40$ minutes.

\begin{figure}[t]
   \centering
   \includegraphics[width = 0.8\columnwidth]{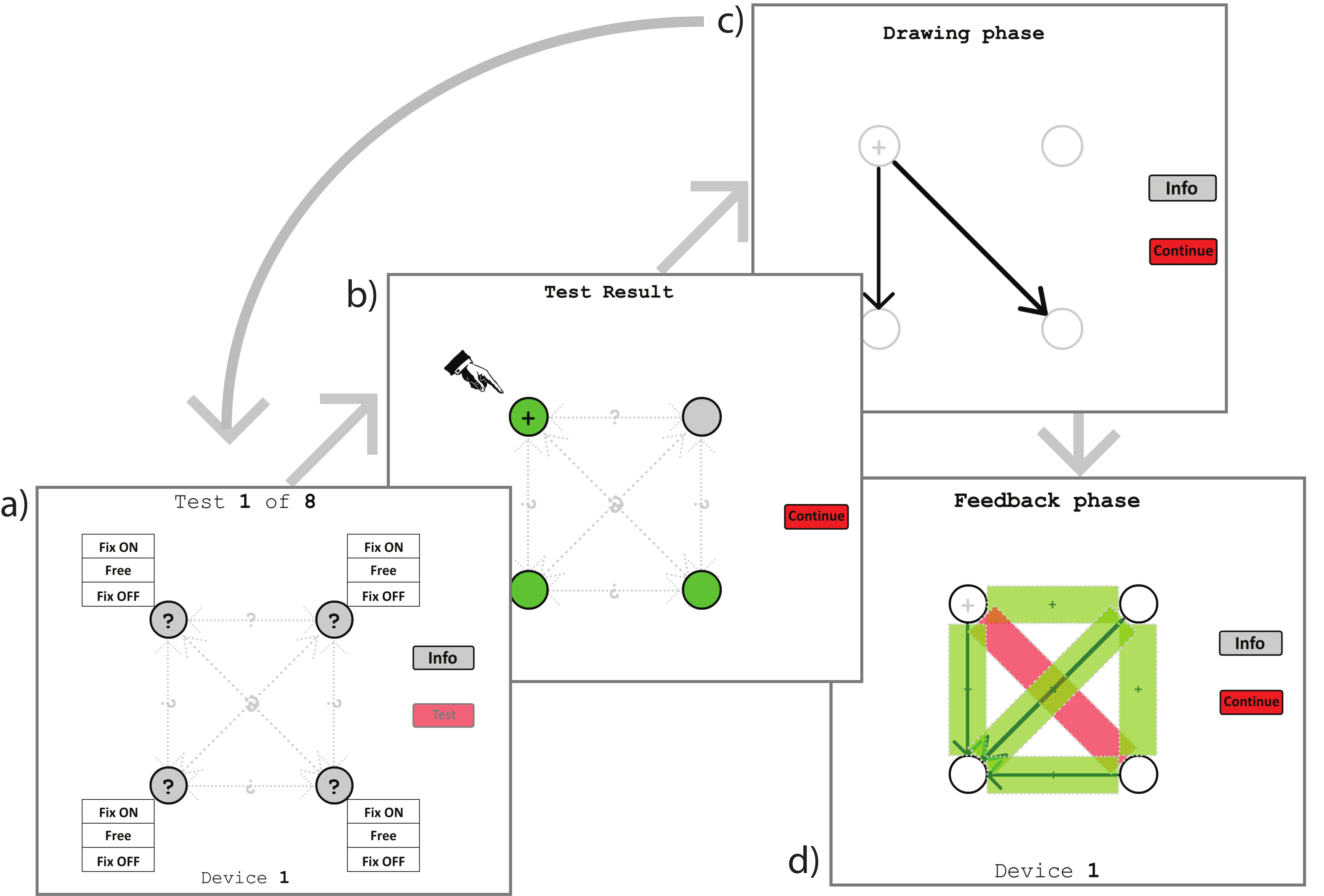}
   \caption{Experimental procedure. a) Selecting a test b) Observing the outcome c) Updating beliefs d) Getting feedback.}
   \label{fig:procedure_e1}
\end{figure}

\subsubsection*{Design}
This study included the five 3-variable problems in \cite{bramley2015fcs} plus five additional 4-variable problems (see Figure~\ref{fig:weighted_judgments_e1}a). There were problems exemplifying three key types of causal structure: forks (diverging connections), chains (sequential connections) and colliders (converging connections).  Within these, the sparseness of the causal connections varied between a single connection (devices 1 and 6) and fully connected (devices 5 and 10).

There were two different levels of causal strength $w_S \in [.9, 0.75]$ and two different levels of background noise $w_B\in [.1,.25]$ making $2\times 2 = 4$ between-subjects conditions. For instance, in condition $w_S=.9; w_B=.1$ the causal systems were relatively reliable, with nodes rarely activating without being intervened on, or caused by, an active parent, and connections rarely failing to cause their effects.  Meanwhile, in condition $w_S=0.75; w_B=0.25$ the outcomes were substantially noisier, with probability .25 that a variable with no active parent would activate, compared to a probability $1-(1-.75)(1-.25)=0.81$ for a variable with one active parent.

\subsubsection*{Procedure}

The task interface was similar to that in \cite{bramley2015fcs}.  Each device was represented as several gray circles on a white background (see Figure~\ref{fig:procedure_e1}). Participants were told that the circles were components of a causal system of binary variables, but were not given any further cover story. Initially, all components were inactive and no connection was marked between them.  Participants performed tests by clicking on the components, setting them at one of three states ``fixed on'', ``fixed off'' and ``free-to-vary'', then clicking ``test'' and observing what happened to the ``free-to-vary''components as a result. The observations were of temporary activity (graphically, activated components would turn green and wobble).

As in \cite{bramley2015fcs}, participants registered their best guess about the underlying structure after each test.  They did this by clicking between the components to select either no connection, or a forward or backward connection (represented as black arrows).  Participants were incentivised to be accurate, but unlike in \cite{bramley2015fcs}, payments were based on randomly selected time points rather than the final judgments.

Participants completed instructions familiarizing them with the task interface; the interpretation of arrows as (probabilistic) causal connections; the incentives for judgment accuracy.  To train $\ww$, participants were told explicitly and then shown unconnected components and forced to test them several times. The frequency with which the components activated reflected the true background noise level. They were then shown a set of two-component causal systems in which component ``$x$'' was a cause of ``$y$'', and were forced to test these systems several times with component $x$ fixed on. This indicated that the frequency with which $y$ activated reflected the level of $w_S$ combined with the background noise they had already learned.

After completing the instructions, participants had to answer four comprehension check questions.  If they got any wrong they had to go back to the start of the instructions and try again.  Then, participants solved a practice problem randomly drawn from the problem set. They then faced the test problems in random order, with randomly oriented unlabeled components. They performed six tests on each three variable problem, and eight tests on each four variable problem.  After the final test for each problem they received feedback telling them the true connections.

To ensure that participants' judgments were always genuine directed acyclic graphs, participants were told in the instructions that the true causal structure would not contain a loop.  Unlike in \cite{bramley2015fcs}, if participants tried to draw a model containing a cyclic structure they would see a message saying ``you have drawn connections that make a loop, change or remove one to continue''. 

As in \cite{bramley2015fcs} conditions \emph{information} and \emph{information + summary}, participants could hover their mouse over a button for a reminder of the key instructions during the task, but unlike condition \emph{information + summary}, they saw no record of their previous tests and outcomes.

The task can be tried out at \url{http://www.ucl.ac.uk/lagnado-lab/el/ns_sup}.

\subsection*{Results and discussion}

\subsubsection*{Judgments}
In spite of the considerably greater noise and complexity than \cite{bramley2015fcs}, participants performed significantly above chance in all four conditions (comparing to chance performance of $\frac{1}{3}$, participants scores differed significantly by t-test with $p<.001$ for all four conditions).  They also significantly underperformed a Bayes optimal observer ($p<.001$ for all four conditions, Figure~\ref{fig:ns_e1}a).  Performance declined as background noise $w_B$ increased $F(1,118)=4.3, \eta^2 = .04, p=.04$ but there was no evidence for a relationship with strength $w_S$  $F(1,118)=2.7, \eta^2 = .04, p=0.1$.  Judgment accuracy was no lower for four compared to three variable problems $t(238) = 0.76, p=0.44$.  Table~\ref{table:accuracy_by_device} shows accuracy by device type across all experiments.  Accuracy differed by device type $\chi^2=(4)=22, p<.001$.  Consistent with the idea that people struggle most to distinguish the chain from the fork or the fully connected model, accuracy was lowest for chains (devices 3; 8) and second lowest for fully connected (5; 10) models. 

In all four conditions, participants' final accuracy was closer to that of the \emph{Neurath's ship} simulations than the \emph{simple endorser}, \emph{win-stay, lose sample}  or \emph{random} responder or ideal (\emph{passive}) responding (Figure~\ref{fig:ns_e1}a).\footnote{On the rare occasions where the \emph{simple endorser} procedure would induce a cycle (0.4\% of trials), the edges were left in their original state.}

\begin{table}[ht]
\centering
\caption{Proportion of Edges Correctly Identified by Device Type in All Experiments}
\label{table:accuracy_by_device}
\footnotesize{
\begin{tabularx}{\columnwidth}{lXXXXXXX}
\toprule
Experiment & Variables & Single & Fork & Chain & Collider & Fully-connected & No connection \\
\hline
Bramley et al \citeyearpar{bramley2015fcs}          & 3         & 0.69   & 0.63 & 0.66  & 0.71     & 0.67            &                \\
Exp 1: Learning larger models         & 3         & 0.57   & 0.56 & 0.51  & 0.6      & 0.55            &                \\
Exp 1: Learning larger models         & 4         & 0.56   & 0.58 & 0.48  & 0.57     & 0.49            &                \\
Exp 2: Unknown strengths         & 3         & 0.62   & 0.65 & 0.61  & 0.6      & 0.61            & 0.61     \\
\hline
All        &           & 0.61 & 0.61 & 0.57 & 0.62 & 0.58            & 0.61 \\
\bottomrule    
\end{tabularx}}
\end{table}

\subsubsection*{Sequential dependence}

Table~\ref{table:sequential_dependence} summarizes the number of edits (additions, removals or reversals of edges) participants made between each judgment in all experiments.  Inspecting the table and Figure~\ref{fig:ns_e1}b we see participants judgments (both high and low performing) show a pattern of rapidly decreasing probability for larger edit distances mimicked by both \emph{Neurath's ship} and \emph{simple endorsement} simulations.  In contrast, \emph{random} or \emph{posterior} sampling lead to quite different signatures with larger jumps being more probable.  Choices simulated from \emph{Neurath's ship} and \emph{simple endorsement} were more sequentially dependent than participants' on average but have the expected decreasing shape.  \emph{Win-stay, lose sample} produces a different pattern with a maximum at zero changes but a second peak in the same location as for \emph{posterior sampling} but has an average edit distance very close to that averaged over participants.  However, we expect any random or inattentive responding to inflate average edit distances, and indeed find a strong negative correlation between edit distance and score $F(1, 118)=34, \beta = -6.7, \eta^2 = .34, p<.001$.  A simple way to illustrate this is to compare the edits of higher and lower performers.  Scores of $\frac{22}{45}$ or more differ significantly from chance performance (around $\frac{15}{45}$) by $\chi^2$ test.  The 79 participants that scored $22$ or more made markedly smaller edits than those that scored under 22 ($0.85 \pm 0.95$ compared to $1.3 \pm 1.12$ for three variable, and $1.4 \pm 1.5$ compared to $2.4 \pm 1.8$ for four variable problems), putting the clearly successful participants patterns closer to the ``Neurath's ship'' and ``simple endorser'' simulations.  Additionally, we expect individual differences in search length $\lambda$ under the \emph{Neurath's ship} model and here only simulate assuming a mean search length of 1.5.  Aggregating over a wider set of simulated learners with different capacities to search for updates would lead to a heavier-tailed distribution of edit distances that would resemble the participants' choices more faithfully.  %

\begin{table}[]
\centering
\caption{Edit Distance Between Consecutive Judgments in All Experiments.}
\label{table:sequential_dependence}
\footnotesize{
\begin{tabularx}{\columnwidth}{lXXXXXXXXXXXXXX}
\toprule
 Experiment                          & Var  & Participants &      & Random &    & SE     &     & WSLS &      &  NS   &      & Posterior &       \\
                                     &      & M            & SD   & M      & SD & M      & SD  & M    & SD   &  M    & SD   & M        & SD    \\
\hline
Bramley et al \citeyearpar{bramley2015fcs} & 3  & 0.66 & 0.97 & 1.99 & 0.83 & 0.51 & 0.74 & 0.55 & 0.88 & 0.38 & 0.57 & 1.02 & 0.93 \\  
Exp 1: Learning larger models              & 3  & 0.92 & 1.01 & 1.99 & 0.83 & 0.44 & 0.64 & 0.73 & 0.98 & 0.50 & 0.63 & 1.47 & 0.94 \\  
Exp 1: Learning larger models              & 4  & 1.69 & 1.63 & 3.96 & 1.16 & 0.65 & 0.88 & 1.64 & 1.78 & 0.55 & 0.68 & 2.94 & 1.40 \\  
Exp 2: Unknown strengths (remain)          & 3  & 0.73 & 0.85 & 2.02 & 0.81 & 0.43 & 0.62 & 1.06 & 1.13 & 0.59 & 0.69 & 1.86 & 0.85 \\  
Exp 2: Unknown strengths (disappear)       & 3  & 1.02 & 0.99 & 2.02 & 0.81 & 0.43 & 0.62 & 1.06 & 1.13 & 0.59 & 0.69 & 1.86 & 0.85 \\  
\hline
All                                        &    & 0.99 & 1.09 & 2.40 & 0.89 & 0.50 & 0.71 & 0.99 & 1.17 & 0.51 & 0.65 & 1.80 & 1.00 \\ 
\bottomrule
\end{tabularx}
}
NOTE: Var = number of variables.  NS = \emph{Neurath's ship} simulations with $\lambda=1.5$ and $\omega = 10$, WSLS = \emph{Win-stay, lose-sample} simulations, SW = \emph{Simple Endorser} simulations, M=mean, SD = standard deviation.
\end{table}

\subsubsection*{Interventions}
Globally focused active learning favored a mixture of ``one-on'' and ``one-on, one-off'' interventions (and several others including ``one-on, two-off'' in the four variable problems).  The number and nature of the fixed variables it favored depended strongly on the condition, favoring fixing more variables off when $w_S$ was high.   It would also shift dramatically over trials always favoring ``one-on'' interventions for the first trials but these dropping below 50\% of choices by the final test.  Participants' choices were much less reactive to condition or trial.  There were no clear differences in intervention choices by condition (see supplementary figures available at \url{http://www.ucl.ac.uk/lagnado-lab/el/ns_sup}) but participants were a little more likely to select ``one on'' interventions on their first test 57\% compared to their last 50\% test $t(238) = 1.7, p=.01$.  Like the \emph{ideal} or the \emph{effects} focused simulations, and like in \cite{bramley2015fcs}, learners favored ``one-on'' tests.  However,  in line with an \emph{edge} or \emph{confirmation} they also selected a substantial number of ``two-on'' and ``one-on, one-off'' interventions, doing so on early as well as late tests while the ideal learner only predicted using ``one-on, one-off'' tests on the last few trials.  As in \cite{bramley2015fcs} and consistent with \emph{confirmatory} testing, participants were more likely to fix ``on'' components with at least one child according to their latest hypothesis $b^{t-1}$:  49\% compared to 30\% $t(238) = 5.5, p<.001$.    The overall pattern was not clearly consistent with any one local focus but might be consistent with a mixture of all three.

\begin{figure}[t]
  \centering
   \includegraphics[width = \columnwidth]{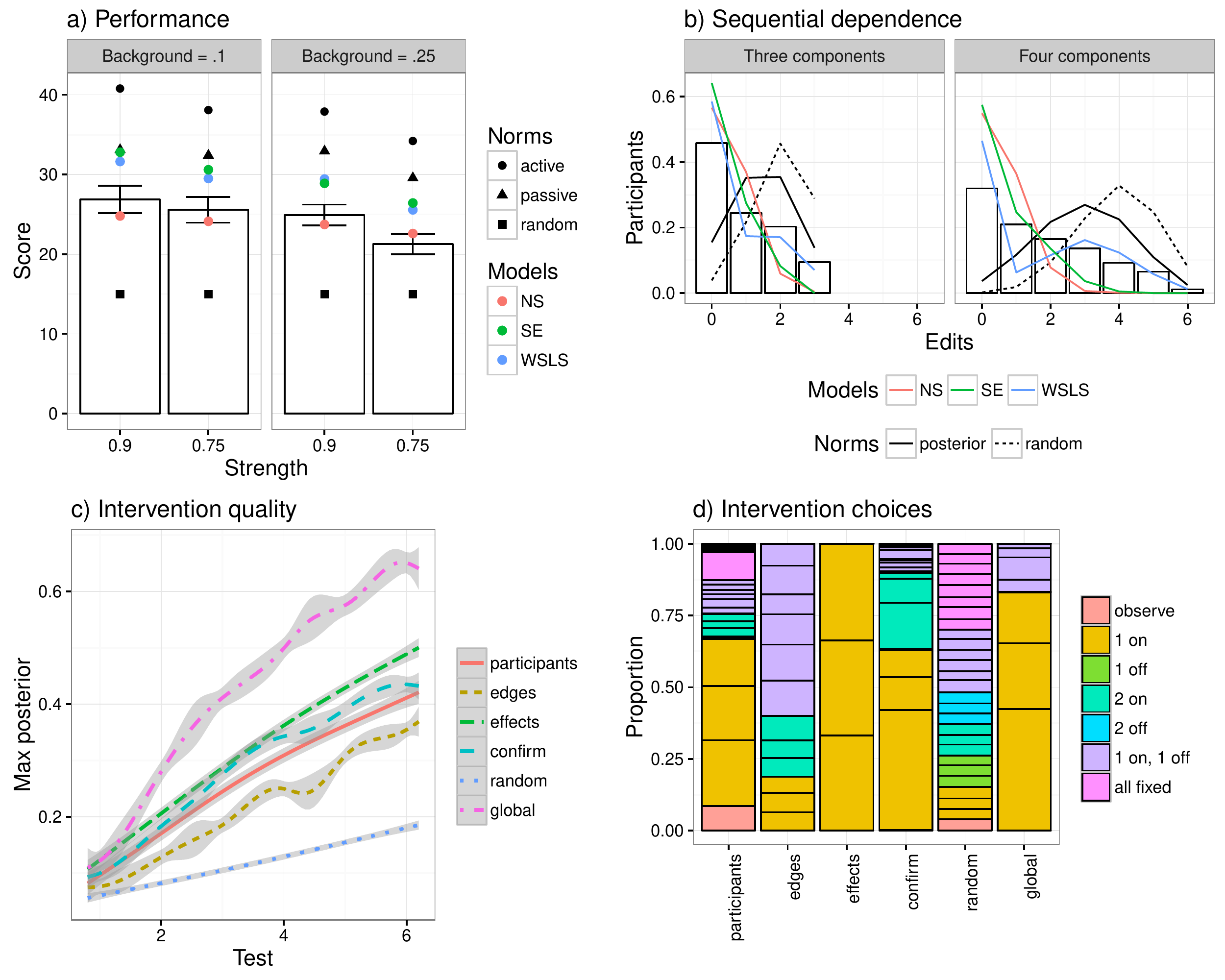}
   \caption{Experiment 1: Learning larger causal models; performance and interventions. a) Accuracy by condition.   Bars are participants and points compare with the models.  b) Sequential dependence.  Bars show the number of edits made by participants between successive judgments.  Lines compare with the models. c)  Quality of participants' and simulated learners' intervention choices in the three variable problems as in Figure~\ref{fig:ns_fcs}b.  d) The proportion of interventions of different types chosen by participants compared to simulated learners in the three variable problems, otherwise as in Figure~\ref{fig:ns_fcs}d.}
   \label{fig:ns_e1}
\end{figure}

\section*{Experiment 2: Unknown
  strengths}\label{section:exp_4} 

In this experiment, we focused on cases where participants are not pretrained on $\ww$ (see Appendix~\ref{app:a} for the computational level details of how to incorporate uncertainty over $\ww$ in model inference and intervention choice).\footnote{Experiments 2 is also reported briefly in \cite{bramley2015staying} but without discussion of the intervention choices or current model.}

We took advantage of the fact that participants would experience
substantially greater uncertainty given ignorance about $\ww$ to assess
their ability to estimate local uncertainty based on recently observed
data $\cald_r^t$ in order to choose where to focus subsequent tests. This is
central to any scheme for intervention selection.  Thus, in Experiment
3, we elicited the participants' confidence about the edges in each
judgment.  If participants track local uncertainties based on recent
evidence, we should expect these to correlate with uncertainties given
$\cald_r^t$.  In particular, given the representation associated with
\emph{Neurath's ship}, we might also expect the local confidences to be evaluated while leaning on the rest of the model for support.  This means they should reflect conditional uncertainty in the edge $H(E_{ij}|E_{\setminus ij}, \cald_r^t;\calc_r^t)$ more closely than the marginal uncertainty $H(E_{ij}|\cald_r^t;\calc_r^t)$ which involves averaging across all the possible states of the other edges.

We also elicited predictions about the outcome of each chosen test
before the outcome was revealed.  If participants maintained only a
single hypothesis, we expected this to be reflected in their predictions.  Thus, for a \emph{Neurath's ship} learner, it would be predominantly the
predictive distribution under their current hypothesis rather than the
average across models. 

Finally, in \cite{bramley2015fcs} and Experiment 1, participants' intervention selections
showed hints of being motivated by a mixture of local aspects of the
overall uncertainty, with overall patterns most consistent with focus on
a mixture of different local aspects of uncertainty.  To test this idea
more thoroughly, in the final problem in Experiment 2 we
explicitly probed participants' beliefs about their intervention choices through eliciting free responses which we go on to code and compare to our model predictions.

\subsection*{Methods}

\subsubsection*{Participants}

111 UCL undergraduates (mean $\pm$ SD age $18.7\pm0.9$, 22 male) took
part in Experiment 2 as part of a course. They were incentivised to be
accurate based on randomly selected trials as before, but this time with
the opportunity to win \emph{Amazon}$^{\text{\small \texttrademark}}$ vouchers
rather than money.  Participants were split randomly into 8 groups of
mean size $13.8\pm3.4$, each of which was presented with a different
condition in terms of the value of $\ww$ and the way that they had to
register their responses.

\subsubsection*{Design and procedure}

Experiment 2 used the same task interface as the other experiments, but focused just on the three variable problems (devices 1-5) and an additional device (6) in which none of the components was connected (Figure~\ref{fig:weighted_judgments_e2}). Like in Experiment 1, there were two causal strength conditions $w_S\in[0.9,0.75]$ and two background noise conditions $w_B\in[0.1,0.25]$. However, unlike Experiment 1, participants were not trained on these parameters, but only told that: \emph{``the connections do not always work''}, and \emph{``sometimes components can activate by chance''}.

To assess the extent the different reporting conditions drove lower sequential dependence in \cite{bramley2015fcs} relative to Experiment 1, we examined two reporting conditions between subjects: \emph{remain} and \emph{disappear}. In the \emph{remain} condition, judgments stayed on the screen into the next test, so participants did not have to change anything if they wanted to register the same judgment at $t$ as at $t-1$. In the \emph{disappear} condition, the previous judgment disappeared as soon as participants entered a new test. They then had explicitly to make a choice for every connection after each test.

In addition to the structure judgments and interventions, we also elicited additional probability measures from participants.  First, after selecting a test, but before seeing the outcome, participants were asked to predict what would happen to the variables they had left free.  To do this they would set a slider for each variable they had left free to vary.  The left pole of the slider was labeled ``Sure off'', the right pole ``Sure on''and the middle setting indicated maximal uncertainty (Figure~\ref{fig:procedure_extras_e2}a). Second, after drawing their best guess about the causal model by setting each edge between the variables, participants were asked how sure they were about each edge.  Again they would respond by setting a slider, this time between ``Guess'' on the left indicating maximal uncertainty and ``Sure'' on the right indicating high confidence that edge judgment was correct (Figure~\ref{fig:procedure_extras_e2}b).  Participants were trained and tested on interpretation of the slider extremes and midpoints in an additional interactive page during the instructions.

Participants faced the six devices in random order, with six tests per device followed by feedback as in Experiments 1 and 2.  Then they faced one additional test problem.  On this problem, the true structure was always a chain (Figure~\ref{fig:weighted_judgments_e2}, device 7). On this final problem, participants did not have to set sliders.  Instead, after they selected each test, but before seeing its outcome, they were asked why they had selected that intervention.  Labels would appear on the nodes and participants were invited to \emph{``Explain why you chose this combination of fixed and unfixed components.  Use labels `A' `B', `C' to talk about particular components or connections''} in a text box that would appear below the device.  Responses were constrained to be at least 5 characters long.  The chain (device 3) was chosen for this problem because in \cite{bramley2015fcs} and Experiments 1 and 2, participants often did not select the crucial $\Do[x\!=1,y\!=0]$ intervention that would allow them to distinguish a chain from a fully connected model (device 5) making this an interesting case for exploring divergence between participants' behavior and ideal active learning.

Finally, at the end of the experiment participants were asked to estimate the reliability $w_S$ of the true connections: \emph{``In your opinion, how reliable were the devices?  i.e. How frequently would fixing a cause component ON make the effect component turn ON too?''} and the level of background noise $w_B$: \emph{``In your opinion, how frequently did components activate by themselves (when they were not fixed by you, or caused by any of the device's other components)?''} by setting sliders between \emph{``0\% (never)''} and \emph{``100\% (always)''}.

A demo of Experiment 2 can be viewed at \url{http://www.ucl.ac.uk/lagnado-lab/el/ns_sup}.

\begin{figure}[t]
   \centering
   \includegraphics[width = 0.5\columnwidth]{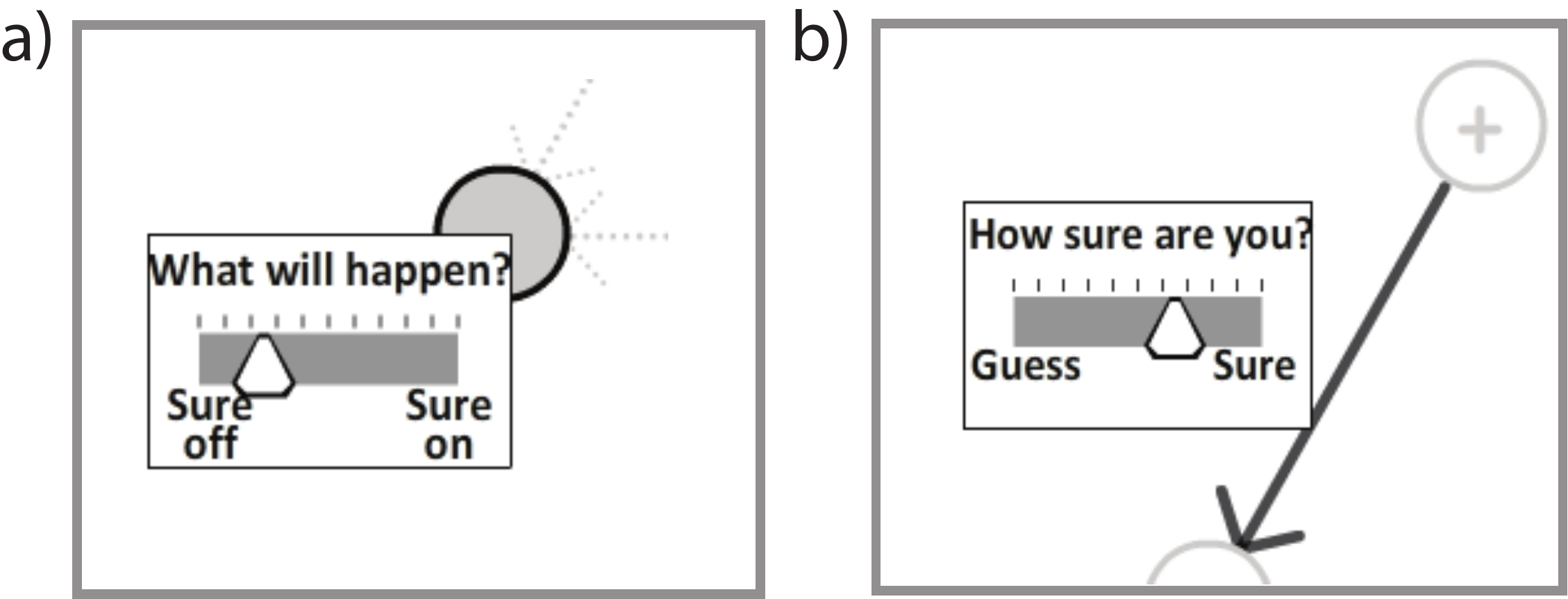}
   \caption{Exp 2: Unknown strengths;  additional measures - a) Outcome expectation sliders b) Edge confidence sliders.}
   \label{fig:procedure_extras_e2}
\end{figure}

\begin{figure}[t]
   \centering
   \includegraphics[width = 0.8\columnwidth]{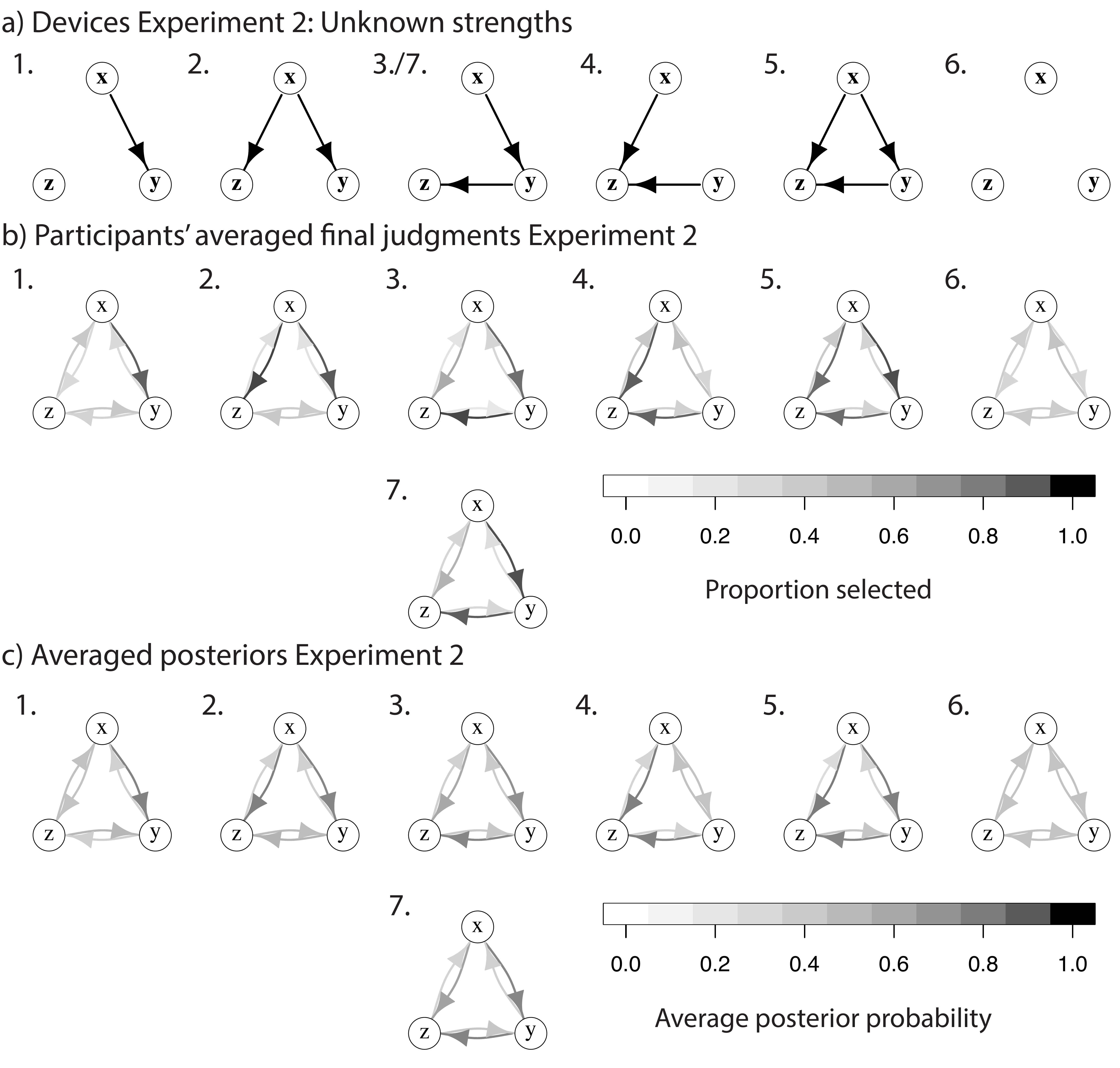}
   \caption{Experiment 2: Unknown strengths; true models and final judgments. a) The true models faced by participants. b) Weighted average
     final judgments by participants. Darker arrows indicate that a
     larger proportion of participants marked this link in their final
     model.  Note that problem 7 was the a repeat of the chain (problem 3) with the write aloud protocol. c) Bayes-optimal marginal probability of each edge in
     $\int_{\ww}P(M|D^T;C^T)p(\ww)~\mathrm{d} \ww$
     averaged over participants' data assuming a uniform independent
     prior over $\ww\in[0,1]^2$.}
   \label{fig:weighted_judgments_e2}
\end{figure}

\subsection*{Results and discussion}

\subsubsection*{Judgments}

As in the experiments where participants were trained on $\ww$, accuracy was significantly higher than chance in all conditions (all 8 t statistics $> 6.1$ all $p$ values $<0.001$) and underperformed a Bayes optimal observer observing the same data as them. Because the noise was unspecified, we explored several reasonable priors on $\ww$ (always assuming that $w_S$ and $w_B$ were independent) when computing posteriors.  Firstly, we considered a \emph{uniform-uniform} prior that made no assumptions about either $w_S$ or $w_B$ (UU) where $\ww\sim$ Uniform$(0,1)^2$.  We also considered a \emph{strong-uniform} (SU) variant, following \citep{yeung2011estimating}, expecting causes to be reliable -- $w_S \sim$ Beta$(2,10)$, but making no assumptions about background noise -- $w_B \sim$ Uniform$(0,1)$.  Additionally, we considered a \emph{sparse-strong} (SS) variant following Lu et al \citeyearpar{lu2008bayesian}, encoding an expectation of high edge reliability -- $w_S \sim $ Beta$(2,10)$, and relatively little background noise -- $w_B \sim$ Beta$(10,2)$.  The choice of parameter prior made little difference to the Bayes optimal observer's judgment accuracy.
Thus participants significantly underperformed the Bayes optimal observer in all conditions regardless of the assumed prior, except for condition $w_S=0.75; w_B=0.1$, \emph{remain}) under the SU prior, and $w_S=0.75; w_B=0.25$, \emph{remain} under all three considered priors.

\begin{figure}[t]
   \centering
   \includegraphics[width = \columnwidth]{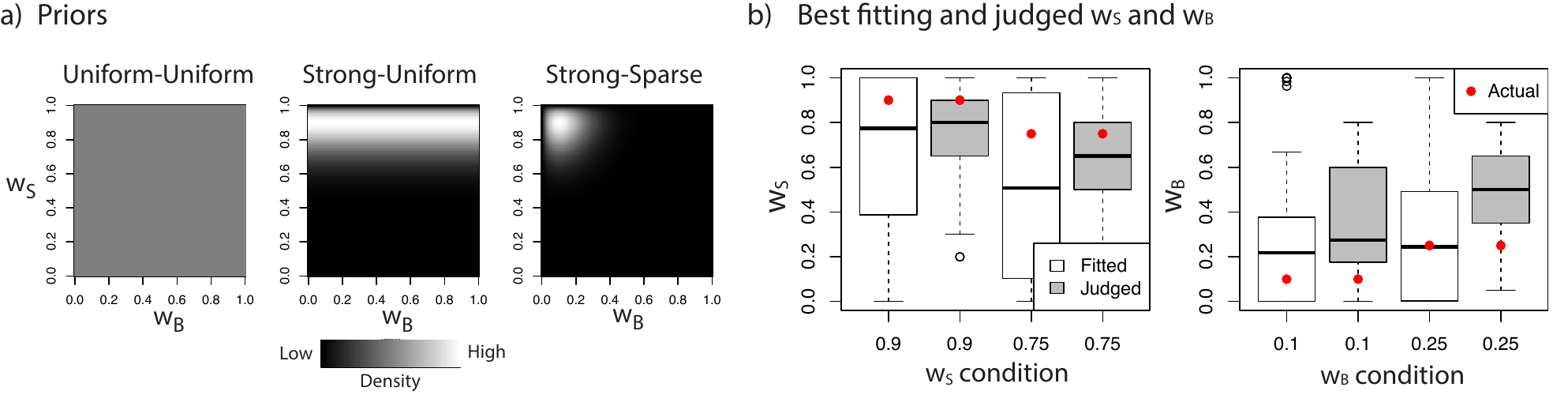}
   \caption{Experiment 2: Unknown strengths; priors and strength estimates. a) Visualizations of Uniform-Uniform,
     Strong-Uniform and Strong-Sparse priors on $w_S$ and $w_B$ b)
     Participants final judgments of amount of background noise ($w_B$)
     and strength ($w_S$), rescaled from 100 point scale to 0-1, and
     best-fitting $w_S^*$ and $w_B^*$ estimates assuming ideal Bayesian
     updating. Boxplots show the medians and
     quartiles.}
   \label{fig:performance_e2}
\end{figure}

\begin{figure}[t]
  \centering
   \includegraphics[width = .8\columnwidth]{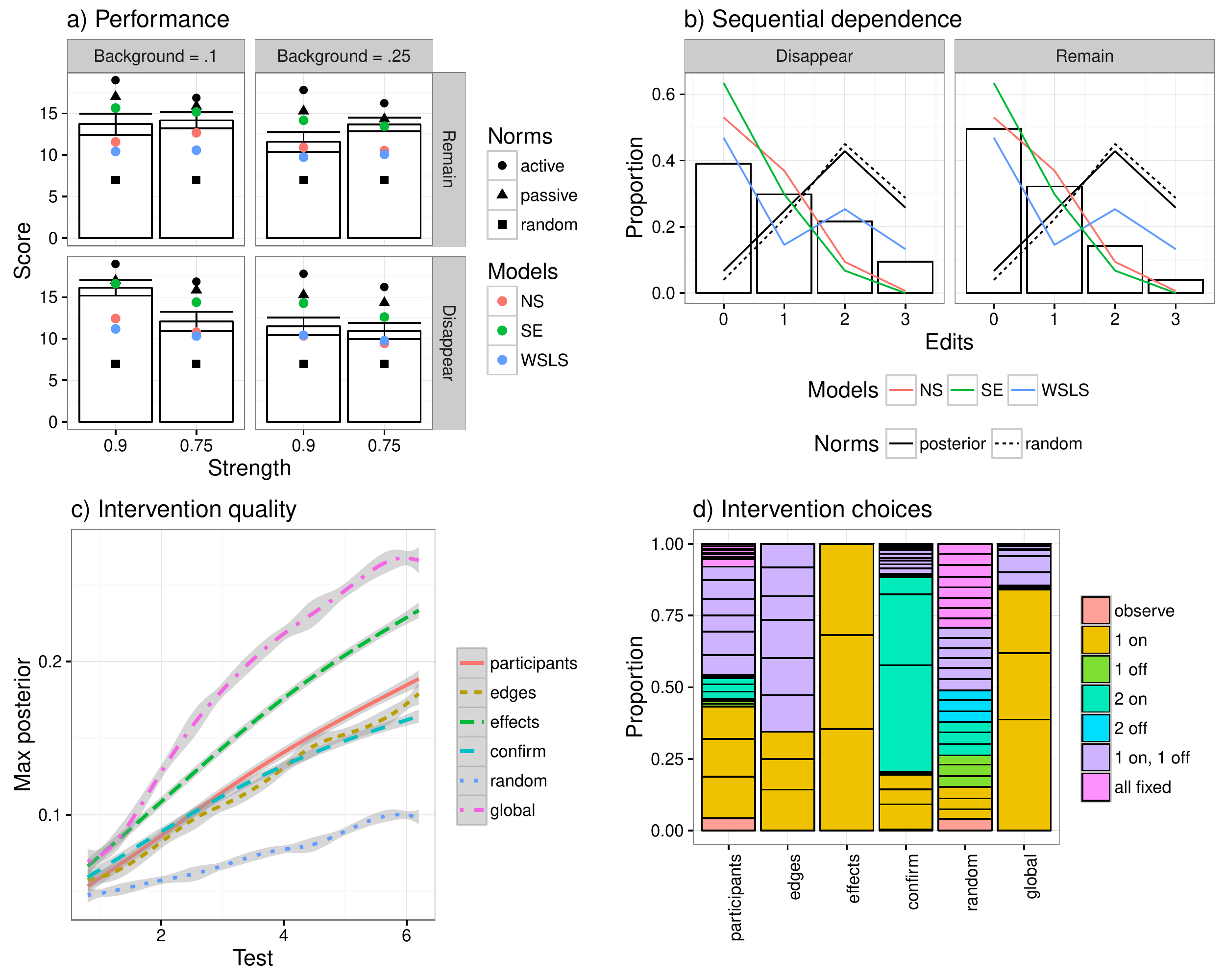}
   \caption{Experiment 2: Unknown strengths; performance and interventions.  Subplots as in Figure~\ref{fig:ns_e1}.}
   \label{fig:ns_e2}
\end{figure}

\subsubsection*{Comparison with known strength experiments}

Performance in Experiment 2 was comparable to the 3-variable problems in 
Experiment 1 where the underlying $\ww$ conditions were identical.  Mean accuracy was actually slightly higher
$0.61 \pm 0.21$ compared to $0.56 \pm 0.21$ for the matched problems in
Experiment 1, although not significantly so $t(229)=1.9, p=.054$. This suggests that participants were able make reasonable structure
judgments without knowledge of the exact parameters.  We found that participants' final judgments of $w_S$ and
$w_B$ and best fitting estimates assuming rational updating $w^*_S$ and
$w^*_B$ suffered bias and variance (Figure~\ref{fig:performance_e2} b).\footnote{Fifty-eight participants' final $w_B$ judgments were incorrectly stored, so the N for $w_B$
  judgments was 53 rather than 111.}

As with Experiments 1 and 2, participants were not affected by the reliability
of the connections themselves $w_S$ $t(106)=0.88, p=0.37$ but were
affected by higher levels of background noise $w_B$ $t(108)=2.7,
p=0.008$.  There was no difference in performance between the two
judgment elicitation conditions $t(108)=0.67, p=0.50$.

Participants were no more or less accurate on the final problem when identifying
a chain structure for the second time (device 7).  %
The most frequent error once again was mistaking the chain structure for the fully connected structure, made by 17/111 participants, although this was reduced to 11/111 when facing the chain structure again on device 7, with only a single participant making the same error twice.

Average edit distance between sequential judgments about the same device was significantly increased by removing the record of previous judgments between trials, going from .73 in the \emph{remain} condition to 1.0 in the \emph{disappear} condition $t(109)=3.5, p<.001$.  Edit distances even in the \emph{disappear} condition were still significantly lower than those predicted by UU, SU or SS posterior or random sampling (all p's $<.001$).  As in Experiment 1 there was a strong negative relationship between number of edits and performance $F(1,109)=102, \beta = 6.4, \eta^2=.48, p<.001$.  The edit-distance--performance relationship interacted weakly with condition $t(108)=1.9, \beta=1.3, p=.049$ becoming stronger in the \emph{disappear} condition. Again, the 71 participants who scored significantly above chance ($\frac{12}{21}$ or higher by $\chi^2$ test) had lower edit distances of $0.66 \pm 0.29$ than the remaining 40 participants' $1.3 \pm 0.44$.

\subsubsection*{Additional measures}

Participants' edge confidence judgments increased significantly over trials $\chi^2(1) = 2060, \beta=.04, SE = .0008, p<.001$, going from $.57\pm .20$ on the first trial to $.78\pm .19$ by the final trial.
  The probability of changing an edge at the next time point was weakly inversely related to the learners' reported confidence in it $\chi^2(1) = 67, \beta = -.03, SE=.004, p<.001$.
Reported edge confidences were correlated with both the conditional probability of the edge states given the the rest of the current model  $r^{cond}=$.20 and the \emph{marginal} probability of the edge-state in the full posterior under the UU prior $r^{mar}=$.17 but these correlations did not differ significantly.

As predicted, reported outcome predictions were more closely related to the predictive distribution under the participants' latest structure judgment $b^{t-1}$: $\chi^2(1) = 1044, \beta=.35, SE=.010, p<.001$ than marginalized over the full posterior  $\chi^2(1) = 580, \beta=.29, SE=.012, p<.001$.  The latest-structure to prediction relationship was significantly stronger than the marginal posterior to prediction relationship by Cox test $Z=10.9, p<.001$.  

\subsubsection*{Interventions}

The overall distribution of intervention choices was broadly similar to the other Experiments (Figure~\ref{fig:ns_e2}).  ``One-on'' interventions were the most frequently chosen, making up $39\%$ of selections.  However, unlike the previous Experiments, and consistent with \emph{edge} focused learning, the constrained ``one-on one-off'' interventions were almost as common as single ``one-on'' interventions, making up $38\%$ of tests compared to $12\%$ across 3-variable problems in Experiment 1.  The intervention selections and informativeness of intervention sequences were not closely consistent with global expected information, nor any single type of local focus, but could again be consistent with a mixture of local \emph{effect} focused, \emph{edge} focused and \emph{confirmation} focused queries.

\subsubsection*{Free explanations}\label{section:frc}

For device 7, participants gave free explanations for their
intervention choices on each of their six tests.  The overall
distribution of intervention choices did not differ significantly from
the original presentation of the chain (device 3) $\chi^2 = 31, p=0.21$
suggesting that the different response format did not affect the
intervention choices that participants made.  In order to assess what
the explanations tell us about participants' intervention choices, we
asked two independent coders to categorize the free responses into 8
categories.  The categories were chosen in a partly data-driven, partly
hypothesis-driven way: 1. An initial set of categories were selected,
with the goal of distinguishing the approximations introduced in
\emph{A local uncertainty schema} from global strategies like
uncertainty sampling or expected information maximization.  2. A subset
of the data was then checked and the categories were refined to better
delineate their responses with minimal membership ambiguity.

The eight resulting categories were:

\begin{enumerate}
\item The participant just wanted to learn about one specific
  connection. [Corresponding to \emph{edge focused} testing]
\item The participant wanted to learn about two specific connections.
\item The participant wanted to learn about all three connections. [Corresponding to \emph{globally focused} testing]
\item The participant wanted to learn what a particular component can
  affect but did not mention a specific pattern of
  connections. [Corresponding to \emph{effect focused} testing]
\item The participant wanted to test / check / confirm their current hypothesis. [Corresponding to \emph{confirmatory} testing]
\item The participant wanted to learn about the randomness in the system
  (as opposed to the location of the connections). [Corresponding to a focus on learning about noise rather than structure]
\item The participant chose randomly / by mistake / to use up unwanted
  tests / they say they did not understand what they are doing /it is
  clear they were not engaging with the task. 
\item The participant's explanation was complex / underspecified / did
  not seem to fall in any of the above categories.
\end{enumerate}

A supplementary file (available at \url{http://www.ucl.ac.uk/lagnado-lab/el/ns_sup}) contains all the materials given to coders and the full set of participant responses. Coders were permitted to assign more than one category per response, but had to select a primary category.  When the category referred to particular component label(s), the rater would record these, and when it referred to a specific connection they would record which direction (if specified) and the components involved.  These details will be used to facilitate a quantitative comparison between participants' explanations and our model fits in the next Section. Raters normally just selected one category per response, selecting additional categories on only $8\%$ of trials. Inter-rater agreement on the primary category was 0.73, and Cohen's $\kappa= 0.64\pm 0.04$, both higher than their respective heuristic criteria for adequacy of 0.7 and 0.6 \citep{landis1977measurement,krippendorff2012content}.

Figure~\ref{fig:coding_summary_time}, shows the proportion of responses in the different categories across the six trials.  On the first trial participants were most likely to be categorized as 4. -- focused on identifying what a particular variable could effect.  On subsequent trials they most frequently categorized as 1. -- focusing on learning about a specific connection.  Toward the end, explanations became more diverse and were increasingly categorized as 5. -- confirmatory testing or 6. learning about the noise in the system.  Individuals almost always gave a range of different explanations across their six tests, falling under $3.0\pm 0.99$ different categories on average, with only 5/111 participants providing explanations from the same category all six times (3 all-fours, 1 all-threes. and 1 all-eights).

Explanation type was predictive of performance $F(8,657) = 13.75, \eta^2
= 0.14, p<0.001$.  Taking category 7 -- unprincipled or random
intervening -- as the reference category with low average performance of
10.2 points out of a possible 21, categories 1,2,4,5, and 6 were all
associated significantly higher final scores $[14.5, 12.9, 13.9, 13.9,
13.9]$ points, all p's $<0.001$. 

\begin{figure}[t]
   \centering
   \includegraphics[width = \columnwidth]{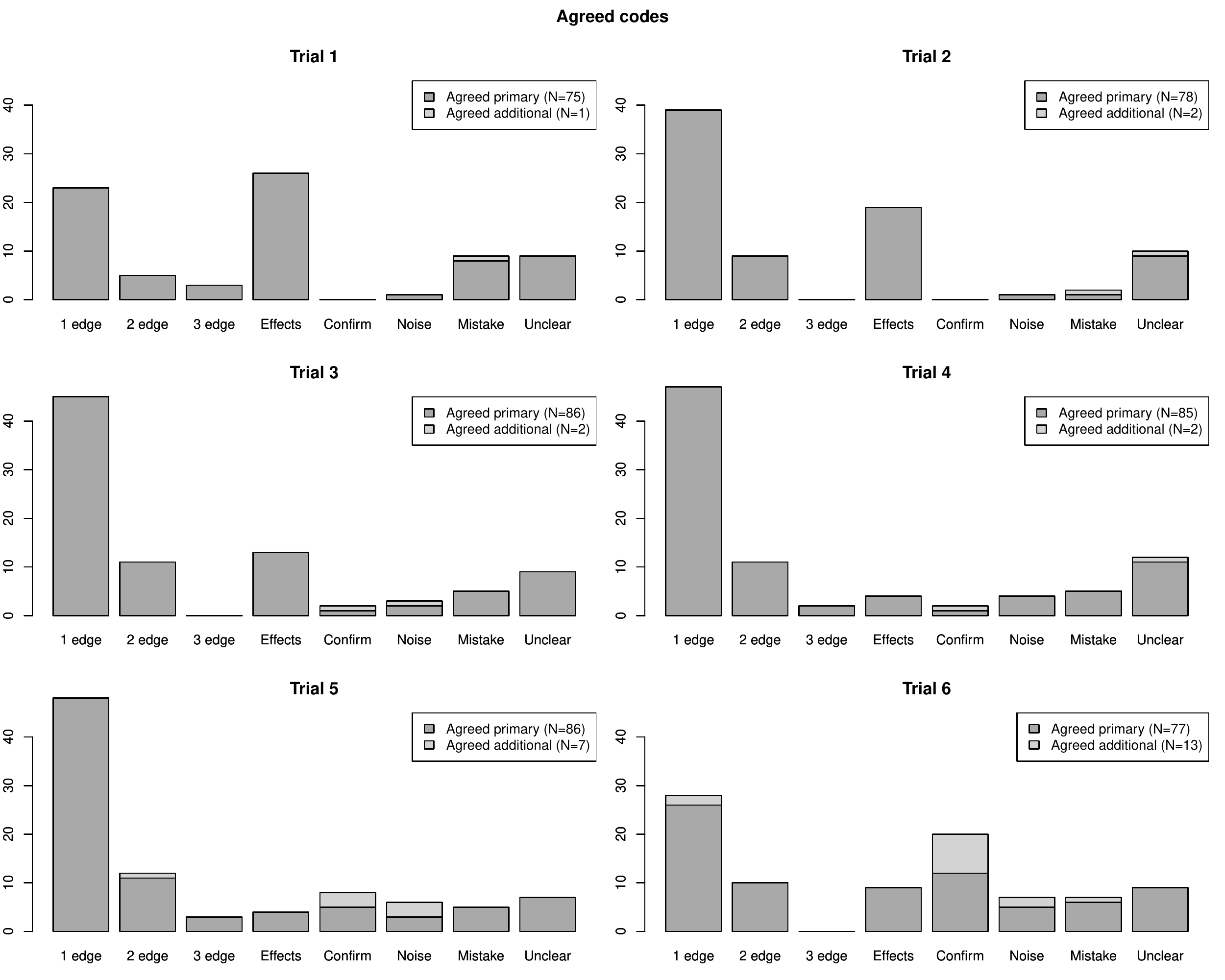}
   \caption{Experiment 2: Unknown strengths; free explanations for interventions agreed codes over the six tests in problem 7.}
   \label{fig:coding_summary_time}
\end{figure}

\subsection*{Summary of Experiments}\label{section:exps_discussion}

In all these experiments, participants were clearly able to generate plausible causal models but also did so suboptimally.  Averaged across participants, final model judgments resembled the posterior over models (e.g. Figures~\ref{fig:weighted_judgments_e1}c and \ref{fig:weighted_judgments_e2}c), however individuals' trajectories typically exhibited strong sequential dependence, with the probability of moving to a new model decreasing with its edit distance from the previous model.  This is consistent with our hypothesis that individuals normally maintain a single hypothesis and update it piece by piece.  As found in previous research, participants were worst at separating the direct and indirect causes in the chain (3; 8) and fully-connected (5; 10) models.  A closer look at participants' intervention choices suggests that this was due to a common failure to generate the constrained interventions, such as $\Do[x\!=\!1, y\!=\!0]$, necessary to disambiguate these options.  The \emph{simple endorser} model predicts this error by proposing that people ignore the dependencies between the different edges.  Our framework provides a more nuanced explanation.  Whether a learner will correctly disambiguate these options depends on whether they focus on $x-z$ before or after having inferred $x\rightarrow y$ and $y\rightarrow z$.   If the consider $x-z$ after, then they will tend to fix $y$ ``off'',  realizing it is necessary to prevent the indirect path from confounding the outcome of their test.  However, if they have no connection marked from $x$ to $y$ or from $y$ to $z$, they will not expect this confounding activation and so have no motivation to fix $y$ ``off'' when testing $x-z$.

Participants' overall distributions of intervention selections resembled a mixture of \emph{edge}, \emph{effect} and \emph{confirmation} focused testing, but their distributions of choices were relatively invariant across conditions and trials while the efficient learners' were much more dynamic.  Comparison with the final global information gathered revealed that they did not select which variables to target particularly efficiently, leading to a considerable discrepancy between the total information gathered by participants compared to an ideal active learner.  However, participants also displayed hints of adaptation of strategy over the trials: with a preference for confirmatory testing, being more likely to fix variables ``on'' when they had children according to their latest hypothesis $b^{t-1}$, and displaying a modest shift toward more constrained interventions in later trials.

In Experiment 2 we saw that people were able to identify causal structure effectively without specific parameter knowledge.  Comparing a range of plausible prior assumptions about edge reliability $w_S$ and the level of background noise $w_B$ yielded little difference in judgment or intervention choice predictions.  Participants' overall judgment accuracy was not affected by the
\emph{remain/disappear} reporting condition, but this did affect
sequential dependence, especially for lower performers who may have often forgotten their previous judgment when making their next one.  The idea, common to the three judgment rules we consider, that people represent one model at a time was also supported by the additional
measures elicited from participants during the task.  
With a single hypothesis rather than distributional beliefs, intervention outcome predictions could only be generated by the current hypothesis rather than averaged and weighted over all possible models.  Consistent with this idea, we found participants' expectation judgments were more in line with their current hypothesis than the marginal likelihoods, although we note that these measures were quite noisy and the effects quite small.

\section*{Modeling individual behavior}\label{section:model}

Across all three examined experiments we found a qualitative correspondence, both between our \emph{Neurath's ship} simulations and participants' judgments, and between the two stage local intervention schema and participants interventions.  However, both \emph{simple endorsement} and \emph{win-stay, lose-sample} also appeared to do a good job of capturing qualitative  judgment patterns.  In order to validate quantitatively which of these models better describes participants' behavior, we fit the models to the data and assessed their competence relative also to \emph{win-stay, lose-sample} and \emph{simple endorsement}.  By fitting the models separately to individual participants we also assessed individual differences in learning behavior, and thus gained a finer-grained picture of the processes involved.

\subsection*{Judgments}

\subsubsection*{Models}

We compared six models to participants judgments, the three process models we considered in the experiment \emph{Neurath's ship} (NS), \emph{simple endorser} (SE), \emph{win-stay,lose-sample} (WSLS), alonside an efficient Bayesian learner (\emph{Rational}) and two null models \emph{Baseline} and NS-RE.

For NS, we fit three parameters:
\begin{enumerate}
  \item An average search length parameter $\lambda$ controlling the probability of searching for different lengths $k$ on each belief update.
  \item A search behavior parameter $\omega$ controlling how strongly the learner moves toward the more likely state for an edge when updating it (recalling that $\omega=1$ leads to probability matching, while $\omega=\infty$ leads to deterministic hill climbing and $\omega=0$ to making random local edits).
  \item A lapse parameter $\epsilon$ controlling a mixture between the model predictions and a uniform distribution.
\end{enumerate}
Including the last parameter into equation~\ref{eq:model_gibbs}, this
resulted in the following equation
\begin{equation}
P(b^t=m|\cald_r^t,\calc_r^t,\ww, b^{t-1}, \omega,
\lambda)=(1-\epsilon){\sum\nolimits_{0}^\infty} 
\frac{\lambda^k e^{-\lambda}}{k!} [(R^\omega_t)^k]_{b^{t-1}m}
+\epsilon\mathrm{Unif}(M) 
\end{equation}
where $R$ is a Markov matrix expressing the options for local
improvement.

We operationalized the \emph{Simple endorser} (SE) \citep{bramley2015fcs} with two parameters. One is the probability $\rho\in [0,1]$ with which the belief state is updated from $b^{t-1}$ include extra edges from any currently fixed ``on'' node(s) to any activated nodes and to exclude edges from any currently fixed ``on'' node(s) to any non-activated nodes (we write $b^{t-1}_{+SE}$).  With the complementary probability $1-\rho$, it stays the same as $b^{t-1}$.  As with the NS model we also included a lapse parameter mixing in a probability of choosing something at random, giving

\begin{equation}
P(b^t=m|\dd,\ww) = (1-\epsilon)(\rho~b^{t-1}_{+SE} + (1-\rho)b^{t-1}) + \epsilon Unif(M)
\end{equation}

 \emph{Win-stay, lose-sample} (WSLS) \citep{bonawitz2014win} predicts that participants stick with their current model $b^{t-1}$ with probability $p(\dd^t|b^{t-1},\ww, c^t)$ or else draw a sample from the full posterior with probability $1-p(\dd^t|b^{t-1},\ww, c^t)$.  The fitted version of this model had a single lapse parameter $\epsilon$ giving

\begin{equation}
P(b^t=m|D^t,\ww) = (1-\epsilon) \Big( (1-P(\dd^t|b^{t-1},\ww, c^t)) P(M|D^t,\ww)_t + P(\dd^t|b^{t-1},\ww, c^t)[m=b^{t-1}] \Big) + {\epsilon} \mathrm{Unif}(M)
\end{equation}

The final model, \emph{Rational} was a variant of the Bayes-optimal observer (Section~2) that attempted to select the maximum a posteriori causal structure $\max P(M|D^t,\ww;C^t)$ with each judgment, with a soft maximization \citep{luce1959choice} governed by inverse temperature parameter $\theta$ and a lapse parameter $\epsilon$. For this, we considered

\begin{equation}
P(b^t=m|D^t,\ww) = (1-\epsilon)\frac{\exp(P(M|D^t,\ww)_t \theta)}{\sum_{m'\in M} \exp(P(m'|D^t,\ww)_t \theta)} + {\epsilon} \mathrm{Unif}(M)
\end{equation}

\emph{Baseline} is a parameter-free baseline that
assumes each judgment to be  a random draw from all possible causal models
\begin{equation}
p(b^t = m) = \mathrm{Unif}(M)
\end{equation}
(leading to a probability of approximately $\frac{1}{3}$ for each edge).

One concern with this baseline is that judgments might exhibit sequential dependence yet be unrelated to data $\cald_r^t$.  Therefore we also considered a baseline variant of the NS model in which the search behavior parameter $\omega$ was fixed to 0, resulting in a (R)andom (E)dit model (NS-RE) that walks randomly around the hypothesis space for $k$ steps on each update.  For
this model, small $k$ simply denotes more inertia.

Each of these belief models output a likelihood based on the probability that the model assigns to a belief of $b^t$, given the most recent outcome $\dd^t$ (SE), outcomes since the last belief change $\cald_r^t$ (NS), or all outcomes $D^t$ (WSLS, \emph{Rational}), and the most recent judgment $b^{t-1}$.  Because the choice of prior for Experiment 2 made negligible difference to our results, we only report models assuming uniform (UU) priors on $\ww$.  For Experiment 2, we also marginalized over the unknown values of $\ww$ rather than conditioning as in the other experiments as detailed in Appendix~\ref{app:b}.

\subsubsection*{Evaluation}

To compare these models quantitatively, we used maximum likelihood optimization as implemented by R's {\tt optim} function to fit the model separately to each of the 370 participants across all three experiments.\footnote{In Appendix~\ref{app:b} we provide additional detail on how the models were fit.}  We used Bayesian Information Criterion \citep[BIC,][]{schwarz1978estimating} to compare the models while accommodating their differing numbers of parameters. \emph{Baseline} acts as the null model for computing BICs and pseudo-$R^2$s \citep{dobson2010introduction} for the other models. In \cite{bramley2015fcs} participants were not forced to select something for each edge immediately, although once they did so they could not return to ``unspecified'', and they could also respond with cyclic causal model if they wanted.  Therefore, we fit only the 75\% of tests where the participants report a fully specified non-cyclic belief, taking the $b^{t-1}$ to be the unconnected model on the first fully specified judgment, as we do with $b^{0}$ in the other Experiments.  Recalculating the transition probabilities on the fly in the optimization of $\omega$ was infeasibly computationally intensive for the four-variable problems.  So for Experiment 1 we first fit all three parameters to the three-variable problems only, then used the best fitting $\omega$ parameters from this fit when fitting the $\lambda$ and $\epsilon$ on the full data.  In \cite{bramley2015fcs} and Experiment 2 we were able to fit all three parameters.

\subsubsection*{Results and discussion}

Table~\ref{table:belief_mods} details the results of the model fits to all experiments.  Summed across all participants, NS has the lowest total BIC (93381) with the SE in second place (94326), followed by WSLS with (97643), then NS-RE (101837), \emph{Rational} (1207209) and finally \emph{Baseline} with (149313).  NS was also the best fitting model for \cite{bramley2015fcs} and Experiment 1, with SE winning in Experiment 2.  Thus, all three heuristics substantially beat an exact Bayesian inference account of causal judgment here, but \emph{Neurath's ship}, with its ability to capture a graded dependence on prior beliefs, outperformed WSLS substantially, and the heuristic SE to a lesser degree.  In terms of number of individuals best fit,  Table~\ref{table:belief_mods} shows a broad spread across models: WSLS -- 102, NS -- 85, SE -- 80, NS-RE -- 70, \emph{Rational} -- 28 , \emph{Baseline} -- 4.

\begin{table}[ht]
\caption{Belief Model Fits.}
\centering
\footnotesize{
\begin{tabularx}{\columnwidth}{lXXXXXXXXXXrrrXXXr}
\toprule
\multicolumn{11}{l}{\cite{bramley2015fcs}}\\
\hline
  Model &  $\lambda$ M & $\lambda$ SD & $\omega$ M & $\omega$ SD & $\theta$ M & $\theta$ SD &$\rho$ M & $\rho$ SD & $\epsilon$ M & $\epsilon$ SD &  N fit & M acc & $\log$L & $R^2$ & BIC \\ 
  \hline
  Baseline &  &  &  &  &  &  &  &  &  &  & 1 & 0.27 & -17836 & 0 & 35672 \\ 
      NS-RE & 0.17 & 51.5 &  &  &  &  &  &  & 0.21 & 0.19 & 32 & 0.36 & -9379 & 0.49 & 19762 \\ 
  SE &  &  &  &  &  &  & 0.18 & 0.18 & 0.21 & 0.24 & 13 & 0.60 & -8819 & 0.53 & 18642 \\ 
  WSLS &  &  &  &  &  &  &  &  & 0.14 & 0.28 & 56 & 0.85 & -9117 & 0.52 & 18736 \\ 

  \textbf{NS} & \textbf{1.20} & \textbf{100.1} & \textbf{6} & \textbf{243.8} &  &  &  &  & \textbf{0.05} & \textbf{0.20} & \textbf{27} & \textbf{0.66} & \textbf{-8197} & \textbf{0.56} & \textbf{17901} \\ 
  Rational &  &  &  &  & 5 & 124.8 &  &  & 0.00 & 0.39 & 10 & 0.93 & -12089 & 0.36 & 25182 \\
  \hline
  \multicolumn{11}{l}{Exp 1: Learning larger models}\\
   \hline
   Model &  $\lambda$ M & $\lambda$ SD & $\omega$ M & $\omega$ SD & $\theta$ M & $\theta$ SD &$\rho$ M & $\rho$ SD & $\epsilon$ M & $\epsilon$ SD &  N fit & M acc & $\log$L & $R^2$ & BIC \\ 
  \hline
   Baseline &  &  &  &  &  &  &  &  &  &  & 1 & 0.33 & -41814 & 0 & 83628 \\ 
        NS-RE & 1.23 & 1.5 &  &  &  &  &  &  & 0.25 & 0.28 & 33 & 0.55 & -29235 & 0.30 & 59490 \\ 
   SE &  &  &  &  &  &  & 0.51 & 0.32 & 0.56 & 0.29 & 30 & 0.49 & -27736 & 0.34 & 56492 \\ 
   WSLS &  &  &  &  &  &  &  &  & 0.32 & 0.34 & 29 & 0.52 & -28772 & 0.31 & 58053 \\ 

  \textbf{NS} & \textbf{1.63} & \textbf{1.7} & \textbf{4} & \textbf{241.0} &  &  &  &  & \textbf{0.20} & \textbf{0.29} & \textbf{27} & \textbf{0.56} & \textbf{-27234} & \textbf{0.35} & \textbf{55896} \\ 
   Rational &  &  &  &  & 13 & 143.2 &  &  & 0.14 & 0.36 & 0 &  & -36362 & 0.13 & 73743 \\ 
  \hline
   \multicolumn{11}{l}{Exp 2: Unknown strengths}\\
  \hline
   Model &  $\lambda$ M & $\lambda$ SD & $\omega$ M & $\omega$ SD & $\theta$ M & $\theta$ SD &$\rho$ M & $\rho$ SD & $\epsilon$ M & $\epsilon$ SD & \specialcell[t]{N fit\\rem/diss} & M acc & $\log$L & $R^2$ & BIC rem/diss \\ 
   \hline
   Baseline &  &  &  &  &  &  &  &  &  &  &  0/2 & 0.21 & -15006 & 0 & 14330/15682 \\ 
      NS-RE & 0.9 & 3 &  &  &  &  &  &  & 0.10 & 0.29 & 3/2 & 0.36 & -10877 & 0.28 & 10050/12535 \\ 
   \textbf{SE} &  &  &  &  &  &  & \textbf{0.80} & \textbf{0.30} & \textbf{0.43} & \textbf{0.27} & \textbf{21/16} & \textbf{0.75} & \textbf{-9181} & \textbf{0.39} & \textbf{8257~/10936} \\ 
   WSLS &  &  &  &  &  &  &  &  & 0.00 & 0.31 & 9/8 & 0.50 & -10220 & 0.32 & 9402/11452 \\ 

   NS & 1.8 & 121 & 9.2 & 247 &  &  &  &  & 0.03 & 0.20 & 13/18 & 0.55 & -9170 & 0.39 & 8620/10964 \\ 
   Rational &  &  &  &  & 31.5 & 282 &  &  & 0.00 & 0.26 & 7/12 & 0.70 & -10482 & 0.30 & 10116/11678 \\
  \bottomrule
\end{tabularx}
}
Note: Columns: M = median estimated parameters across all participants, SD = standard deviation of parameter estimate across all participants, N fit = number of participants best fit by each model, M acc = average proportion of edges identified correctly by participants best fit by this model, LogL = total log likelihood of model over all participants, $R^2$ = median McFadden's pseudo-$R^2$ across all participants, BIC = aggregate Bayesian information criterion across all participants.  For Exp 2, rem = remain condition, diss =disappear condition.  Best fitting model denoted with boldface.\raggedright
\label{table:belief_mods}
\end{table}

The diversity of individual fits across strategies raises the question of the identifiability of the different models.  To assess how reliably genuine followers of the different proposed strategies would be identified by our modeling procedure, we simulated participants using the fitted parameters for each model for each of the actual participants in all three examined experiments.  We then fit all six models to these simulated participants report the rates at which simulations are best-captured by each model.  Table~\ref{table:model_recovery} provides the complete results for this recovery analysis.  Overall, the generating model was recovered 74\% of the time for \cite{bramley2015fcs}, 82\% for Experiment 1 and 75\% for Experiment 2 (chance would be 17\%).  In all three experiments, data generated by \emph{Baseline, WSLS} and SE was nearly always correctly recaptured, indicating that we can treat cases where participants are well described by these models as genuine.   Additionally NS almost never captured data generated by any of the other models, providing reassurance that NS is not simply fitting participants who are doing something more in line with SE or WSLS.  However, data actually generated by NS was frequently recaptured by the NS-RE (random edit) null model that makes NS-style local edits
but does not preferentially approach more likely models.  This was true in the majority of cases in \cite{bramley2015fcs} and Experiment 2.  Some of the cases where NS-RE captures NS-generated simulations are based on participants who were better described by NS-RE in the first place (e.g. whose search behavior was too random to justify  $\omega$'s inclusion).  We find a similar effect whereby simulated \emph{Rational} participants with relatively low $\theta$s or high $\epsilon$s are more parsimoniously described by \emph{Baseline}.  This is supported by looking at the more complex four variable problems in Experiment 1, NS simulations were identified the majority of the time, and when restricted to simulations based on parameters from participants who were actually best described by NS, 24/27 were recovered successfully.    Thus it is plausible that some of the 70 NR-RE participants were infact doing something more in line with NS.  There is a suggestion of this in Experiment 1, where the mean accuracy of the NS-RE participants is commensurate with SE, WSLS and NS.

The performance of a handful of participants -- 10 in \cite{bramley2015fcs}, and 19 in Experiment 2 --  were best fit by the \emph{Rational} model, which has one fewer parameter than NS. Naturally, these participants performed particularly well, scoring near ceiling in \cite{bramley2015fcs} (identifying 14.0 of the 15 connections) and as high as the ideal learning simulations in Experiment 2 -- 14.7/21 compared to an average of 15.5 for perfect Bayesian integration.  This, along with the lower recovery rates for these experiments, suggests that their design  -- both being motivated primarily to look closely at intervention choice -- may not have been difficult enough to separate the process from the normative predictions about the judgments.

Figures~\ref{fig:model_parameters}a and b show the range of the fitted $\lambda$ and $\omega$ parameters under NS.  In line with our predictions, participants' average fitted search lengths ($\lambda$) were mainly small, with medians between 1 and 2 in all three experiments.\footnote{A few participants made judgments that were sequentially anti-correlated leading to $\lambda$ parameters at the limit of the optimization routine's precision and correspondingly large standard deviations in Experiments 1 and 2.} Because this parameter merely encodes a participant's average search length this means that the same participant would sometimes not search at all, staying exactly where they are ($k=0$), or might also sometimes search much longer (e.g. $k\gg\lambda$).  The median fitted $\omega$s of 6, 4 and 9.2 across the three experiments are suggestive of moderate hill-climbing. A substantial number of participants had very large values of $\omega$ indicative of near-deterministic hill climbing.  We discuss this trade-off further in the General Discussion.  However, note that we were only able to fit these values to the easier three variable problems. It might be that the largest values would have been tempered if they could have been fit to the four variable problems as well.

\begin{table}[ht]
\caption{Belief Model Recovery Analysis}
\centering
\label{table:model_recovery}
\scriptsize{
\begin{tabularx}{\columnwidth}{lXXXXXXlXXXXXX}
\toprule
 \multicolumn{7}{l}{\textbf{\cite{bramley2015fcs}}}\\
 \multicolumn{7}{l}{\emph{All simulated participants}} &  \multicolumn{7}{l}{\emph{Best fit participants}}\\ 
  & Baseline & Rational & WSLS & SE & \mbox{NS-RE} & NS  & & Baseline & Rational & WSLS & SE & \mbox{NS-RE} & NS \\ 
  \hline
  Baseline & \textbf{13}4 & 0 & 4 & 0 & 1 & 0 &           Random & \textbf{1} & 0 & 0 & 0 & 0 & 0 \\ 
  Rational & 39 & \textbf{95} & 4 & 0 & 0 & 1 &             Rational & 0 & \textbf{10} & 0 & 0 & 0 & 0 \\ 
  WSLS & 5 & 0 & \textbf{133} & 1 & 0 & 0 &                 WSLS & 0 & 0 & \textbf{56} & 0 & 0 & 0 \\ 
  SE & 1 & 0 & 2 & \textbf{119} & 17 & 0 &                  SE & 0 & 0 & 0 & \textbf{13} & 2 & 0 \\ 
  NS-RE & 2 & 0 & 1 & 53 & \textbf{82} & 1 &                NS-RE & 0 & 0 & 1 & 9 & \textbf{20} & 0 \\ 
  NS & 3 & 0 & 3 & 11 & \textbf{72} & 50 &                  NS & 0 & 0 & 1 & 1 & \textbf{16} & 9 \\ 
  \hline
    \multicolumn{7}{l}{\textbf{Exp 1: Learning larger models}} \\
     \multicolumn{7}{l}{\emph{All simulated participants}} &  \multicolumn{7}{l}{\emph{Best fit participants}}\\
 & Baseline & Rational & WSLS & SE & \mbox{NS-RE} & NS  &  & Baseline & Rational & WSLS & SE & \mbox{NS-RE} & NS \\ 
  \hline
  Baseline & \textbf{114} & 1 & 3 & 0 & 2 & 0   &   Baseline & \textbf{1} & 0 & 0 & 0 & 0 & 0   \\ 
  Rational & 42 & \textbf{75} & 2 & 0 & 1 & 0 &   Rational & 0 & 0 & 0 & 0 & 0 & 0 \\ 
  WSLS & 4 & 1 & \textbf{115} & 0 & 0 & 0     &   WSLS & 1 & 1 & \textbf{27} & 0 & 0 & 0    \\ 
  SE & 6 & 0 & 9 & \textbf{94} & 11 & 0       &   SE & 0 & 0 & 1 & \textbf{31} & 0 & 0      \\ 
  NS-RE & 2 & 0 & 1 & 3 & \textbf{112} & 2    &   NS-RE & 0 & 0 & 0 & 1 & \textbf{29} & 1   \\ 
  NS & 3 & 0 & 3 & 2 & 30 & \textbf{82}       &   NS & 0 & 0 & 0 & 0 & 4 & \textbf{23}      \\ 
  \hline
       \multicolumn{7}{l}{\textbf{Exp 2: Unknown strengths}} \\
        \multicolumn{7}{l}{\emph{All simulated participants}} &  \multicolumn{7}{l}{\emph{Best fit participants}}\\
 & Baseline & Rational & WSLS & SE & \mbox{NS-RE} & NS  &  & Baseline & Rational & WSLS & SE & \mbox{NS-RE} & NS \\ 
  \hline
  Baseline & \textbf{107} & 1 & 3 & 0 & 0 & 0     &  Baseline & \textbf{2} & 0 & 0 & 0 & 0 & 0 \\ 
    Rational & 17 & \textbf{9}2 & 2 & 0 & 0 & 0   &    Rational & 1 & \textbf{18} & 0 & 0 & 0 & 0 \\ 
    WSLS & 7 & 0 & \textbf{100} & 2 & 2 & 0       &    WSLS & 1 & 0 & \textbf{16} & 0 & 1 & 0 \\ 
    SE & 9 & 0 & 6 & \textbf{93} & 3 & 0          &    SE & 0 & 0 & 0 & \textbf{30} & 0 & 0 \\ 
    NS-RE & 9 & 0 & 8 & 3 & \textbf{89} & 2       &    NS-RE & 0 & 0 & 0 & 0 & \textbf{5} & 0 \\ 
    NS & 5 & 2 & 9 & 2 & \textbf{70} & 23         &    NS & 0 & 2 & 3 & 0 & \textbf{22} & 10 \\ 
   \bottomrule
\end{tabularx}
}
Note: Rows denote simulation rule, and columns the model used to fit the simulated choices.  The number in each cell shows how many of the simulations using this rule were best fit by that model.  Right hand side restricts this to simulations using the parameters taken from participants who were actually best fit by each model.  Boldface denotes the model most frequently best fit in each case.\raggedright
\end{table}

\subsection*{Interventions}

\subsubsection*{Models}

We compared our local model of intervention choice (Section~4) to a
globally-focused and a
baseline model. 
Each intervention model output a likelihood for an intervention choice
of $\eee^t$, depending on $\cald_r^t,\calc_r^t$ and $b^{t-1}$.

We compared the overall distribution of participants' intervention
selections and final performance with \emph{edge} focused, \emph{effect}
focused and \emph{confirmation} focused tests.  We found that none of
these models alone closely resembled participants' response patterns, but overall
distributions were consistent with a mixture of different types of local
tests.  This was also supported by the the free-response coding in
Experiment 2, showing that participants would typically report targeting
a mixture of specific edges, effects of specific variables and
confirming the current hypothesis.  Therefore, we considered four
locally driven intervention selection models, one for each of the three
foci, plus a mixture. 

For the \emph{edge} model, the possible foci
$\call$ included the 3 (or 6) edges in the model.  For the
\emph{effect} model, it comprised the 3 components (or 4 in the
4-variable case).  The confirmation model always had the same focus --
comparing $b^{t}$ to null $b^0$ of no
connectivity.  The \emph{mixed} model contained all 7 (or 11) foci. As in Equations \ref{eq:stage1} and \ref{eq:stage2} in Section~4, each
model would first compute a soft-max probability of choosing each
possible focus $\lll^t\in\call$.  Within each chosen focus it
would also calculate the soft-max probability of selecting each
intervention, governed by another inverse temperature parameter $\eta\in
[0,\infty]$.  The total likelihood of the next intervention choice was
thus a soft-maximization-weighted average of choice probabilities across
possible focuses
\begin{equation}
P(\eee^t|\eta,\rho,\cald_r^t,b^{t-1},\ww) = \sum_{\lll\in L}P(\eee|\lll,\eta,b^{t-1},\ww) \frac{\exp (H(\lll|\cald_r^t,b^{t-1},\ww;\calc_r^t)\rho) } { \sum_{\lll'\in \call} \exp( H(\lll'|\cald_r^t,b^{t-1},\ww;\calc_r^t)\rho)} \label{eq:local_intervention_fitting_model}
\end{equation}
where
\begin{equation}
P(\eee|\lll,\eta,b^{t-1},\ww) =
\frac{\exp\left(\E_{{\dd}\in\mathcal{D}_{\eee^t}} \left[\Delta
  H(\lll|{\dd}, b^{t-1},\ww;\eee\right]\eta\right)}{\sum_{\eee'\in
\mathcal{C}}\exp\left(\E_{{\dd}\in\mathcal{D}_{\eee}} \left[\Delta
  H(\lll|{\dd}, b^{t-1},\ww;\eee'\right]\eta\right)} 
\end{equation}

Positive values of $\rho\in [-\infty,\infty]$ encode a preference for focusing on areas where the learner should be most uncertain, $\rho=0$ encodes random selection of local focus, and negative $\rho$ encodes a preference for focusing on areas where the learner should be most certain.

For comparison, \emph{Baseline} is a parameter-free model that assumed
each intervention was a random draw from all possible interventions
\begin{equation}
P(\eee^t) = \mathrm{Unif}(\mathcal{C})
\end{equation}
\emph{Global} is a variant of the globally efficient intervention
selection (Section~2) that attempted
to select the globally most informative greedy test $\arg\max_{\eee\in
  \mathcal{C}}\E_{{\dd} \in \mathcal{D}_{\eee}}\left[\Delta H(M|{\dd},
D^{t-1},\ww;C^{t-1},\eee)\right]$.  It has one inverse temperature parameter
$\theta\in[0,\infty]$ governing soft maximization \citep{luce1959choice}
over the global expected information gains. For this, we considered
\begin{equation}
P(\eee^t|D^{t-1},\ww;C^{t-1}) = \frac{\exp(\E_{\dd \in
    \mathcal{D}_{\eee^t}} \left[\Delta
  H(M|{\dd},D^{t-1},\ww;\eee^t)\right]\theta)}{\sum_{\eee\in
    \mathcal{C}}\exp(\E_{\dd \in \mathcal{D}_{\eee}} \left[\Delta
  H(M|{\dd},D^{t-1},\ww;\eee)\right]\theta)} 
\end{equation}

As with the belief modeling, for Experiment 2 we marginalized
over the the unknown values of $\ww$ rather than conditioning as in
Experiments 1-2 as detailed in Appendix~\ref{app:b}.%

\subsubsection*{Evaluation}

All six models were fit to the data from all three experiments in the
same way as the belief models.  The results are detailed in
Table~\ref{table:int_mods}. 

Additionally, to compare model predictions of local focus choice
$\lll^t$ to participants' self reports in problem $7$ in Experiment 2,
we computed the likelihood of each local focus prediction on each test.
This was done by calculating $P(\eee|\lll,\eta,b^{t-1},\ww)$ for each of
the local foci we considered, using a fixed common $\eta=20$ to capture strong but nondeterministic preference for the most useful intervention(s).
For each data point $c^t$, we then calculated which $l^t$ assigned the most probability to $c^t$ the intervention actually chosen by the participant.  Figure~\ref{fig:response_model_correspondence} plots the most likely focus of participants' intervention choices in the final problem against the code assigned to their free responses.

\subsubsection*{Results and discussion}

\begin{table}[ht]
\caption{Intervention Model Fits}
\centering
\footnotesize{
\begin{tabularx}{\columnwidth}{lXXXXXXXrrrr}
  \toprule
&  \multicolumn{7}{l}{\cite{bramley2015fcs}}\\
  \hline
Model &  $\eta$ M & $\eta$ SD & $\rho$ M & $\rho$ SD & $\theta$ M & $\theta$ SD & 
  N fit & M acc & $\log$L & $R^2$ & BIC \\
  \hline
Baseline &  &  &  &  &  &  & 3 & 0.38 & -18262 &  & 36524 \\ 
Edges & 10.4 & 26 & 0.9 & 80 &  &  & 9 & 0.68 & -14222 & 0.23 & 29338 \\ 
\textbf{Effects} & \textbf{7.7} & \textbf{16} & \textbf{0.5} & \textbf{5} &  &  & \textbf{82} & \textbf{0.71} & \textbf{-10701} & \textbf{0.41} & \textbf{22296} \\ 
Confirmatory & 3.9 & 74 &  &  &  &  & 9 & 0.43 & -15368 & 0.14 & 31182 \\ 
Mixed & 15.1 & 152 & 0.7 & 33 &  &  & 32 & 0.66 & -11145 & 0.39 & 23185 \\ 
Global &  &  &  &  & 6.1 & 4.1 & 4 & 0.85 & -15619 & 0.14 & 31686 \\ 
  \hline
\multicolumn{7}{l}{Exp 1: Learning larger models}\\
    \hline
Model &  $\eta$ M & $\eta$ SD & $\rho$ M & $\rho$ SD & $\theta$ M & $\theta$ SD & 
 N fit & M acc & $\log$L & $R^2$ & BIC \\
  \hline
Baseline &  &  &  &  &  &  & 18 & 0.35 & -32958 &  & 65917 \\ 
Edges & 9.3 & 138 & 25.9 & 1012 &  &  & 2 & 0.37 & -28588 & 0.13 & 58196 \\ 
Effects & 6.2 & 39 & 1.5 & 6 &  &  & 31 & 0.59 & -24213 & 0.27 & 49445 \\ 
Confirmatory & 3.8 & 8 &  &  &  &  & 24 & 0.49 & -28721 & 0.13 & 57951 \\ 
\textbf{Mixed} & \textbf{8.8} & \textbf{139} & \textbf{17.6} & \textbf{414} &  &  & \textbf{27} & \textbf{0.61} & \textbf{-23944} & \textbf{0.27} & \textbf{48907} \\ 
Global &  &  &  &  & 4.9 & 4 & 18 & 0.66 & -26652 & 0.19 & 53813 \\ 
  \hline
\multicolumn{7}{l}{Exp 2: Unknown strengths}\\
  \hline
Model &  $\eta$ M & $\eta$ SD & $\rho$ M & $\rho$ SD & $\theta$ M & $\theta$ SD & 
 N fit & M acc & $\log$L & $R^2$ & BIC \\
  \hline
Baseline &  &  &  &  &  &  & 14 & 0.35 & -15365 &  & 30730 \\ 
Edges & 4.0 & 8 & 2.7 & 77 &  &  & 24 & 0.74 & -13010 & 0.18 & 26850 \\ 
Effects & 3.2 & 9 & 2.9 & 218 &  &  & 7 & 0.52 & -12992 & 0.14 & 26815 \\ 
Confirmatory & 2.5 & 14 &  &  &  &  & 12 & 0.46 & -14180 & 0.04 & 28776 \\ 
\textbf{Mixed} & \textbf{3.9} & \textbf{9} & \textbf{5.4} & \textbf{285} &  &  & \textbf{15} & \textbf{0.63} & \textbf{-12550} & \textbf{0.17} & \textbf{25931} \\ 
Global &  &  &  &  & 3.0 & 8 & 39 & 0.70 & -12850 & 0.14 & 26114 \\ 
  \bottomrule

\end{tabularx}
}
Note: Columns as in belief model (Table~\ref{table:belief_mods}).\raggedright
\label{table:int_mods}
\end{table}

The \emph{mixed} local focus model was the best fitting model over the
three experiments with the lowest total BIC of 97757, followed by
\emph{effects} then by the \emph{global} focused model, then by
\emph{edges} and finally by \emph{confirmation} and then baseline.
However, there was a great deal of individual variation, suggesting that
a single model does not capture the population well.  More participants
were best described by an \emph{effects} focus (121)  than a \emph{mixed} focus (77),  but each model received some support, with 58, 43, 36 and 35 individuals best fit by \emph{global}, \emph{confirmation} and \emph{edge} focused and \emph{baseline} models respectively.  Additionally, the \emph{effect} focus was was the best fitting model overall in \cite{bramley2015fcs} where there was a strong tendency for participants to fix a single variable on at a time.

As Table~\ref{table:int_mods} shows, \emph{mixed} was the best overall
fitting model for Experiments 1 and 2, and the majority of
participants 277/370 were fit by one of the local uncertainty driven models.
Furthermore, Figure~\ref{fig:response_model_correspondence} shows that
for effect and edge queries, there was a strong correspondence between
the most likely choice of focus $l$ on Experiment 2
problem 7 and the coded explanation of that intervention's goal.  This
was not the case for tests where explanations were categorized as
\emph{confirmatory}.  These were most frequently best described as
effect focused tests of the root variable of the true model (labeled
``$x$'' in the plots).

As with the case of judgments, a moderate number of chance-level
performing participants (35/370) were best described by the
\emph{Baseline} model.  However, 58 participants across the three
experiments were better described by the \emph{Global}ly efficient
testing model than any local testing models. However, these were not the
highest performing participants in Experiment 2, with lower
average scores than those described by the \emph{edge} focused
model.  This suggests that we do not yet have a good model of these
participants' choices.

\begin{figure}[t]
   \centering
   \includegraphics[width = .6\columnwidth]{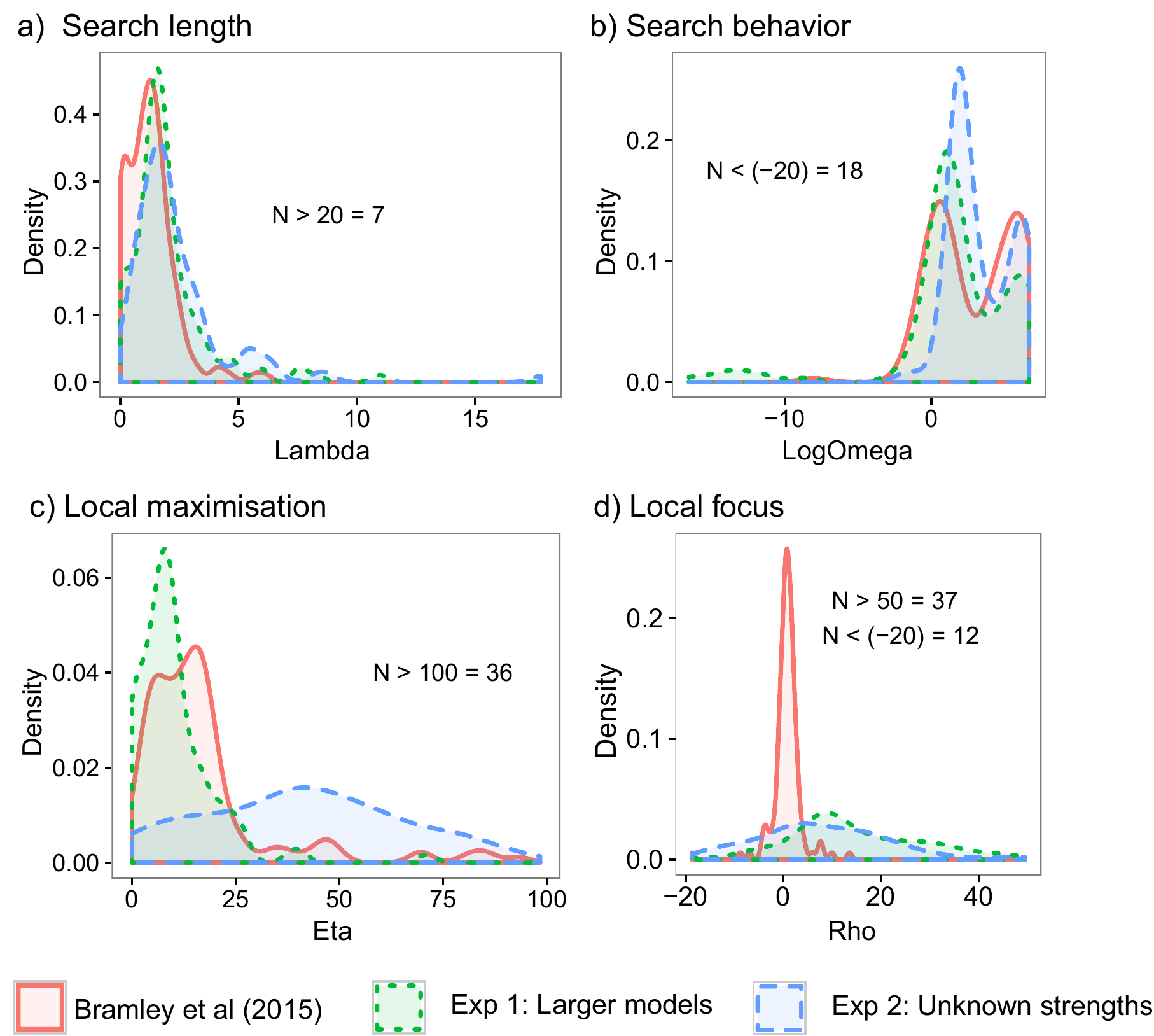}
   \caption{Gaussian kernel densities over fitted model parameters for all participants.  a) Search length $\lambda$ and b) log search behavior $\log(\omega)$ according to \emph{Neurath's  ship} belief update model. c) Local maximization parameter $\eta$ and d) local focus choice parameter $\rho$ under \emph{mixed} local uncertainty based model. Since we report all participants fits, there are some extreme values ---  poorly described by either model --- that are not plotted.  Annotations give the number of parameters above and below the range plotted.}
   \label{fig:model_parameters}
\end{figure}

\begin{figure}[t]
  \centering
   \includegraphics[width = \columnwidth]{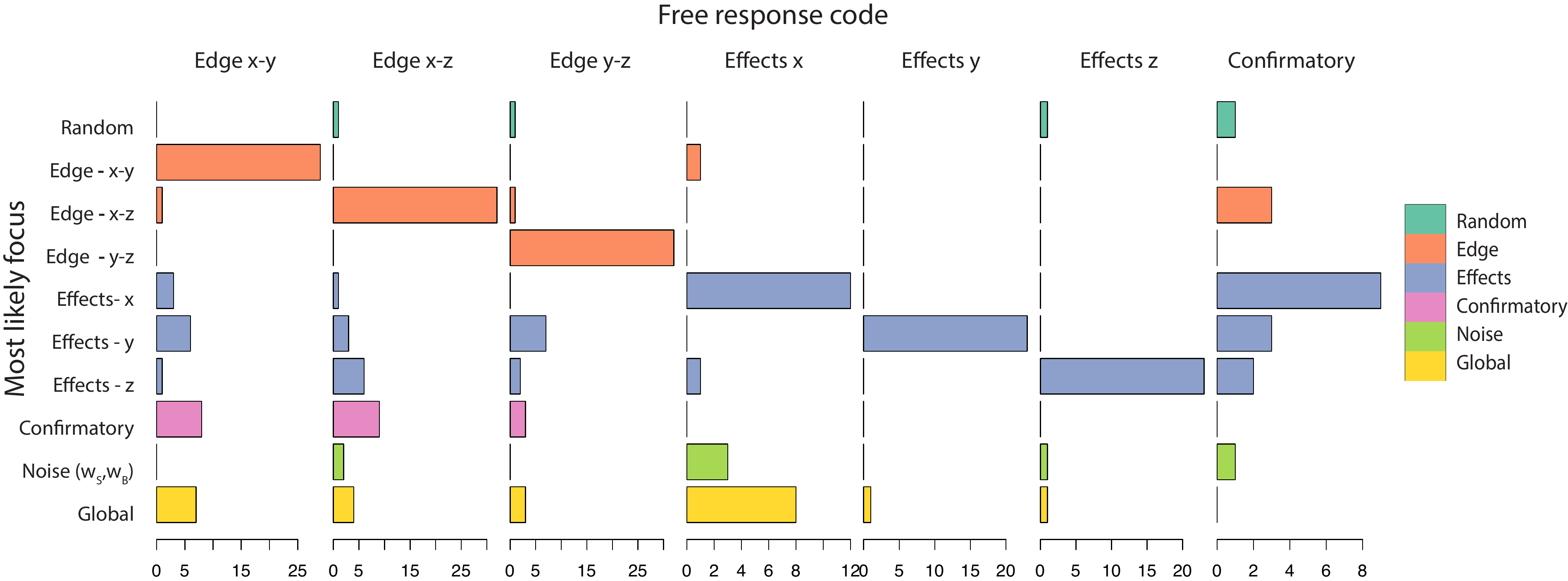}
   \caption{Model and free response correspondence.  Each plot is for trials assigned a particular free response code, each bar is for the number of trials for which that local focus was most likely given the intervention choice.  \emph{Effect} and \emph{edge} coded queries were also diagnosed as such by the model fitting while \emph{confirmatory} coded queries were most likely to be diagnosed as querying the effects of the root node(s) in the true model which always was (or included) $x$.}
   \label{fig:response_model_correspondence}

\end{figure}

\section*{General Discussion}\label{section:GD}

Actively learning causal models is key to higher-level cognition and yet
is radically intractable. We explored how people manage to identify
causal models despite their limited computational resources.  In three
experiments, we found that participants' judgments somewhat reflected
the true posterior, while exhibiting sequential dependencies. Further,
participants' choices of interventions reflected average expected
information, but were insufficiently reactive to the evidence that had
already been observed and were consistent with being locally focused.

We could capture participants' judgment patterns by assuming that they maintained a single causal model rather than a full distribution.  We proposed that participants considered local changes to improve the ability of their single model to explain the latest data and compared this account to two other proposals, one based on the idea that participants occasionally resample from the full posterior, and the other, a heuristic based on ignoring the possibility of indirect effects.  While our \emph{Neurath's ship} proposal fit best overall, all three proposals had merit, with \emph{simple endorsement} winning out in Experiment 2 and more individuals better fit by \emph{win-stay lose-sample}.

We captured participants' interventions by assuming they focused stochastically on different local aspects of the overall uncertainty and tried to resolve these, leading to behavior that was comparatively invariant to the prior.  Our modelling suggested a broad spread of local focuses both between and within participants.

By casting our modeling in the language of machine learning, we were
able to make strong connections between our \emph{Neurath's ship} model
and established techniques for approximating distributions--sequential
Monte-Carlo particle filtering and MCMC (specifically Gibbs) sampling.
Likewise, we were able to explicate intervention selections using the
language of expected uncertainty reduction but relaxing the assumption
that the goal was the minimization of global uncertainty in the full distribution.  The
combination of a single hypothesis (particle) and a Gibbs-esque search,
nicely reflects the \emph{Neurath's ship} intuition that theory change
is necessarily piecemeal and that changes are evaluated against the
backdrop of the rest of the existing theory.

\subsection*{Limitations of Neurath's ship}

Like any theory, Neurath's ship was evaluated against a backdrop of a
number of assumptions. We discuss some of these here.

\subsubsection*{Measurement effects}

In order to explore incremental belief change it was necessary to elicit
multiple judgments and to make two strong assumptions: (1) that these
judgments reflected participants' true and latest beliefs; and (2) that
the repeated elicitations did not fundamentally alter learning
processes. To mitigate problems of these, we both incentivised
participants to draw their best and latest guess at every time point
during the tasks, and examined different reporting conditions to explore
the influence of the elicitations on the learning process.  

In \cite{bramley2015fcs} and the \emph{remain} condition in Experiment 2,
participants could leave parts of their hypothesis untouched if they did
not want to change them.  This had the strength of being minimally
invasive; it did not push the learner to reconsider an edge that they
would otherwise not have done merely because they have been asked about
it again. However this came at the cost of conflating genuine incremental
change in the learner's psychological representation with response
laziness. To assuage this concern, in Experiment 1 and Experiment 2
\emph{disappear}, we removed the participants' previous judgment after
they had seen the outcome of the subsequent intervention, meaning that
they would have to remember and re-report any edges they had previously
judged (and not yet reconsidered).  The slight reduction of dependence between \emph{remain} and \emph{disappear} conditions in Experiment 2, is consistent
with the idea that being forced to re-report edges made it more likely
that they would be reconsidered and potentially changed.

The \emph{Neurath's ship} approach is related to anchor-and-adjust
models \citep{einhorn1986judging,petrov2005dynamics} of sequential
magnitude estimation.  Hogarth and Einhorn found that when mean
estimates are repeatedly elicited from participants as they see a
sequence of numbers, the sequence of responses can be captured by a
process whereby one stores a single value and adjusts it a portion of
the way toward each new observed value.  When judgments were elicited at
the end of the sequence, participants behaved more like they had stored
a subset of the values and averaged them at the end.  In the same way,
we can think of \emph{Neurath's ship} as a process in which the current
model acts as an anchor, and adjustments are made toward new data as it
is observed.  However, the higher complexity of causal inference, and
the greater storage requirements for the individual episodes will
presumably lead to greater pressure to use a sequential strategy rather
than store.  Arguably, step-by-step elicitation is a closer analogue to
real-world causal inference than end-of-sequence because causal beliefs
are presumably in frequent use while learning instances may be spread
out, with no clear start or end.

\subsubsection*{Acyclicity}

We adopted the directed acyclic graph as our model of causal
representation here because it is a standard approach in the literature
and is mathematically convenient. Furthermore, cyclic graphs were quite
rare choices in \cite{bramley2015fcs} (where participants were permitted to draw
them). Thus, we simply opted to to rule them out in the instructions in
later experiments.

However, in tasks where people draw causal models of real-world
phenomena, they often draw cyclic or reciprocal relationships
\citep{nikolic2015there, kim2002clinical}, and many real world processes
are characterized by bidirectional causality, such as supply and demand in
economics or homeostasis in biological systems.  There are various
ways to represent dynamic systems.  One proposal is the dynamic Bayesian
network \citep{dean1989model}, which can be ``unfolded'' to form regular
acyclic network with causal influences passing forward through time.
Another is the chain graph \citep{lauritzen2002chain}, in which
undirected edges are mixed with directed edges and used to model the
equilibria of the cyclic parts of the system. 

Exploring these structures would require a change in the the semantics
of the experiment so that people could understand what they were
reporting in the presence of dynamical interactions. However, given
this, NS would offer a way of performing sequential, on-line,
inference for such structures, using standard likelihood calculations
for dynamic Bayes nets and chain graphs.

\subsubsection*{Evaluation of evidence}

Another pragmatic limitation of the current modeling was the assumption
of the noisy-OR functional form for the true underlying causal
models. While we did take care to train participants on the sources of
noise in both Experiments and the exact values in \cite{bramley2015fcs}, our
own past work suggests that people may have simpler ways of evaluating the
how likely models would be to produce different patterns -- for example, in \cite{bramley2015staying}, we found
participants' judgments could be captured by assuming they lumped
sources of noise together and just counted the number of surprising
outcomes under each model.

One possibility is that people actually formed likelihood estimates
through simulation with an internal causal model. For instance, one
might perform a mental intervention, activating a component of one's own
internal causal model and keeping track of where the activation
propagates.  By simulating multiple times, a learner could estimate the
likelihood of different outcomes under their current model
\citep{hamrick2015think}, and by simulating under variations of the
model, the learner could compare likelihoods generated on the fly.  This
simulation-based view provides a possible explanation for why
participants more readily accommodated internal noise $w_S$ than
background noise $w_B$.  The former can be ``built in'' to the inferred connections in their model and reveal itself in mental simulation, while $w_B$ is more of a mathematical ``catch all'' for all possible influences coming from outside the variables under focus.  The Neurath's ship perspective suggests that people lean on their surrounding network of assumptions about surrounding causes, controlling for these if they get in the way of local inference.  By being omnipresent and affecting all the variables equally $w_B$ was not possible to accommodate in this way.

Future experiments and modeling might relax the assumption of noisy-OR
likelihoods and allow induction of more diverse functional forms, or
focus on well known domains where priors can be measured before the
task.  Another approach might be to render the noisy-OR formalization
more transparent by visualizing the sources of exogenous noise alongside
the target variables, for instance displaying varying numbers of
nuisance background variables on screen for different background noise
conditions. 

\subsubsection*{Antifoundationalism}

The core of Neurath's ship is the strong assumption that people consider
only a single global hypothesis and make local changes within this.
This is the ``antifoundationalism'' captured by the Duhem--Quine thesis --- any
local theoretical claim is necessarily supported by surrounding
assumptions.  However, this may be too strong for some of the easier
problems we considered here where the worlds may have been small and
constrained enough for some people to reason at the global level.  For
the three variable problems in particular, some participants may have
been able to consider alternatives at the level of the whole model, and
thus able to shift from common cause to chain etc with a single step.

While participants' judgments showed high sequential dependence, they
did occasionally change their model abruptly.   The theory of unexpected
uncertainty \citep{yu2003expected}, and substantial work on changepoint
tasks \citep{speekenbrink2010learning} are associated with the notion
that people will sometimes ``start over'' if they are having
consistently poor predictions from their existing model. This relates to
the idea, in philosophy of science, of a ``paradigm shift''
\citep{kuhn1962structure}.  The current \emph{Neurath's ship} models do
not naturally capture this but accommodate occasional large jumps by
assuming a variable search length ($k$), meaning the search will
sometimes be long enough to allow the learner to move to a radically
different model in a single update.  However we might also extend the
\emph{Neurath's ship} framework to include a threshold on prediction
accuracy below which a learner will start afresh, for example by randomly
sampling a model, or sampling from a hitherto unexplored part of the
space. At present this is captured by the $\epsilon$ probability of sampling a new $b^t$ at random on a given trial (which ranged between a probability of .03 in Experiment 2 and .2 in Experiment 1).

\subsubsection*{Selective memory}

We assumed that participants' judgment updates were based on the recent data $\cald_r^t$, collected since the last time they changed their hypothesis. This is quite frugal in the current context, as the learner rarely has to store more than a few tests worth of evidence.  It also captures the idea of semanticization -- that as one gradually absorbs episodic evidence into one's hypothesis, it becomes safe to forget it.

However, the particular choice of $\cald_r^t$ is certainly a simplification.
People may frequently remember evidence from before their latest change,
and fail to store recent evidence, especially once their beliefs become
settled.  They might also collect summary evidence at the level of
individual edges, counting how often pairs of components activate
together for example, or remember evidence about some components but not
others, or only store evidence when it is surprising under the current
model.  In order to fit the models it was necessary to make simplifying
assumptions that captured some form of halfway house between remembering
everything and relying entirely on your hypothesis.  Future studies
might probe exactly what learners can remember during and after learning
to get a finer-grained understanding of the trade-off between
remembering evidence and absorbing it into beliefs.

Related to this, we fit a static search behavior parameter to
participants, finding evidence of moderate hill climbing.  However, a
more realistic depiction might be something more akin to simulated
annealing \citep{hwang1988simulated}.  Learners might begin searching with
more exploratory moves $\omega\approx 1$ so as to explore the space
broadly, and transition toward hill climbing $\omega=\infty$ as they
start to choose what judgment to report.  Alternatively they might
gradually reduce their search length $k$ as pressure to settle on a model increases.

\subsection*{Alternative approximations and representations}

The choice of Gibbs sampling, together with a single particle
approximation, is just one of numerous possible models of structure
inference.  For example we found (data not shown) fairly good fits by
replacing Gibbs sampling with a form of Metropolis-Hastings MCMC
sampling -- using an $MC^3$ proposal and acceptance distribution
\citep{madigan1995bayesian,madigan1994model}.  The two approaches make
similar behavioral predictions but differ somewhat in their internal
architecture -- a Metropolis-Hastings sampler would first generate a
wholesale alternative to the current belief, then make an accept-reject
decision about whether to accept this alternative, while the Gibbs
sampler focuses on one subpart at a time and updates this conditional on
the rest.  Ultimately, the Gibbs sampler did a better job, helping
justify the broader ideas of locality of inference implicit in the
\emph{Neurath's ship} proposal.

An interesting alternative approach to complex model induction via local
computations \citep{fernbach2009causal,waldmann2008causal}, comes from
variational Bayes
\citep{weierstrass1902mathematische,bishop2006pattern}. The idea behind
this is that one can simplify inference by replacing an intractable
distribution, here the distribution over all possible models, with a
simpler one which has degrees of freedom that can be used to allow it to
fit as best as possible. A common choice of simpler distribution
involves factorization, with a multiplicative combination of a set of
simpler parametrized distributions.  Thus, for causal inference one
might make a mean-field approximation \citep{georges1996dynamical} and
suppose the true distribution over models factorizes into independent
distributions for each causal connection.  Divergence between this
approximation and the full model can then be minimized mathematically by
updating each of the local distributions in turn \citep{jaakkola2001tutorial}.
This provides a different perspective on global inference based on local
updates.  Rather than a process of local search where only a single
model is represented at any time, variational Bayes suggests people
maintain many local distributions and try to minimize the
inconsistencies between them.  The biases induced by this process make
the two approaches distinguishable in principle \citep{sanborn2015types}, meaning that an
interesting avenue for future work may be to design experiments that
distinguish between the two approaches to approximation in cognition
.  The truth in our case may be somewhere in between.  For instance, in the current work, we assumed people were able to use recent evidence to estimate their local uncertainty conditional on the rest of the structure, and thus choose where to focus interventions.  To the extent that learners really represent their beliefs with lots of local uncertainties, their representation becomes increasingly variational.

\subsection*{Choosing interventions aboard Neurath's ship}

The largest difference in intervention choices between experiments was that in Experiment 2 constrained interventions (e.g. $\Do[x \! = \!
1,y \! =\! 0]$) were chosen much more frequently.  One explanation for
this is that participants might have been forced to focus their
attention more narrowly in Experiment 2, to compensate for their
additional uncertainty about the noise by using more focused testing.
Another possibility is that the different subject pools drove this
difference.  It is possible that mTurk's older and educationally diverse
participants (Experiments 1) gathered evidence differently from the
young scientifically trained UCL undergraduates (Experiment 2).  This
might have driven the tendency toward more tightly constrained tests in
in Experiment 2.

The idea that people relied on asking a mixture of different types of locally focused
question, was borne out by our
analysis of the coding of participants' free explanations.  Explanations
almost always focused on one specific aspect of the problem, most
frequently on a particular causal connection, or what a particular
component can affect, but also sometimes on parameter uncertainty or, on later tests,
confirming their current hypothesis.  Furthermore, participants almost
always referred to a mix of different local query types over the course
of their six tests.  The apparent shift toward confirmatory testing on the last trial is sensible, since participants knew they would not have more tests to follow up anything new they might discover.  Indeed this shift would be normative in various settings.

Subjective explanations are notoriously problematic
\citep{russo1989validity,ericsson1980verbal,ericsson1993protocol}.  Therefore, we must be careful in interpreting these results.  One common
issue is that eliciting responses concurrently with performing a task
can change behavior, invalidating conclusions about the original
behavior.  We minimized this issue by eliciting explanations
just after each intervention was chosen, before its outcome was
revealed.  Additionally, we did not find any difference in the
distribution of interventions on the free response trials and those
chosen the first time participants identified the chains structure.

A second issue is that there are limits on the kinds of processes people
can describe effectively in natural language, with rule based
explanations being typically easier to express than those involving more
complex statistical weighting and averaging.  That is, even if someone
weighed several factors in coming to a decision, they might explain
this by mentioning only the most significant, or recently
considered of these factors, falsely appearing to have relied on a
one-reason decision strategy.  There is an active debate about this,
including suggestions that people's explanations for their choices are,
in general, post-hoc rationalizations rather than genuine descriptions
of process \citep{dennett1991consciousness,johansson2005failure}, but
also refutations of this interpretation \citep{newell2014unconscious}.

In sum, taken with appropriate caution, we suggest that this analysis
does provides a valuable window on participants' subjective sense of
their active testing, with their relatively specific focus on
one aspect of the uncertainty at a time consistent with the idea that
they rely on a mixture of heuristic questions.

The models pinned down interventions less tightly than beliefs in the
sense that there was a great deal of spread in the individuals best fit
across the models, and the proportional reductions in BIC were
smaller. There are various possible reasons for this.  Firstly, the
models of belief change generally predicted one or few likely models,
whereas there are typically many interventions of roughly equal
informativeness to an ideal learner (see
Figure~\ref{fig:example_local_interventions}), which could be performed
in many different orders. This sets the bar for predictability for
interventions much lower than for the causal judgments.  

Secondly, to the extent that learners chose interventions based on a
reduced encoding of the hypothesis space, we are also forced to average
over our additional uncertainty about exactly which hypotheses or
alternatives they were considering at the moment of choice
\citep{markant2010category}.  

A third issue is that of whether and how learners represented current
uncertainty, and recruited this in choosing what to focus on.  In the
current work we assumed that learners were somewhat able to track the
current local uncertainties and use these to choose what to target next.
The modeling revealed that, relative to the local intervention schema,
the majority of participants did tend to focus on the areas of high
current uncertainty (shown by the predominantly positive $\rho$ in
Figure~\ref{fig:model_parameters} d) but we do not yet have a model for
how they did this.  It is plausible that learners used a heuristic to
estimate their local confidence.  For example, a simple option would be
to accrue confidence in an edge, (or analogously in the the descendants
of a variable or in the current hypothesis) for every search step for
which it is considered and remains unchanged, reducing confidence every
time it changes.  In this way confidence in locales that survive more
data and search become stronger, approximately mimicking reduction in
local uncertainty.

We considered just three of a multitude of possible choices of local
focus. These encompass most extant proposals for human search
heuristics, encapsulating modular \citep{markant2015self} constraint
seeking \citep{ruggeri2014learning} and confirmatory
\citep{klayman1989hypothesis} testing, placing all three
within a unified schema and also showing that many learners 
dynamically switch between them.

Participants' free responses provided a complementary perspective,
suggesting that even initial tests were generated as solutions to
uncertainty about some specific subpart of the overall uncertainty space
-- often the descendants of some particular variable or the presence of
some particular connection.  This suggests that the the most important
step in an intervention selection may not be the final choice of action
but the prior choice of what to focus on next.  This is captured in our
model, under which the values of different interventions for a chosen
focus do not depend on $D^{t-1}$.  This means learners need not do
extensive prospective calculation on every test but can learn gradually,
for instance through experience and preplay \citep{pfeiffer2013hippocampal},
which interventions are likely to be informative relative to generic types of
local focus.  This knowledge could then be transferred to subsequent tests,
and translated to tests with different targets -- e.g. if $Do[x=1]$ is
effective for identifying the effects of $x$ then $Do[y=1]$ will be
effective for identifying the effects of $y$. 

It is worth noting from these data that even when participants'
interventions were relatively uninformative from the perspective of
ideal or even our heuristic learners, their explanations would generally
reveal that they \emph{were} informative with respect to some other
question or source of uncertainty.  For example, participants' tests
that were uninformative with respect to identifying structure were often
revealed, through our free response coding, to have been motivated by a
desire to reduce uncertainty about internal $w_S$ or background $w_B$
noise.\footnote{We might have extended the computational model of
  Bayesian inference to incorporate joint inference over models and
  parameters which would have incorporated this aspect of testing.
  However, this would have complicated analyses since participants were
  ultimately only incentivised to identify the right connections}  From
this perspective we might think of even the completely uninformative
intervention choices (e.g. fixing all the variables) as legitimate tests
of illegitimate hypotheses -- e.g.hypotheses that were outside of the
space of possibilities we intended participants to consider -- such as
whether fixed variables actually always took the states they were fixed
to.  
More research is needed to explicate these internal steps
leading up to an active learning action, but the implication based on
the current research is that the solution will not require that the
learner evaluate all possible outcomes of all possible actions under all
possible models, but rather reflect a mixture of heuristics that can
guide the gradual improvement of the learner's current theory.

\subsection*{The navy of one}

At the start we argued that our Neurath's ship model could be seen as a
single particle combined with an MCMC search.  As such, we are claiming
\emph{Neurath's ship} as a form of boundedly rational approximate
Bayesian inference.  However, it is important to consider the point at
which an approximation becomes so degenerate that it is merely a
complicated way to describe a simple heuristic.  Many would argue that
this line is crossed long before reaching particle filters containing a
single particle, or Markov chains lasting only 1 or 2 steps.  It is
certainly a leap to claim that such a process is calculating a proper
posterior.

One alternative to starting from a normative computational level account
and accepting a distant algorithmic approximation, is to start from the
algorithm, i.e., the simple rules, and consider a computational account
such as satisficing \citep{simon1982models} that provides adequate
license. Our account shares two important problems with this, but avoids
two others.

One shared problem is the provenance of the rules - i.e., the
situation-specific heuristics. We saw this in the manifold choice of
local foci for the choice of intervention -- we do not have an account
of whence these hail. This is a common problem in the context of the
adaptive toolbox \citep{gigerenzer2001adaptive} -- it is hard to have a theory of the
collection of tools.

A second shared problem follows on from this - namely how to choose
which rule to apply under which circumstance. In our case, this is
evident again in the mixtures of local focus rules -- we were not able
to provide a satisfying account of how participants make their selection
of focus on a particular trial. The meta-problem of choosing the correct
heuristic is again a common issue for satisficing approaches.

By contrast with a toolbox approach, though, our account smoothly
captures varying degrees of sophistication between individuals. For
instance, with the \emph{Take the best} heuristic,
\cite{gigerenzer1999simple} give an attractive description of one-reason
decision making that often outperforms regression in describing people's
decisions from multiple cues.  However, subsequent analyses have
revealed that participants behave somewhere between the two
\citep{parpartinpressridge,newell2003take} often using more than one
cue, but certainly less than all the information available.  Thus to
understand their processing we must be able to express the halfway
houses between ideal and overly simplistic processing
\citep{lieder2015use}.  In the same way, the approximate Bayesian
perspective allows us to express different levels of approximation lying
between fully probabilistic and fully heuristic processing, with the
simplest form of Neurath's ship lying at the heuristic end of this road.

A further benefit of our account is the ease of generalization between
tasks. Heuristic models are typically designed for, and are competent
at, specific paradigms. Since they lack a more formal relationship with
approximate rationality, they are hard to combine or often to apply in
different or broader circumstances.

Here, we assumed that learners made updates at the level of individual
directed edges.  Again this is just one illustrative choice, but our
model is consistent with the idea that the learners altered beliefs by
making changes local to arbitrary sub-spaces of an unmanageable learning
problem.  We showed that so long as the learner's updates are
conditioned on the rest of their model, and are appropriately balanced,
the connection to approximate Bayesian inference can be maintained
through the ideas of MCMC sampling and a single-particle particle
filter.  A sophisticated learner might be able to update several edges
of their causal model at a single time, with a more complex proposal
distribution. However, on a larger scale this is still likely to be a
small subset of all potential relata that a learner has encountered,
meaning even the most sophisticated learner must lean on their broader
beliefs for support.

In lower level cognition, inference takes place over simple quantities like magnitudes and is certainly probabilistic in the sense that humans can achieve near optimal integration of noisy signals in a variety of tasks including estimation \citep{miyazaki2005testing} and motor control \citep[e.g.][]{kording2004bayesian}. 
At the top end of higher level cognition we have a global world-view, and explicit reasoning characterized by its single track nature.  Rather than claiming these are completely different processes \citep{evans2003two}, the approximate probabilistic inference perspective can accommodate the whole continuum. At the lower level the brain can average over many values, as in particle filtering \citep{abbott2011exploring}, with a whole fleet of Neurath's ships, or via lots of long chains \citep{lieder2012burn,gershman2012multistability}.  In higher level cognition, however, the hypothesis space becomes increasingly unwieldy, and inference becomes increasingly approximate as it must rely on smaller fleets, i.e., fewer hypotheses, and more local alterations in the face of evidence.  At the very top we have a navy of one, grappling with a single global model that can only be updated incrementally. It is worth noting that individuals can then play the role of particles again in group behavior \citep{courville2007rat}, giving us approximate inference all the way up.

In sum, retaining the Bayesian machinery is valuable even as it becomes degenerate, because it allows us to express heuristic behavior without resorting to separate process model or abandoning close connections to an appropriate computational level understanding.

\subsection*{Scope of the theory}

We modeled causal belief change as a process of gradually updating a
single representation through local, conditional edits.  While we chose
to focus on causal structure inference within the causal Bayes net
framework here, there is no reason why this approach should be limited
to this domain.  By taking the \emph{Neurath's ship} metaphor to reveal
an intuitive answer as to how people sidestep the intractability of
rational theory formation \citep{van2014rational}, we can start to build
more realistic models of how people generate the theories that they do
and how and why they get stuck.  We might explain the induction and
adaptation of many of the rich representations utilized in cognition by
analogous processes. Future work could explore the piecemeal induction
of models involving multinomial, continuous
\citep{pacer2011rational,nodelman2002continuous} or latent variables
\citep{lucas2014discovering}; unrestricted functional forms
\citep{griffiths2009modeling}; hierarchical organization
\citep{griffiths2009theory,williamson2005recursive}; and temporal
\citep{pacer2012elements} and spatial
\citep{ullman2012theory,ullman2014learning,battaglia2013simulation} 
semantics.  We are currently exploring the combination of production
rules \citep{goodman2008rational} and local search to model discovery of
new hypotheses in situations where the space of possibilities is
theoretically infinite.  The sequential conditional re-evaluation process
illustrated by our \emph{Neurath's ship} model shows how this radical
antifoundationalism need not be fatal for theory building in
general.

\section*{Conclusions}

In this paper, we proposed a new model of causal theory change, based on
an old idea from philosophy of science -- that learners cannot maintain
a distribution over all possible beliefs, and so must rely on sequential
local changes to a single representation when updating beliefs to
incorporate new evidence.  We showed that we can provide a good account
of participants' sequences of judgments in three experiments and argued
that our model offers a flexible candidate for explaining how complex
representations can be formed in cognition.  We also analyzed
participants' information-gathering behavior, finding it consistent with
the thesis that learners focus on resolving manageable areas of local
uncertainty rather than global uncertainty, showing cognizance of their
learning limitations. Together these accounts show how people manage to
construct rich, causally-structured representations through their
interactions with a complex noisy world.


\clearpage

\rhead{}
\bibliographystyle{apacite}
\bibliography{refs/myrefs}

\begin{thebibliography}{}

\bibitem [\protect \citeauthoryear {%
Abbott%
, Austerweil%
\BCBL {}\ \BBA {} Griffiths%
}{%
Abbott%
\ \protect \BOthers {.}}{%
{\protect \APACyear {2012}}%
}]{%
abbott2012human}
\APACinsertmetastar {%
abbott2012human}%
\begin{APACrefauthors}%
Abbott, J\BPBI T.%
, Austerweil, J\BPBI L.%
\BCBL {}\ \BBA {} Griffiths, T\BPBI L.%
\end{APACrefauthors}%
\unskip\
\newblock
\APACrefYearMonthDay{2012}{}{}.
\newblock
{\BBOQ}\APACrefatitle {Human memory search as a random walk in a semantic
  network.} {Human memory search as a random walk in a semantic
  network.}{\BBCQ}
\newblock
\BIn{} \APACrefbtitle {{Advances in Neural Information Processing Systems}}
  {{Advances in Neural Information Processing Systems}}\ (\BPGS\ 3050--3058).
\PrintBackRefs{\CurrentBib}

\bibitem [\protect \citeauthoryear {%
Abbott%
\ \BBA {} Griffiths%
}{%
Abbott%
\ \BBA {} Griffiths%
}{%
{\protect \APACyear {2011}}%
}]{%
abbott2011exploring}
\APACinsertmetastar {%
abbott2011exploring}%
\begin{APACrefauthors}%
Abbott, J\BPBI T.%
\BCBT {}\ \BBA {} Griffiths, T\BPBI L.%
\end{APACrefauthors}%
\unskip\
\newblock
\APACrefYearMonthDay{2011}{}{}.
\newblock
{\BBOQ}\APACrefatitle {Exploring the influence of particle filter parameters on
  order effects in causal learning} {Exploring the influence of particle filter
  parameters on order effects in causal learning}.{\BBCQ}
\newblock
\BIn{} \APACrefbtitle {{Proceedings of the 33\textsuperscript{rd} Annual
  Meeting of the Cognitive Science Society}.} {{Proceedings of the
  33\textsuperscript{rd} Annual Meeting of the Cognitive Science Society}.}
\newblock
\APACaddressPublisher{Austin, TX}{Cognitive Science Society}.
\PrintBackRefs{\CurrentBib}

\bibitem [\protect \citeauthoryear {%
Anderson%
}{%
Anderson%
}{%
{\protect \APACyear {1990}}%
}]{%
anderson1990adaptive}
\APACinsertmetastar {%
anderson1990adaptive}%
\begin{APACrefauthors}%
Anderson, J.%
\end{APACrefauthors}%
\unskip\
\newblock
\APACrefYear{1990}.
\newblock
\APACrefbtitle {The Adaptive Character of thought} {The adaptive character of
  thought}.
\newblock
\APACaddressPublisher{}{Erlbaum}.
\PrintBackRefs{\CurrentBib}

\bibitem [\protect \citeauthoryear {%
Austerweil%
\ \BBA {} Griffiths%
}{%
Austerweil%
\ \BBA {} Griffiths%
}{%
{\protect \APACyear {2011}}%
}]{%
austerweil2011seeking}
\APACinsertmetastar {%
austerweil2011seeking}%
\begin{APACrefauthors}%
Austerweil, J\BPBI L.%
\BCBT {}\ \BBA {} Griffiths, T\BPBI L.%
\end{APACrefauthors}%
\unskip\
\newblock
\APACrefYearMonthDay{2011}{}{}.
\newblock
{\BBOQ}\APACrefatitle {Seeking confirmation is rational for deterministic
  hypotheses} {Seeking confirmation is rational for deterministic
  hypotheses}.{\BBCQ}
\newblock
\APACjournalVolNumPages{Cognitive Science}{35}{3}{499--526}.
\PrintBackRefs{\CurrentBib}

\bibitem [\protect \citeauthoryear {%
Battaglia%
, Hamrick%
\BCBL {}\ \BBA {} Tenenbaum%
}{%
Battaglia%
\ \protect \BOthers {.}}{%
{\protect \APACyear {2013}}%
}]{%
battaglia2013simulation}
\APACinsertmetastar {%
battaglia2013simulation}%
\begin{APACrefauthors}%
Battaglia, P\BPBI W.%
, Hamrick, J\BPBI B.%
\BCBL {}\ \BBA {} Tenenbaum, J\BPBI B.%
\end{APACrefauthors}%
\unskip\
\newblock
\APACrefYearMonthDay{2013}{}{}.
\newblock
{\BBOQ}\APACrefatitle {Simulation as an engine of physical scene understanding}
  {Simulation as an engine of physical scene understanding}.{\BBCQ}
\newblock
\APACjournalVolNumPages{{Proceedings of the National Academy of
  Sciences}}{110}{45}{18327--18332}.
\PrintBackRefs{\CurrentBib}

\bibitem [\protect \citeauthoryear {%
Bishop%
}{%
Bishop%
}{%
{\protect \APACyear {2006}}%
}]{%
bishop2006pattern}
\APACinsertmetastar {%
bishop2006pattern}%
\begin{APACrefauthors}%
Bishop, C\BPBI M.%
\end{APACrefauthors}%
\unskip\
\newblock
\APACrefYear{2006}.
\newblock
\APACrefbtitle {Pattern recognition and machine learning} {Pattern recognition
  and machine learning}.
\newblock
\APACaddressPublisher{}{Springer, New York}.
\PrintBackRefs{\CurrentBib}

\bibitem [\protect \citeauthoryear {%
Bonawitz%
, Denison%
, Gopnik%
\BCBL {}\ \BBA {} Griffiths%
}{%
Bonawitz%
\ \protect \BOthers {.}}{%
{\protect \APACyear {2014}}%
}]{%
bonawitz2014win}
\APACinsertmetastar {%
bonawitz2014win}%
\begin{APACrefauthors}%
Bonawitz, E\BPBI B.%
, Denison, S.%
, Gopnik, A.%
\BCBL {}\ \BBA {} Griffiths, T\BPBI L.%
\end{APACrefauthors}%
\unskip\
\newblock
\APACrefYearMonthDay{2014}{}{}.
\newblock
{\BBOQ}\APACrefatitle {Win-Stay, Lose-Sample: A simple sequential algorithm for
  approximating {Bayes}ian inference} {Win-stay, lose-sample: A simple
  sequential algorithm for approximating {Bayes}ian inference}.{\BBCQ}
\newblock
\APACjournalVolNumPages{Cognitive Psychology}{74}{}{35--65}.
\PrintBackRefs{\CurrentBib}

\bibitem [\protect \citeauthoryear {%
Bramley%
, Dayan%
\BCBL {}\ \BBA {} Lagnado%
}{%
Bramley%
, Dayan%
\BCBL {}\ \BBA {} Lagnado%
}{%
{\protect \APACyear {2015}}%
}]{%
bramley2015staying}
\APACinsertmetastar {%
bramley2015staying}%
\begin{APACrefauthors}%
Bramley, N\BPBI R.%
, Dayan, P.%
\BCBL {}\ \BBA {} Lagnado, D\BPBI A.%
\end{APACrefauthors}%
\unskip\
\newblock
\APACrefYearMonthDay{2015}{}{}.
\newblock
{\BBOQ}\APACrefatitle {Staying afloat on {N}eurath's boat: {H}euristics for
  sequential causal learning} {Staying afloat on {N}eurath's boat: {H}euristics
  for sequential causal learning}.{\BBCQ}
\newblock
\BIn{} \APACrefbtitle {{Proceedings of the 37\textsuperscript{th} Annual
  Meeting of the Cognitive Science Society}} {{Proceedings of the
  37\textsuperscript{th} Annual Meeting of the Cognitive Science Society}}\
  (\BPGS\ 262--267).
\newblock
\APACaddressPublisher{Austin, TX}{Cognitive Science Society}.
\PrintBackRefs{\CurrentBib}

\bibitem [\protect \citeauthoryear {%
Bramley%
, Gerstenberg%
\BCBL {}\ \BBA {} Lagnado%
}{%
Bramley%
, Gerstenberg%
\BCBL {}\ \BBA {} Lagnado%
}{%
{\protect \APACyear {2014}}%
}]{%
bramley2014order}
\APACinsertmetastar {%
bramley2014order}%
\begin{APACrefauthors}%
Bramley, N\BPBI R.%
, Gerstenberg, T.%
\BCBL {}\ \BBA {} Lagnado, D\BPBI A.%
\end{APACrefauthors}%
\unskip\
\newblock
\APACrefYearMonthDay{2014}{}{}.
\newblock
{\BBOQ}\APACrefatitle {The order of things: Inferring causal structure from
  temporal patterns} {The order of things: Inferring causal structure from
  temporal patterns}.{\BBCQ}
\newblock
\BIn{} \APACrefbtitle {{Proceedings of the 36\textsuperscript{th} Annual
  Meeting of the Cognitive Science Society}} {{Proceedings of the
  36\textsuperscript{th} Annual Meeting of the Cognitive Science Society}}\
  (\BPGS\ 236--242).
\newblock
\APACaddressPublisher{Austin, TX}{Cognitive Science Society}.
\PrintBackRefs{\CurrentBib}

\bibitem [\protect \citeauthoryear {%
Bramley%
, Lagnado%
\BCBL {}\ \BBA {} Speekenbrink%
}{%
Bramley%
, Lagnado%
\BCBL {}\ \BBA {} Speekenbrink%
}{%
{\protect \APACyear {2015}}%
}]{%
bramley2015fcs}
\APACinsertmetastar {%
bramley2015fcs}%
\begin{APACrefauthors}%
Bramley, N\BPBI R.%
, Lagnado, D\BPBI A.%
\BCBL {}\ \BBA {} Speekenbrink, M.%
\end{APACrefauthors}%
\unskip\
\newblock
\APACrefYearMonthDay{2015}{}{}.
\newblock
{\BBOQ}\APACrefatitle {{Conservative forgetful scholars: How people learn
  causal structure through interventions}} {{Conservative forgetful scholars:
  How people learn causal structure through interventions}}.{\BBCQ}
\newblock
\APACjournalVolNumPages{Journal of Experimental Psychology: Learning, Memory \&
  Cognition}{41}{3}{708--731}.
\PrintBackRefs{\CurrentBib}

\bibitem [\protect \citeauthoryear {%
Bramley%
, Nelson%
, Speekenbrink%
, Crupi%
\BCBL {}\ \BBA {} Lagnado%
}{%
Bramley%
, Nelson%
\BCBL {}\ \protect \BOthers {.}}{%
{\protect \APACyear {2014}}%
}]{%
bramley2014should}
\APACinsertmetastar {%
bramley2014should}%
\begin{APACrefauthors}%
Bramley, N\BPBI R.%
, Nelson, J\BPBI D.%
, Speekenbrink, M.%
, Crupi, V.%
\BCBL {}\ \BBA {} Lagnado, D\BPBI A.%
\end{APACrefauthors}%
\unskip\
\newblock
\APACrefYearMonthDay{2014}{}{}.
\newblock
\APACrefbtitle {{What should causal learners value?}} {{What should causal
  learners value?}}
\newblock
\APAChowpublished {{Poster presented at the Annual Meeting of the Psychonomic
  Society}}.
\PrintBackRefs{\CurrentBib}

\bibitem [\protect \citeauthoryear {%
Buchanan%
, Tenenbaum%
\BCBL {}\ \BBA {} Sobel%
}{%
Buchanan%
\ \protect \BOthers {.}}{%
{\protect \APACyear {2010}}%
}]{%
buchanan2010edge}
\APACinsertmetastar {%
buchanan2010edge}%
\begin{APACrefauthors}%
Buchanan, D\BPBI W.%
, Tenenbaum, J\BPBI B.%
\BCBL {}\ \BBA {} Sobel, D\BPBI M.%
\end{APACrefauthors}%
\unskip\
\newblock
\APACrefYearMonthDay{2010}{}{}.
\newblock
{\BBOQ}\APACrefatitle {Edge replacement and nonindependence in causation} {Edge
  replacement and nonindependence in causation}.{\BBCQ}
\newblock
\BIn{} \APACrefbtitle {{Proceedings of the 32\textsuperscript{nd} Annual
  Meeting of the Cognitive Science Society}} {{Proceedings of the
  32\textsuperscript{nd} Annual Meeting of the Cognitive Science Society}}\
  (\BPGS\ 919--924).
\newblock
\APACaddressPublisher{Austin, TX}{Cognitive Science Society}.
\PrintBackRefs{\CurrentBib}

\bibitem [\protect \citeauthoryear {%
Buhrmester%
, Kwang%
\BCBL {}\ \BBA {} Gosling%
}{%
Buhrmester%
\ \protect \BOthers {.}}{%
{\protect \APACyear {2011}}%
}]{%
buhrmester2011amazon}
\APACinsertmetastar {%
buhrmester2011amazon}%
\begin{APACrefauthors}%
Buhrmester, M.%
, Kwang, T.%
\BCBL {}\ \BBA {} Gosling, S\BPBI D.%
\end{APACrefauthors}%
\unskip\
\newblock
\APACrefYearMonthDay{2011}{}{}.
\newblock
{\BBOQ}\APACrefatitle {Amazon's Mechanical Turk a new source of inexpensive,
  yet high-quality, data?} {Amazon's mechanical turk a new source of
  inexpensive, yet high-quality, data?}{\BBCQ}
\newblock
\APACjournalVolNumPages{Perspectives on Psychological Science}{6}{1}{3--5}.
\PrintBackRefs{\CurrentBib}

\bibitem [\protect \citeauthoryear {%
Cheng%
}{%
Cheng%
}{%
{\protect \APACyear {1997}}%
}]{%
cheng1997from}
\APACinsertmetastar {%
cheng1997from}%
\begin{APACrefauthors}%
Cheng, P\BPBI W.%
\end{APACrefauthors}%
\unskip\
\newblock
\APACrefYearMonthDay{1997}{}{}.
\newblock
{\BBOQ}\APACrefatitle {{From covariation to causation: A causal power theory.}}
  {{From covariation to causation: A causal power theory.}}{\BBCQ}
\newblock
\APACjournalVolNumPages{Psychological Review}{104}{2}{367--405}.
\PrintBackRefs{\CurrentBib}

\bibitem [\protect \citeauthoryear {%
Coenen%
, Rehder%
\BCBL {}\ \BBA {} Gureckis%
}{%
Coenen%
\ \protect \BOthers {.}}{%
{\protect \APACyear {2015}}%
}]{%
coenen2015strategies}
\APACinsertmetastar {%
coenen2015strategies}%
\begin{APACrefauthors}%
Coenen, A.%
, Rehder, R.%
\BCBL {}\ \BBA {} Gureckis, T\BPBI M.%
\end{APACrefauthors}%
\unskip\
\newblock
\APACrefYearMonthDay{2015}{}{}.
\newblock
{\BBOQ}\APACrefatitle {Strategies to intervene on causal systems are adaptively
  selected} {Strategies to intervene on causal systems are adaptively
  selected}.{\BBCQ}
\newblock
\APACjournalVolNumPages{Cognitive Psychology}{79}{}{102--133}.
\PrintBackRefs{\CurrentBib}

\bibitem [\protect \citeauthoryear {%
Cooper%
}{%
Cooper%
}{%
{\protect \APACyear {1990}}%
}]{%
cooper1990computational}
\APACinsertmetastar {%
cooper1990computational}%
\begin{APACrefauthors}%
Cooper, G\BPBI F.%
\end{APACrefauthors}%
\unskip\
\newblock
\APACrefYearMonthDay{1990}{}{}.
\newblock
{\BBOQ}\APACrefatitle {The computational complexity of probabilistic inference
  using {Bayes}ian belief networks} {The computational complexity of
  probabilistic inference using {Bayes}ian belief networks}.{\BBCQ}
\newblock
\APACjournalVolNumPages{Artificial Intelligence}{42}{2}{393--405}.
\PrintBackRefs{\CurrentBib}

\bibitem [\protect \citeauthoryear {%
Cooper%
\ \BBA {} Herskovits%
}{%
Cooper%
\ \BBA {} Herskovits%
}{%
{\protect \APACyear {1992}}%
}]{%
cooper1992bayesian}
\APACinsertmetastar {%
cooper1992bayesian}%
\begin{APACrefauthors}%
Cooper, G\BPBI F.%
\BCBT {}\ \BBA {} Herskovits, E.%
\end{APACrefauthors}%
\unskip\
\newblock
\APACrefYearMonthDay{1992}{}{}.
\newblock
{\BBOQ}\APACrefatitle {A {B}ayesian method for the induction of probabilistic
  networks from data} {A {B}ayesian method for the induction of probabilistic
  networks from data}.{\BBCQ}
\newblock
\APACjournalVolNumPages{Machine Learning}{9}{4}{309--347}.
\PrintBackRefs{\CurrentBib}

\bibitem [\protect \citeauthoryear {%
Courville%
\ \BBA {} Daw%
}{%
Courville%
\ \BBA {} Daw%
}{%
{\protect \APACyear {2007}}%
}]{%
courville2007rat}
\APACinsertmetastar {%
courville2007rat}%
\begin{APACrefauthors}%
Courville, A\BPBI C.%
\BCBT {}\ \BBA {} Daw, N\BPBI D.%
\end{APACrefauthors}%
\unskip\
\newblock
\APACrefYearMonthDay{2007}{}{}.
\newblock
{\BBOQ}\APACrefatitle {The rat as particle filter} {The rat as particle
  filter}.{\BBCQ}
\newblock
\BIn{} \APACrefbtitle {{Advances in Neural Information Processing Systems}}
  {{Advances in Neural Information Processing Systems}}\ (\BPGS\ 369--376).
\PrintBackRefs{\CurrentBib}

\bibitem [\protect \citeauthoryear {%
Crump%
, McDonnell%
\BCBL {}\ \BBA {} Gureckis%
}{%
Crump%
\ \protect \BOthers {.}}{%
{\protect \APACyear {2013}}%
}]{%
crump2013evaluating}
\APACinsertmetastar {%
crump2013evaluating}%
\begin{APACrefauthors}%
Crump, M\BPBI J.%
, McDonnell, J\BPBI V.%
\BCBL {}\ \BBA {} Gureckis, T\BPBI M.%
\end{APACrefauthors}%
\unskip\
\newblock
\APACrefYearMonthDay{2013}{}{}.
\newblock
{\BBOQ}\APACrefatitle {Evaluating Amazon's Mechanical Turk as a tool for
  experimental behavioral research} {Evaluating amazon's mechanical turk as a
  tool for experimental behavioral research}.{\BBCQ}
\newblock
\APACjournalVolNumPages{PloS One}{8}{3}{e57410}.
\PrintBackRefs{\CurrentBib}

\bibitem [\protect \citeauthoryear {%
Dean%
\ \BBA {} Kanazawa%
}{%
Dean%
\ \BBA {} Kanazawa%
}{%
{\protect \APACyear {1989}}%
}]{%
dean1989model}
\APACinsertmetastar {%
dean1989model}%
\begin{APACrefauthors}%
Dean, T.%
\BCBT {}\ \BBA {} Kanazawa, K.%
\end{APACrefauthors}%
\unskip\
\newblock
\APACrefYearMonthDay{1989}{}{}.
\newblock
{\BBOQ}\APACrefatitle {A model for reasoning about persistence and causation}
  {A model for reasoning about persistence and causation}.{\BBCQ}
\newblock
\APACjournalVolNumPages{Computational Intelligence}{5}{2}{142--150}.
\PrintBackRefs{\CurrentBib}

\bibitem [\protect \citeauthoryear {%
DeCarlo%
}{%
DeCarlo%
}{%
{\protect \APACyear {1992}}%
}]{%
decarlo1992intertrial}
\APACinsertmetastar {%
decarlo1992intertrial}%
\begin{APACrefauthors}%
DeCarlo, L\BPBI T.%
\end{APACrefauthors}%
\unskip\
\newblock
\APACrefYearMonthDay{1992}{}{}.
\newblock
{\BBOQ}\APACrefatitle {Intertrial interval and sequential effects in magnitude
  scaling.} {Intertrial interval and sequential effects in magnitude
  scaling.}{\BBCQ}
\newblock
\APACjournalVolNumPages{Journal of Experimental Psychology: Human Perception
  and Performance}{18}{4}{1080}.
\PrintBackRefs{\CurrentBib}

\bibitem [\protect \citeauthoryear {%
Dennett%
}{%
Dennett%
}{%
{\protect \APACyear {1991}}%
}]{%
dennett1991consciousness}
\APACinsertmetastar {%
dennett1991consciousness}%
\begin{APACrefauthors}%
Dennett, D\BPBI C.%
\end{APACrefauthors}%
\unskip\
\newblock
\APACrefYear{1991}.
\newblock
\APACrefbtitle {Consciousness Explained} {Consciousness explained}.
\newblock
\APACaddressPublisher{London, UK}{Penguin}.
\PrintBackRefs{\CurrentBib}

\bibitem [\protect \citeauthoryear {%
Dobson%
}{%
Dobson%
}{%
{\protect \APACyear {2010}}%
}]{%
dobson2010introduction}
\APACinsertmetastar {%
dobson2010introduction}%
\begin{APACrefauthors}%
Dobson, A\BPBI J.%
\end{APACrefauthors}%
\unskip\
\newblock
\APACrefYear{2010}.
\newblock
\APACrefbtitle {An introduction to generalized linear models} {An introduction
  to generalized linear models}.
\newblock
\APACaddressPublisher{}{CRC press}.
\PrintBackRefs{\CurrentBib}

\bibitem [\protect \citeauthoryear {%
Dolan%
\ \BBA {} Dayan%
}{%
Dolan%
\ \BBA {} Dayan%
}{%
{\protect \APACyear {2013}}%
}]{%
dolan2013goals}
\APACinsertmetastar {%
dolan2013goals}%
\begin{APACrefauthors}%
Dolan, R\BPBI J.%
\BCBT {}\ \BBA {} Dayan, P.%
\end{APACrefauthors}%
\unskip\
\newblock
\APACrefYearMonthDay{2013}{}{}.
\newblock
{\BBOQ}\APACrefatitle {Goals and habits in the brain} {Goals and habits in the
  brain}.{\BBCQ}
\newblock
\APACjournalVolNumPages{Neuron}{80}{2}{312--325}.
\PrintBackRefs{\CurrentBib}

\bibitem [\protect \citeauthoryear {%
Duhem%
}{%
Duhem%
}{%
{\protect \APACyear {1991}}%
}]{%
duhem1991aim}
\APACinsertmetastar {%
duhem1991aim}%
\begin{APACrefauthors}%
Duhem, P\BPBI M\BPBI M.%
\end{APACrefauthors}%
\unskip\
\newblock
\APACrefYear{1991}.
\newblock
\APACrefbtitle {The aim and structure of physical theory} {The aim and
  structure of physical theory}.
\newblock
\APACaddressPublisher{}{Princeton University Press}.
\PrintBackRefs{\CurrentBib}

\bibitem [\protect \citeauthoryear {%
Einhorn%
\ \BBA {} Hogarth%
}{%
Einhorn%
\ \BBA {} Hogarth%
}{%
{\protect \APACyear {1986}}%
}]{%
einhorn1986judging}
\APACinsertmetastar {%
einhorn1986judging}%
\begin{APACrefauthors}%
Einhorn, H\BPBI J.%
\BCBT {}\ \BBA {} Hogarth, R\BPBI M.%
\end{APACrefauthors}%
\unskip\
\newblock
\APACrefYearMonthDay{1986}{}{}.
\newblock
{\BBOQ}\APACrefatitle {Judging probable cause.} {Judging probable
  cause.}{\BBCQ}
\newblock
\APACjournalVolNumPages{Psychological Bulletin}{99}{1}{3}.
\PrintBackRefs{\CurrentBib}

\bibitem [\protect \citeauthoryear {%
Ericsson%
\ \BBA {} Simon%
}{%
Ericsson%
\ \BBA {} Simon%
}{%
{\protect \APACyear {1980}}%
}]{%
ericsson1980verbal}
\APACinsertmetastar {%
ericsson1980verbal}%
\begin{APACrefauthors}%
Ericsson, K\BPBI A.%
\BCBT {}\ \BBA {} Simon, H\BPBI A.%
\end{APACrefauthors}%
\unskip\
\newblock
\APACrefYearMonthDay{1980}{}{}.
\newblock
{\BBOQ}\APACrefatitle {Verbal reports as data.} {Verbal reports as
  data.}{\BBCQ}
\newblock
\APACjournalVolNumPages{Psychological Review}{87}{3}{215}.
\PrintBackRefs{\CurrentBib}

\bibitem [\protect \citeauthoryear {%
Ericsson%
\ \BBA {} Simon%
}{%
Ericsson%
\ \BBA {} Simon%
}{%
{\protect \APACyear {1993}}%
}]{%
ericsson1993protocol}
\APACinsertmetastar {%
ericsson1993protocol}%
\begin{APACrefauthors}%
Ericsson, K\BPBI A.%
\BCBT {}\ \BBA {} Simon, H\BPBI A.%
\end{APACrefauthors}%
\unskip\
\newblock
\APACrefYearMonthDay{1993}{}{}.
\newblock
\APACrefbtitle {Protocol Analysis.} {Protocol analysis.}
\newblock
\APACaddressPublisher{}{Cambridge, MA: MIT Press}.
\PrintBackRefs{\CurrentBib}

\bibitem [\protect \citeauthoryear {%
Evans%
}{%
Evans%
}{%
{\protect \APACyear {2003}}%
}]{%
evans2003two}
\APACinsertmetastar {%
evans2003two}%
\begin{APACrefauthors}%
Evans, J.%
\end{APACrefauthors}%
\unskip\
\newblock
\APACrefYearMonthDay{2003}{}{}.
\newblock
{\BBOQ}\APACrefatitle {In two minds: dual-process accounts of reasoning} {In
  two minds: dual-process accounts of reasoning}.{\BBCQ}
\newblock
\APACjournalVolNumPages{Trends in Cognitive Sciences}{7}{10}{454--459}.
\PrintBackRefs{\CurrentBib}

\bibitem [\protect \citeauthoryear {%
Fernbach%
\ \BBA {} Sloman%
}{%
Fernbach%
\ \BBA {} Sloman%
}{%
{\protect \APACyear {2009}}%
}]{%
fernbach2009causal}
\APACinsertmetastar {%
fernbach2009causal}%
\begin{APACrefauthors}%
Fernbach, P\BPBI M.%
\BCBT {}\ \BBA {} Sloman, S\BPBI A.%
\end{APACrefauthors}%
\unskip\
\newblock
\APACrefYearMonthDay{2009}{}{}.
\newblock
{\BBOQ}\APACrefatitle {Causal learning with local computations.} {Causal
  learning with local computations.}{\BBCQ}
\newblock
\APACjournalVolNumPages{Journal of Experimental Psychology: Learning, Memory \&
  Cognition}{35}{3}{678}.
\PrintBackRefs{\CurrentBib}

\bibitem [\protect \citeauthoryear {%
Geman%
\ \BBA {} Geman%
}{%
Geman%
\ \BBA {} Geman%
}{%
{\protect \APACyear {1984}}%
}]{%
geman1984stochastic}
\APACinsertmetastar {%
geman1984stochastic}%
\begin{APACrefauthors}%
Geman, S.%
\BCBT {}\ \BBA {} Geman, D.%
\end{APACrefauthors}%
\unskip\
\newblock
\APACrefYearMonthDay{1984}{}{}.
\newblock
{\BBOQ}\APACrefatitle {Stochastic relaxation, Gibbs distributions, and the
  {Bayes}ian restoration of images} {Stochastic relaxation, gibbs
  distributions, and the {Bayes}ian restoration of images}.{\BBCQ}
\newblock
\APACjournalVolNumPages{Pattern Analysis and Machine
  Intelligence}{}{6}{721--741}.
\PrintBackRefs{\CurrentBib}

\bibitem [\protect \citeauthoryear {%
Georges%
, Kotliar%
, Krauth%
\BCBL {}\ \BBA {} Rozenberg%
}{%
Georges%
\ \protect \BOthers {.}}{%
{\protect \APACyear {1996}}%
}]{%
georges1996dynamical}
\APACinsertmetastar {%
georges1996dynamical}%
\begin{APACrefauthors}%
Georges, A.%
, Kotliar, G.%
, Krauth, W.%
\BCBL {}\ \BBA {} Rozenberg, M\BPBI J.%
\end{APACrefauthors}%
\unskip\
\newblock
\APACrefYearMonthDay{1996}{}{}.
\newblock
{\BBOQ}\APACrefatitle {Dynamical mean-field theory of strongly correlated
  fermion systems and the limit of infinite dimensions} {Dynamical mean-field
  theory of strongly correlated fermion systems and the limit of infinite
  dimensions}.{\BBCQ}
\newblock
\APACjournalVolNumPages{Reviews of Modern Physics}{68}{1}{13}.
\PrintBackRefs{\CurrentBib}

\bibitem [\protect \citeauthoryear {%
Gershman%
, Vul%
\BCBL {}\ \BBA {} Tenenbaum%
}{%
Gershman%
\ \protect \BOthers {.}}{%
{\protect \APACyear {2012}}%
}]{%
gershman2012multistability}
\APACinsertmetastar {%
gershman2012multistability}%
\begin{APACrefauthors}%
Gershman, S\BPBI J.%
, Vul, E.%
\BCBL {}\ \BBA {} Tenenbaum, J\BPBI B.%
\end{APACrefauthors}%
\unskip\
\newblock
\APACrefYearMonthDay{2012}{}{}.
\newblock
{\BBOQ}\APACrefatitle {Multistability and perceptual inference} {Multistability
  and perceptual inference}.{\BBCQ}
\newblock
\APACjournalVolNumPages{Neural Computation}{24}{1}{1--24}.
\PrintBackRefs{\CurrentBib}

\bibitem [\protect \citeauthoryear {%
Gigerenzer%
}{%
Gigerenzer%
}{%
{\protect \APACyear {2001}}%
}]{%
gigerenzer2001adaptive}
\APACinsertmetastar {%
gigerenzer2001adaptive}%
\begin{APACrefauthors}%
Gigerenzer, G.%
\end{APACrefauthors}%
\unskip\
\newblock
\APACrefYearMonthDay{2001}{}{}.
\newblock
{\BBOQ}\APACrefatitle {The Adaptive Toolbox} {The adaptive toolbox}.{\BBCQ}
\newblock
\BIn{} G.~Gigerenzer\ \BBA {} R.~Selten\ (\BEDS), \APACrefbtitle {Bounded
  Rationality} {Bounded rationality}\ (\BPGS\ 37--50).
\newblock
\APACaddressPublisher{}{MIT Press}.
\PrintBackRefs{\CurrentBib}

\bibitem [\protect \citeauthoryear {%
Gigerenzer%
, Todd%
\BCBL {}\ \BBA {} {ABC Research Group}%
}{%
Gigerenzer%
\ \protect \BOthers {.}}{%
{\protect \APACyear {1999}}%
}]{%
gigerenzer1999simple}
\APACinsertmetastar {%
gigerenzer1999simple}%
\begin{APACrefauthors}%
Gigerenzer, G.%
, Todd, P\BPBI M.%
\BCBL {}\ \BBA {} {ABC Research Group}.%
\end{APACrefauthors}%
\unskip\
\newblock
\APACrefYear{1999}.
\newblock
\APACrefbtitle {Simple heuristics that make us smart} {Simple heuristics that
  make us smart}.
\newblock
\APACaddressPublisher{New York}{Oxford University Press}.
\PrintBackRefs{\CurrentBib}

\bibitem [\protect \citeauthoryear {%
Gilden%
}{%
Gilden%
}{%
{\protect \APACyear {2001}}%
}]{%
gilden2001cognitive}
\APACinsertmetastar {%
gilden2001cognitive}%
\begin{APACrefauthors}%
Gilden, D\BPBI L.%
\end{APACrefauthors}%
\unskip\
\newblock
\APACrefYearMonthDay{2001}{}{}.
\newblock
{\BBOQ}\APACrefatitle {Cognitive emissions of 1/f noise.} {Cognitive emissions
  of 1/f noise.}{\BBCQ}
\newblock
\APACjournalVolNumPages{Psychological Review}{108}{1}{33}.
\PrintBackRefs{\CurrentBib}

\bibitem [\protect \citeauthoryear {%
Goodman%
, Tenenbaum%
, Feldman%
\BCBL {}\ \BBA {} Griffiths%
}{%
Goodman%
\ \protect \BOthers {.}}{%
{\protect \APACyear {2008}}%
}]{%
goodman2008rational}
\APACinsertmetastar {%
goodman2008rational}%
\begin{APACrefauthors}%
Goodman, N\BPBI D.%
, Tenenbaum, J\BPBI B.%
, Feldman, J.%
\BCBL {}\ \BBA {} Griffiths, T\BPBI L.%
\end{APACrefauthors}%
\unskip\
\newblock
\APACrefYearMonthDay{2008}{}{}.
\newblock
{\BBOQ}\APACrefatitle {A Rational Analysis of Rule-Based Concept Learning} {A
  rational analysis of rule-based concept learning}.{\BBCQ}
\newblock
\APACjournalVolNumPages{Cognitive Science}{32}{1}{108--154}.
\PrintBackRefs{\CurrentBib}

\bibitem [\protect \citeauthoryear {%
Gosling%
, Vazire%
, Srivastava%
\BCBL {}\ \BBA {} John%
}{%
Gosling%
\ \protect \BOthers {.}}{%
{\protect \APACyear {2004}}%
}]{%
gosling2004should}
\APACinsertmetastar {%
gosling2004should}%
\begin{APACrefauthors}%
Gosling, S\BPBI D.%
, Vazire, S.%
, Srivastava, S.%
\BCBL {}\ \BBA {} John, O\BPBI P.%
\end{APACrefauthors}%
\unskip\
\newblock
\APACrefYearMonthDay{2004}{}{}.
\newblock
{\BBOQ}\APACrefatitle {Should we trust web-based studies? A comparative
  analysis of six preconceptions about internet questionnaires.} {Should we
  trust web-based studies? a comparative analysis of six preconceptions about
  internet questionnaires.}{\BBCQ}
\newblock
\APACjournalVolNumPages{American Psychologist}{59}{2}{93}.
\PrintBackRefs{\CurrentBib}

\bibitem [\protect \citeauthoryear {%
Goudie%
\ \BBA {} Mukherjee%
}{%
Goudie%
\ \BBA {} Mukherjee%
}{%
{\protect \APACyear {2011}}%
}]{%
goudie2011efficient}
\APACinsertmetastar {%
goudie2011efficient}%
\begin{APACrefauthors}%
Goudie, R\BPBI J.%
\BCBT {}\ \BBA {} Mukherjee, S.%
\end{APACrefauthors}%
\unskip\
\newblock
\APACrefYearMonthDay{2011}{}{}.
\newblock
\APACrefbtitle {An efficient {Gibbs} sampler for structural inference in
  {Bayes}ian networks} {An efficient {Gibbs} sampler for structural inference
  in {Bayes}ian networks}\ \APACbVolEdTR{}{\BTR{}}.
\newblock
\APACaddressInstitution{}{Centre for Research in Statistical Methodology,
  Warwick}.
\PrintBackRefs{\CurrentBib}

\bibitem [\protect \citeauthoryear {%
Griffiths%
, Lieder%
\BCBL {}\ \BBA {} Goodman%
}{%
Griffiths%
\ \protect \BOthers {.}}{%
{\protect \APACyear {2015}}%
}]{%
griffiths2015rational}
\APACinsertmetastar {%
griffiths2015rational}%
\begin{APACrefauthors}%
Griffiths, T\BPBI L.%
, Lieder, F.%
\BCBL {}\ \BBA {} Goodman, N\BPBI D.%
\end{APACrefauthors}%
\unskip\
\newblock
\APACrefYearMonthDay{2015}{}{}.
\newblock
{\BBOQ}\APACrefatitle {Rational use of cognitive resources: Levels of analysis
  between the computational and the algorithmic} {Rational use of cognitive
  resources: Levels of analysis between the computational and the
  algorithmic}.{\BBCQ}
\newblock
\APACjournalVolNumPages{Topics in Cognitive Science}{7}{}{217--229}.
\PrintBackRefs{\CurrentBib}

\bibitem [\protect \citeauthoryear {%
Griffiths%
, Lucas%
, Williams%
\BCBL {}\ \BBA {} Kalish%
}{%
Griffiths%
\ \protect \BOthers {.}}{%
{\protect \APACyear {2009}}%
}]{%
griffiths2009modeling}
\APACinsertmetastar {%
griffiths2009modeling}%
\begin{APACrefauthors}%
Griffiths, T\BPBI L.%
, Lucas, C.%
, Williams, J.%
\BCBL {}\ \BBA {} Kalish, M\BPBI L.%
\end{APACrefauthors}%
\unskip\
\newblock
\APACrefYearMonthDay{2009}{}{}.
\newblock
{\BBOQ}\APACrefatitle {Modeling human function learning with {Gauss}ian
  processes} {Modeling human function learning with {Gauss}ian
  processes}.{\BBCQ}
\newblock
\BIn{} \APACrefbtitle {{Advances in Neural Information Processing Systems}}
  {{Advances in Neural Information Processing Systems}}\ (\BPGS\ 553--560).
\PrintBackRefs{\CurrentBib}

\bibitem [\protect \citeauthoryear {%
Griffiths%
\ \BBA {} Tenenbaum%
}{%
Griffiths%
\ \BBA {} Tenenbaum%
}{%
{\protect \APACyear {2007}}%
}]{%
griffiths2007two}
\APACinsertmetastar {%
griffiths2007two}%
\begin{APACrefauthors}%
Griffiths, T\BPBI L.%
\BCBT {}\ \BBA {} Tenenbaum, J\BPBI B.%
\end{APACrefauthors}%
\unskip\
\newblock
\APACrefYearMonthDay{2007}{}{}.
\newblock
{\BBOQ}\APACrefatitle {Two proposals for causal grammars} {Two proposals for
  causal grammars}.{\BBCQ}
\newblock
\BIn{} A.~Gopnik\ \BBA {} L\BPBI E.~Schulz\ (\BEDS), \APACrefbtitle {Causal
  learning: Psychology, Philosophy and Computation.} {Causal learning:
  Psychology, philosophy and computation.}
\newblock
\APACaddressPublisher{Oxford}{Oxford University Press}.
\PrintBackRefs{\CurrentBib}

\bibitem [\protect \citeauthoryear {%
Griffiths%
\ \BBA {} Tenenbaum%
}{%
Griffiths%
\ \BBA {} Tenenbaum%
}{%
{\protect \APACyear {2009}}%
}]{%
griffiths2009theory}
\APACinsertmetastar {%
griffiths2009theory}%
\begin{APACrefauthors}%
Griffiths, T\BPBI L.%
\BCBT {}\ \BBA {} Tenenbaum, J\BPBI B.%
\end{APACrefauthors}%
\unskip\
\newblock
\APACrefYearMonthDay{2009}{}{}.
\newblock
{\BBOQ}\APACrefatitle {Theory-based causal induction} {Theory-based causal
  induction}.{\BBCQ}
\newblock
\APACjournalVolNumPages{Psychological Review}{116}{}{661--716}.
\PrintBackRefs{\CurrentBib}

\bibitem [\protect \citeauthoryear {%
Gureckis%
\ \BBA {} Markant%
}{%
Gureckis%
\ \BBA {} Markant%
}{%
{\protect \APACyear {2009}}%
}]{%
gureckis2009battleship}
\APACinsertmetastar {%
gureckis2009battleship}%
\begin{APACrefauthors}%
Gureckis, T\BPBI M.%
\BCBT {}\ \BBA {} Markant, D.%
\end{APACrefauthors}%
\unskip\
\newblock
\APACrefYearMonthDay{2009}{}{}.
\newblock
{\BBOQ}\APACrefatitle {Active learning strategies in a spatial concept learning
  game} {Active learning strategies in a spatial concept learning game}.{\BBCQ}
\newblock
\BIn{} \APACrefbtitle {{Proceedings of the 31\textsuperscript{st} Annual
  Meeting of the Cognitive Science Society}} {{Proceedings of the
  31\textsuperscript{st} Annual Meeting of the Cognitive Science Society}}\
  (\BPGS\ 3145--3150).
\newblock
\APACaddressPublisher{Austin, TX}{Cognitive Science Society}.
\PrintBackRefs{\CurrentBib}

\bibitem [\protect \citeauthoryear {%
Hamrick%
, Smith%
, Griffiths%
\BCBL {}\ \BBA {} Vul%
}{%
Hamrick%
\ \protect \BOthers {.}}{%
{\protect \APACyear {2015}}%
}]{%
hamrick2015think}
\APACinsertmetastar {%
hamrick2015think}%
\begin{APACrefauthors}%
Hamrick, J\BPBI B.%
, Smith, K\BPBI A.%
, Griffiths, T\BPBI L.%
\BCBL {}\ \BBA {} Vul, E.%
\end{APACrefauthors}%
\unskip\
\newblock
\APACrefYearMonthDay{2015}{}{}.
\newblock
{\BBOQ}\APACrefatitle {Think again? The amount of mental simulation tracks
  uncertainty in the outcome} {Think again? the amount of mental simulation
  tracks uncertainty in the outcome}.{\BBCQ}
\newblock
\BIn{} \APACrefbtitle {{Proceedings of the 37\textsuperscript{th} Annual
  Meeting of the Cognitive Science Society}.} {{Proceedings of the
  37\textsuperscript{th} Annual Meeting of the Cognitive Science Society}.}
\newblock
\APACaddressPublisher{Austin, TX}{Cognitive Science Society}.
\PrintBackRefs{\CurrentBib}

\bibitem [\protect \citeauthoryear {%
Hauser%
\ \BBA {} Schwarz%
}{%
Hauser%
\ \BBA {} Schwarz%
}{%
{\protect \APACyear {2015}}%
}]{%
hauser2015attentive}
\APACinsertmetastar {%
hauser2015attentive}%
\begin{APACrefauthors}%
Hauser, D\BPBI J.%
\BCBT {}\ \BBA {} Schwarz, N.%
\end{APACrefauthors}%
\unskip\
\newblock
\APACrefYearMonthDay{2015}{}{}.
\newblock
{\BBOQ}\APACrefatitle {Attentive Turkers: MTurk participants perform better on
  online attention checks than do subject pool participants} {Attentive
  turkers: Mturk participants perform better on online attention checks than do
  subject pool participants}.{\BBCQ}
\newblock
\APACjournalVolNumPages{Behavior research methods}{}{}{1--8}.
\PrintBackRefs{\CurrentBib}

\bibitem [\protect \citeauthoryear {%
Holyoak%
\ \BBA {} Cheng%
}{%
Holyoak%
\ \BBA {} Cheng%
}{%
{\protect \APACyear {2011}}%
}]{%
holyoak2011causal}
\APACinsertmetastar {%
holyoak2011causal}%
\begin{APACrefauthors}%
Holyoak, K\BPBI J.%
\BCBT {}\ \BBA {} Cheng, P\BPBI W.%
\end{APACrefauthors}%
\unskip\
\newblock
\APACrefYearMonthDay{2011}{{\APACmonth{01}}}{}.
\newblock
{\BBOQ}\APACrefatitle {Causal learning and inference as a rational process: the
  new synthesis.} {Causal learning and inference as a rational process: the new
  synthesis.}{\BBCQ}
\newblock
\APACjournalVolNumPages{Annual Review of Psychology}{62}{}{135--63}.
\PrintBackRefs{\CurrentBib}

\bibitem [\protect \citeauthoryear {%
Hwang%
}{%
Hwang%
}{%
{\protect \APACyear {1988}}%
}]{%
hwang1988simulated}
\APACinsertmetastar {%
hwang1988simulated}%
\begin{APACrefauthors}%
Hwang, C\BHBI R.%
\end{APACrefauthors}%
\unskip\
\newblock
\APACrefYearMonthDay{1988}{}{}.
\newblock
{\BBOQ}\APACrefatitle {Simulated annealing: theory and applications} {Simulated
  annealing: theory and applications}.{\BBCQ}
\newblock
\APACjournalVolNumPages{Acta Applicandae Mathematicae}{12}{1}{108--111}.
\PrintBackRefs{\CurrentBib}

\bibitem [\protect \citeauthoryear {%
Jaakkola%
}{%
Jaakkola%
}{%
{\protect \APACyear {2001}}%
}]{%
jaakkola2001tutorial}
\APACinsertmetastar {%
jaakkola2001tutorial}%
\begin{APACrefauthors}%
Jaakkola, T\BPBI S.%
\end{APACrefauthors}%
\unskip\
\newblock
\APACrefYearMonthDay{2001}{}{}.
\newblock
{\BBOQ}\APACrefatitle {A Tutorial on Variational Approximation Methods} {A
  tutorial on variational approximation methods}.{\BBCQ}
\newblock
\BIn{} M.~Opper\ (\BED), \APACrefbtitle {Advanced mean field methods: theory
  and practice} {Advanced mean field methods: theory and practice}\ (\BPG~129).
\newblock
\APACaddressPublisher{}{MIT Press}.
\PrintBackRefs{\CurrentBib}

\bibitem [\protect \citeauthoryear {%
James%
}{%
James%
}{%
{\protect \APACyear {1890}}%
}]{%
james1890principles}
\APACinsertmetastar {%
james1890principles}%
\begin{APACrefauthors}%
James, W.%
\end{APACrefauthors}%
\unskip\
\newblock
\APACrefYear{1890}.
\newblock
\APACrefbtitle {The Principles of Psychology} {The principles of psychology}.
\newblock
\APACaddressPublisher{}{Dover (2000 reprint)}.
\PrintBackRefs{\CurrentBib}

\bibitem [\protect \citeauthoryear {%
Johansson%
, Hall%
, Sikstr{\"o}m%
\BCBL {}\ \BBA {} Olsson%
}{%
Johansson%
\ \protect \BOthers {.}}{%
{\protect \APACyear {2005}}%
}]{%
johansson2005failure}
\APACinsertmetastar {%
johansson2005failure}%
\begin{APACrefauthors}%
Johansson, P.%
, Hall, L.%
, Sikstr{\"o}m, S.%
\BCBL {}\ \BBA {} Olsson, A.%
\end{APACrefauthors}%
\unskip\
\newblock
\APACrefYearMonthDay{2005}{}{}.
\newblock
{\BBOQ}\APACrefatitle {Failure to detect mismatches between intention and
  outcome in a simple decision task} {Failure to detect mismatches between
  intention and outcome in a simple decision task}.{\BBCQ}
\newblock
\APACjournalVolNumPages{Science}{310}{5745}{116--119}.
\PrintBackRefs{\CurrentBib}

\bibitem [\protect \citeauthoryear {%
Julier%
\ \BBA {} Uhlmann%
}{%
Julier%
\ \BBA {} Uhlmann%
}{%
{\protect \APACyear {1997}}%
}]{%
julier1997new}
\APACinsertmetastar {%
julier1997new}%
\begin{APACrefauthors}%
Julier, S\BPBI J.%
\BCBT {}\ \BBA {} Uhlmann, J\BPBI K.%
\end{APACrefauthors}%
\unskip\
\newblock
\APACrefYearMonthDay{1997}{}{}.
\newblock
{\BBOQ}\APACrefatitle {New extension of the {Kalman} filter to nonlinear
  systems} {New extension of the {Kalman} filter to nonlinear systems}.{\BBCQ}
\newblock
\BIn{} \APACrefbtitle {{Proceedings of the 1997 AeroSense Conference on
  Photonic Quantum Computing}} {{Proceedings of the 1997 AeroSense Conference
  on Photonic Quantum Computing}}\ (\BPGS\ 182--193).
\PrintBackRefs{\CurrentBib}

\bibitem [\protect \citeauthoryear {%
Kahneman%
, Slovic%
\BCBL {}\ \BBA {} Tversky%
}{%
Kahneman%
\ \protect \BOthers {.}}{%
{\protect \APACyear {1982}}%
}]{%
kahneman1982judgment}
\APACinsertmetastar {%
kahneman1982judgment}%
\begin{APACrefauthors}%
Kahneman, D.%
, Slovic, P.%
\BCBL {}\ \BBA {} Tversky, A.%
\end{APACrefauthors}%
\unskip\
\newblock
\APACrefYear{1982}.
\newblock
\APACrefbtitle {Judgment Under Uncertainty: Heuristics and Biases} {Judgment
  under uncertainty: Heuristics and biases}.
\newblock
\APACaddressPublisher{}{Cambridge University Press}.
\PrintBackRefs{\CurrentBib}

\bibitem [\protect \citeauthoryear {%
Kim%
\ \BBA {} Ahn%
}{%
Kim%
\ \BBA {} Ahn%
}{%
{\protect \APACyear {2002}}%
}]{%
kim2002clinical}
\APACinsertmetastar {%
kim2002clinical}%
\begin{APACrefauthors}%
Kim, N\BPBI S.%
\BCBT {}\ \BBA {} Ahn, W\BHBI k.%
\end{APACrefauthors}%
\unskip\
\newblock
\APACrefYearMonthDay{2002}{}{}.
\newblock
{\BBOQ}\APACrefatitle {Clinical psychologists' theory-based representations of
  mental disorders predict their diagnostic reasoning and memory.} {Clinical
  psychologists' theory-based representations of mental disorders predict their
  diagnostic reasoning and memory.}{\BBCQ}
\newblock
\APACjournalVolNumPages{Journal of Experimental Psychology:
  General}{131}{4}{451}.
\PrintBackRefs{\CurrentBib}

\bibitem [\protect \citeauthoryear {%
Klayman%
\ \BBA {} Ha%
}{%
Klayman%
\ \BBA {} Ha%
}{%
{\protect \APACyear {1989}}%
}]{%
klayman1989hypothesis}
\APACinsertmetastar {%
klayman1989hypothesis}%
\begin{APACrefauthors}%
Klayman, J.%
\BCBT {}\ \BBA {} Ha, Y\BHBI w.%
\end{APACrefauthors}%
\unskip\
\newblock
\APACrefYearMonthDay{1989}{}{}.
\newblock
{\BBOQ}\APACrefatitle {Hypothesis testing in rule discovery: Strategy,
  structure, and content.} {Hypothesis testing in rule discovery: Strategy,
  structure, and content.}{\BBCQ}
\newblock
\APACjournalVolNumPages{Journal of Experimental Psychology: Learning, Memory \&
  Cognition}{15}{4}{596}.
\PrintBackRefs{\CurrentBib}

\bibitem [\protect \citeauthoryear {%
K{\"o}rding%
\ \BBA {} Wolpert%
}{%
K{\"o}rding%
\ \BBA {} Wolpert%
}{%
{\protect \APACyear {2004}}%
}]{%
kording2004bayesian}
\APACinsertmetastar {%
kording2004bayesian}%
\begin{APACrefauthors}%
K{\"o}rding, K\BPBI P.%
\BCBT {}\ \BBA {} Wolpert, D\BPBI M.%
\end{APACrefauthors}%
\unskip\
\newblock
\APACrefYearMonthDay{2004}{}{}.
\newblock
{\BBOQ}\APACrefatitle {Bayesian integration in sensorimotor learning} {Bayesian
  integration in sensorimotor learning}.{\BBCQ}
\newblock
\APACjournalVolNumPages{Nature}{427}{6971}{244--247}.
\PrintBackRefs{\CurrentBib}

\bibitem [\protect \citeauthoryear {%
Krippendorff%
}{%
Krippendorff%
}{%
{\protect \APACyear {2012}}%
}]{%
krippendorff2012content}
\APACinsertmetastar {%
krippendorff2012content}%
\begin{APACrefauthors}%
Krippendorff, K.%
\end{APACrefauthors}%
\unskip\
\newblock
\APACrefYear{2012}.
\newblock
\APACrefbtitle {Content analysis: An introduction to its methodology} {Content
  analysis: An introduction to its methodology}.
\newblock
\APACaddressPublisher{}{Sage}.
\PrintBackRefs{\CurrentBib}

\bibitem [\protect \citeauthoryear {%
Kuhn%
}{%
Kuhn%
}{%
{\protect \APACyear {1962}}%
}]{%
kuhn1962structure}
\APACinsertmetastar {%
kuhn1962structure}%
\begin{APACrefauthors}%
Kuhn, T\BPBI S.%
\end{APACrefauthors}%
\unskip\
\newblock
\APACrefYear{1962}.
\newblock
\APACrefbtitle {The structure of scientific revolutions} {The structure of
  scientific revolutions}.
\newblock
\APACaddressPublisher{}{University of Chicago Press}.
\PrintBackRefs{\CurrentBib}

\bibitem [\protect \citeauthoryear {%
Kushnir%
, Gopnik%
, Lucas%
\BCBL {}\ \BBA {} Schulz%
}{%
Kushnir%
\ \protect \BOthers {.}}{%
{\protect \APACyear {2010}}%
}]{%
kushnir2010inferring}
\APACinsertmetastar {%
kushnir2010inferring}%
\begin{APACrefauthors}%
Kushnir, T.%
, Gopnik, A.%
, Lucas, C.%
\BCBL {}\ \BBA {} Schulz, L\BPBI E.%
\end{APACrefauthors}%
\unskip\
\newblock
\APACrefYearMonthDay{2010}{}{}.
\newblock
{\BBOQ}\APACrefatitle {{Inferring hidden causal structure.}} {{Inferring hidden
  causal structure.}}{\BBCQ}
\newblock
\APACjournalVolNumPages{Cognitive Science}{34}{1}{148--60}.
\PrintBackRefs{\CurrentBib}

\bibitem [\protect \citeauthoryear {%
Lagnado%
\ \BBA {} Sloman%
}{%
Lagnado%
\ \BBA {} Sloman%
}{%
{\protect \APACyear {2002}}%
}]{%
lagnado2002learning}
\APACinsertmetastar {%
lagnado2002learning}%
\begin{APACrefauthors}%
Lagnado, D\BPBI A.%
\BCBT {}\ \BBA {} Sloman, S\BPBI A.%
\end{APACrefauthors}%
\unskip\
\newblock
\APACrefYearMonthDay{2002}{}{}.
\newblock
{\BBOQ}\APACrefatitle {Learning causal structure} {Learning causal
  structure}.{\BBCQ}
\newblock
\BIn{} \APACrefbtitle {{Proceedings of the 24\textsuperscript{th} Annual
  Meeting of the Cognitive Science Society}.} {{Proceedings of the
  24\textsuperscript{th} Annual Meeting of the Cognitive Science Society}.}
\newblock
\APACaddressPublisher{}{Erlbaum}.
\PrintBackRefs{\CurrentBib}

\bibitem [\protect \citeauthoryear {%
Lagnado%
\ \BBA {} Sloman%
}{%
Lagnado%
\ \BBA {} Sloman%
}{%
{\protect \APACyear {2004}}%
}]{%
lagnado2004advantage}
\APACinsertmetastar {%
lagnado2004advantage}%
\begin{APACrefauthors}%
Lagnado, D\BPBI A.%
\BCBT {}\ \BBA {} Sloman, S\BPBI A.%
\end{APACrefauthors}%
\unskip\
\newblock
\APACrefYearMonthDay{2004}{}{}.
\newblock
{\BBOQ}\APACrefatitle {The advantage of timely intervention} {The advantage of
  timely intervention}.{\BBCQ}
\newblock
\APACjournalVolNumPages{Journal of Experimental Psychology: Learning, Memory \&
  Cognition}{30}{}{856--876}.
\PrintBackRefs{\CurrentBib}

\bibitem [\protect \citeauthoryear {%
Lagnado%
\ \BBA {} Sloman%
}{%
Lagnado%
\ \BBA {} Sloman%
}{%
{\protect \APACyear {2006}}%
}]{%
lagnado2006time}
\APACinsertmetastar {%
lagnado2006time}%
\begin{APACrefauthors}%
Lagnado, D\BPBI A.%
\BCBT {}\ \BBA {} Sloman, S\BPBI A.%
\end{APACrefauthors}%
\unskip\
\newblock
\APACrefYearMonthDay{2006}{}{}.
\newblock
{\BBOQ}\APACrefatitle {Time as a guide to cause} {Time as a guide to
  cause}.{\BBCQ}
\newblock
\APACjournalVolNumPages{Journal of Experimental Psychology: Learning, Memory \&
  Cognition}{32}{3}{451--60}.
\PrintBackRefs{\CurrentBib}

\bibitem [\protect \citeauthoryear {%
Lagnado%
, Waldmann%
, Hagmayer%
\BCBL {}\ \BBA {} Sloman%
}{%
Lagnado%
\ \protect \BOthers {.}}{%
{\protect \APACyear {2007}}%
}]{%
lagnado2007cues}
\APACinsertmetastar {%
lagnado2007cues}%
\begin{APACrefauthors}%
Lagnado, D\BPBI A.%
, Waldmann, M\BPBI R.%
, Hagmayer, Y.%
\BCBL {}\ \BBA {} Sloman, S\BPBI A.%
\end{APACrefauthors}%
\unskip\
\newblock
\APACrefYearMonthDay{2007}{}{}.
\newblock
{\BBOQ}\APACrefatitle {Beyond covariation: cues to causal structure} {Beyond
  covariation: cues to causal structure}.{\BBCQ}
\newblock
\BIn{} A.~Gopnik\ \BBA {} L\BPBI E.~Schulz\ (\BEDS), \APACrefbtitle {Causal
  Learning: Psychology, Philosophy, and Computation} {Causal learning:
  Psychology, philosophy, and computation}\ (\BPGS\ 154--72).
\newblock
\APACaddressPublisher{London}{Oxford University Press}.
\PrintBackRefs{\CurrentBib}

\bibitem [\protect \citeauthoryear {%
Landis%
\ \BBA {} Koch%
}{%
Landis%
\ \BBA {} Koch%
}{%
{\protect \APACyear {1977}}%
}]{%
landis1977measurement}
\APACinsertmetastar {%
landis1977measurement}%
\begin{APACrefauthors}%
Landis, J\BPBI R.%
\BCBT {}\ \BBA {} Koch, G\BPBI G.%
\end{APACrefauthors}%
\unskip\
\newblock
\APACrefYearMonthDay{1977}{}{}.
\newblock
{\BBOQ}\APACrefatitle {The measurement of observer agreement for categorical
  data} {The measurement of observer agreement for categorical data}.{\BBCQ}
\newblock
\APACjournalVolNumPages{Biometrics}{}{}{159--174}.
\PrintBackRefs{\CurrentBib}

\bibitem [\protect \citeauthoryear {%
Lauritzen%
\ \BBA {} Richardson%
}{%
Lauritzen%
\ \BBA {} Richardson%
}{%
{\protect \APACyear {2002}}%
}]{%
lauritzen2002chain}
\APACinsertmetastar {%
lauritzen2002chain}%
\begin{APACrefauthors}%
Lauritzen, S\BPBI L.%
\BCBT {}\ \BBA {} Richardson, T\BPBI S.%
\end{APACrefauthors}%
\unskip\
\newblock
\APACrefYearMonthDay{2002}{}{}.
\newblock
{\BBOQ}\APACrefatitle {Chain graph models and their causal interpretations}
  {Chain graph models and their causal interpretations}.{\BBCQ}
\newblock
\APACjournalVolNumPages{Journal of the Royal Statistical
  Society}{64}{3}{321--348}.
\PrintBackRefs{\CurrentBib}

\bibitem [\protect \citeauthoryear {%
Lieder%
, Griffiths%
\BCBL {}\ \BBA {} Goodman%
}{%
Lieder%
\ \protect \BOthers {.}}{%
{\protect \APACyear {2012}}%
}]{%
lieder2012burn}
\APACinsertmetastar {%
lieder2012burn}%
\begin{APACrefauthors}%
Lieder, F.%
, Griffiths, T.%
\BCBL {}\ \BBA {} Goodman, N.%
\end{APACrefauthors}%
\unskip\
\newblock
\APACrefYearMonthDay{2012}{}{}.
\newblock
{\BBOQ}\APACrefatitle {Burn-in, bias, and the rationality of anchoring}
  {Burn-in, bias, and the rationality of anchoring}.{\BBCQ}
\newblock
\BIn{} \APACrefbtitle {{Advances in Neural Information Processing Systems}}
  {{Advances in Neural Information Processing Systems}}\ (\BPGS\ 2690--2798).
\PrintBackRefs{\CurrentBib}

\bibitem [\protect \citeauthoryear {%
Lieder%
\ \BBA {} Griffiths%
}{%
Lieder%
\ \BBA {} Griffiths%
}{%
{\protect \APACyear {2015}}%
}]{%
lieder2015use}
\APACinsertmetastar {%
lieder2015use}%
\begin{APACrefauthors}%
Lieder, F.%
\BCBT {}\ \BBA {} Griffiths, T\BPBI L.%
\end{APACrefauthors}%
\unskip\
\newblock
\APACrefYearMonthDay{2015}{}{}.
\newblock
{\BBOQ}\APACrefatitle {When to use which heuristic: A rational solution to the
  strategy selection problem} {When to use which heuristic: A rational solution
  to the strategy selection problem}.{\BBCQ}
\newblock
\BIn{} \APACrefbtitle {{Proceedings of the 37\textsuperscript{th} Annual
  Meeting of the Cognitive Science Society}} {{Proceedings of the
  37\textsuperscript{th} Annual Meeting of the Cognitive Science Society}}\
  (\BPGS\ 1362--1367).
\newblock
\APACaddressPublisher{Austin, TX}{Cognitive Science Society}.
\PrintBackRefs{\CurrentBib}

\bibitem [\protect \citeauthoryear {%
Lieder%
, Griffiths%
, Huys%
\BCBL {}\ \BBA {} Goodman%
}{%
Lieder%
\ \protect \BOthers {.}}{%
{\protect \APACyear {under review}}%
}]{%
liederanchor}
\APACinsertmetastar {%
liederanchor}%
\begin{APACrefauthors}%
Lieder, F.%
, Griffiths, T\BPBI L.%
, Huys, Q\BPBI J\BPBI M.%
\BCBL {}\ \BBA {} Goodman, N\BPBI D.%
\end{APACrefauthors}%
\unskip\
\newblock
\APACrefYearMonthDay{under review}{}{}.
\newblock
{\BBOQ}\APACrefatitle {Reinterpreting anchoring-and-adjustment as rational use
  of cognitive resource} {Reinterpreting anchoring-and-adjustment as rational
  use of cognitive resource}.{\BBCQ}
\newblock

\PrintBackRefs{\CurrentBib}

\bibitem [\protect \citeauthoryear {%
Liu%
\ \BBA {} Chen%
}{%
Liu%
\ \BBA {} Chen%
}{%
{\protect \APACyear {1998}}%
}]{%
liu1998sequential}
\APACinsertmetastar {%
liu1998sequential}%
\begin{APACrefauthors}%
Liu, J\BPBI S.%
\BCBT {}\ \BBA {} Chen, R.%
\end{APACrefauthors}%
\unskip\
\newblock
\APACrefYearMonthDay{1998}{}{}.
\newblock
{\BBOQ}\APACrefatitle {Sequential {Monte Carlo} methods for dynamic systems}
  {Sequential {Monte Carlo} methods for dynamic systems}.{\BBCQ}
\newblock
\APACjournalVolNumPages{Journal of the American Statistical
  Association}{93}{443}{1032--1044}.
\PrintBackRefs{\CurrentBib}

\bibitem [\protect \citeauthoryear {%
Lu%
, Yuille%
, Liljeholm%
, Cheng%
\BCBL {}\ \BBA {} Holyoak%
}{%
Lu%
\ \protect \BOthers {.}}{%
{\protect \APACyear {2008}}%
}]{%
lu2008bayesian}
\APACinsertmetastar {%
lu2008bayesian}%
\begin{APACrefauthors}%
Lu, H.%
, Yuille, A\BPBI L.%
, Liljeholm, M.%
, Cheng, P\BPBI W.%
\BCBL {}\ \BBA {} Holyoak, K\BPBI J.%
\end{APACrefauthors}%
\unskip\
\newblock
\APACrefYearMonthDay{2008}{}{}.
\newblock
{\BBOQ}\APACrefatitle {{Bayes}ian generic priors for causal learning.}
  {{Bayes}ian generic priors for causal learning.}{\BBCQ}
\newblock
\APACjournalVolNumPages{Psychological Review}{115}{4}{955}.
\PrintBackRefs{\CurrentBib}

\bibitem [\protect \citeauthoryear {%
Lucas%
, Bridgers%
, Griffiths%
\BCBL {}\ \BBA {} Gopnik%
}{%
Lucas%
, Bridgers%
\BCBL {}\ \protect \BOthers {.}}{%
{\protect \APACyear {2014}}%
}]{%
lucas2014children}
\APACinsertmetastar {%
lucas2014children}%
\begin{APACrefauthors}%
Lucas, C\BPBI G.%
, Bridgers, S.%
, Griffiths, T\BPBI L.%
\BCBL {}\ \BBA {} Gopnik, A.%
\end{APACrefauthors}%
\unskip\
\newblock
\APACrefYearMonthDay{2014}{}{}.
\newblock
{\BBOQ}\APACrefatitle {When children are better (or at least more open-minded)
  learners than adults: Developmental differences in learning the forms of
  causal relationships} {When children are better (or at least more
  open-minded) learners than adults: Developmental differences in learning the
  forms of causal relationships}.{\BBCQ}
\newblock
\APACjournalVolNumPages{Cognition}{131}{2}{284--299}.
\PrintBackRefs{\CurrentBib}

\bibitem [\protect \citeauthoryear {%
Lucas%
, Holstein%
\BCBL {}\ \BBA {} Kemp%
}{%
Lucas%
, Holstein%
\BCBL {}\ \BBA {} Kemp%
}{%
{\protect \APACyear {2014}}%
}]{%
lucas2014discovering}
\APACinsertmetastar {%
lucas2014discovering}%
\begin{APACrefauthors}%
Lucas, C\BPBI G.%
, Holstein, K.%
\BCBL {}\ \BBA {} Kemp, C.%
\end{APACrefauthors}%
\unskip\
\newblock
\APACrefYearMonthDay{2014}{}{}.
\newblock
{\BBOQ}\APACrefatitle {Discovering hidden causes using statistical evidence}
  {Discovering hidden causes using statistical evidence}.{\BBCQ}
\newblock
\BIn{} \APACrefbtitle {{Proceedings of the 36\textsuperscript{th} Annual
  Meeting of the Cognitive Science Society}.} {{Proceedings of the
  36\textsuperscript{th} Annual Meeting of the Cognitive Science Society}.}
\newblock
\APACaddressPublisher{Austin, TX}{Cognitive Science Society}.
\PrintBackRefs{\CurrentBib}

\bibitem [\protect \citeauthoryear {%
Luce%
}{%
Luce%
}{%
{\protect \APACyear {1959}}%
}]{%
luce1959choice}
\APACinsertmetastar {%
luce1959choice}%
\begin{APACrefauthors}%
Luce, D\BPBI R.%
\end{APACrefauthors}%
\unskip\
\newblock
\APACrefYear{1959}.
\newblock
\APACrefbtitle {Individual choice behavior} {Individual choice behavior}.
\newblock
\APACaddressPublisher{New York}{Wiley}.
\PrintBackRefs{\CurrentBib}

\bibitem [\protect \citeauthoryear {%
Madigan%
\ \BBA {} Raftery%
}{%
Madigan%
\ \BBA {} Raftery%
}{%
{\protect \APACyear {1994}}%
}]{%
madigan1994model}
\APACinsertmetastar {%
madigan1994model}%
\begin{APACrefauthors}%
Madigan, D.%
\BCBT {}\ \BBA {} Raftery, A\BPBI E.%
\end{APACrefauthors}%
\unskip\
\newblock
\APACrefYearMonthDay{1994}{}{}.
\newblock
{\BBOQ}\APACrefatitle {Model selection and accounting for model uncertainty in
  graphical models using Occam's window} {Model selection and accounting for
  model uncertainty in graphical models using occam's window}.{\BBCQ}
\newblock
\APACjournalVolNumPages{Journal of the American Statistical
  Association}{89}{428}{1535--1546}.
\PrintBackRefs{\CurrentBib}

\bibitem [\protect \citeauthoryear {%
Madigan%
, York%
\BCBL {}\ \BBA {} Allard%
}{%
Madigan%
\ \protect \BOthers {.}}{%
{\protect \APACyear {1995}}%
}]{%
madigan1995bayesian}
\APACinsertmetastar {%
madigan1995bayesian}%
\begin{APACrefauthors}%
Madigan, D.%
, York, J.%
\BCBL {}\ \BBA {} Allard, D.%
\end{APACrefauthors}%
\unskip\
\newblock
\APACrefYearMonthDay{1995}{}{}.
\newblock
{\BBOQ}\APACrefatitle {{Bayes}ian graphical models for discrete data}
  {{Bayes}ian graphical models for discrete data}.{\BBCQ}
\newblock
\APACjournalVolNumPages{International Statistical Review}{}{}{215--232}.
\PrintBackRefs{\CurrentBib}

\bibitem [\protect \citeauthoryear {%
Markant%
\ \BBA {} Gureckis%
}{%
Markant%
\ \BBA {} Gureckis%
}{%
{\protect \APACyear {2010}}%
}]{%
markant2010category}
\APACinsertmetastar {%
markant2010category}%
\begin{APACrefauthors}%
Markant, D\BPBI B.%
\BCBT {}\ \BBA {} Gureckis, T\BPBI M.%
\end{APACrefauthors}%
\unskip\
\newblock
\APACrefYearMonthDay{2010}{}{}.
\newblock
{\BBOQ}\APACrefatitle {Category learning through active sampling} {Category
  learning through active sampling}.{\BBCQ}
\newblock
\APACjournalVolNumPages{{Proceedings of the of the 32\textsuperscript{nd}
  Annual Meeting of the Cognitive Science Society}}{}{}{248--253}.
\PrintBackRefs{\CurrentBib}

\bibitem [\protect \citeauthoryear {%
Markant%
\ \BBA {} Gureckis%
}{%
Markant%
\ \BBA {} Gureckis%
}{%
{\protect \APACyear {2012}}%
}]{%
markant2012does}
\APACinsertmetastar {%
markant2012does}%
\begin{APACrefauthors}%
Markant, D\BPBI B.%
\BCBT {}\ \BBA {} Gureckis, T\BPBI M.%
\end{APACrefauthors}%
\unskip\
\newblock
\APACrefYearMonthDay{2012}{}{}.
\newblock
{\BBOQ}\APACrefatitle {{Does the utility of information influence sampling
  behavior?}} {{Does the utility of information influence sampling
  behavior?}}{\BBCQ}
\newblock
\BIn{} \APACrefbtitle {{Proceedings of the 34\textsuperscript{th} Annual
  Meeting of the Cognitive Science Society}.} {{Proceedings of the
  34\textsuperscript{th} Annual Meeting of the Cognitive Science Society}.}
\newblock
\APACaddressPublisher{Austin, TX}{Cognitive Science Society}.
\PrintBackRefs{\CurrentBib}

\bibitem [\protect \citeauthoryear {%
Markant%
, Settles%
\BCBL {}\ \BBA {} Gureckis%
}{%
Markant%
\ \protect \BOthers {.}}{%
{\protect \APACyear {2015}}%
}]{%
markant2015self}
\APACinsertmetastar {%
markant2015self}%
\begin{APACrefauthors}%
Markant, D\BPBI B.%
, Settles, B.%
\BCBL {}\ \BBA {} Gureckis, T\BPBI M.%
\end{APACrefauthors}%
\unskip\
\newblock
\APACrefYearMonthDay{2015}{}{}.
\newblock
{\BBOQ}\APACrefatitle {Self-Directed Learning Favors Local, Rather Than Global,
  Uncertainty} {Self-directed learning favors local, rather than global,
  uncertainty}.{\BBCQ}
\newblock
\APACjournalVolNumPages{Cognitive Science}{}{}{}.
\PrintBackRefs{\CurrentBib}

\bibitem [\protect \citeauthoryear {%
Marr%
}{%
Marr%
}{%
{\protect \APACyear {1982}}%
}]{%
marr1982vision}
\APACinsertmetastar {%
marr1982vision}%
\begin{APACrefauthors}%
Marr, D.%
\end{APACrefauthors}%
\unskip\
\newblock
\APACrefYear{1982}.
\newblock
\APACrefbtitle {Vision} {Vision}.
\newblock
\APACaddressPublisher{New York}{Freeman \& Co}.
\PrintBackRefs{\CurrentBib}

\bibitem [\protect \citeauthoryear {%
Mason%
\ \BBA {} Suri%
}{%
Mason%
\ \BBA {} Suri%
}{%
{\protect \APACyear {2012}}%
}]{%
mason2012conducting}
\APACinsertmetastar {%
mason2012conducting}%
\begin{APACrefauthors}%
Mason, W.%
\BCBT {}\ \BBA {} Suri, S.%
\end{APACrefauthors}%
\unskip\
\newblock
\APACrefYearMonthDay{2012}{}{}.
\newblock
{\BBOQ}\APACrefatitle {Conducting behavioral research on Amazon's Mechanical
  Turk} {Conducting behavioral research on amazon's mechanical turk}.{\BBCQ}
\newblock
\APACjournalVolNumPages{Behavior Research Methods}{44}{1}{1--23}.
\PrintBackRefs{\CurrentBib}

\bibitem [\protect \citeauthoryear {%
McCormack%
, Bramley%
, Frosch%
, Patrick%
\BCBL {}\ \BBA {} Lagnado%
}{%
McCormack%
\ \protect \BOthers {.}}{%
{\protect \APACyear {2016}}%
}]{%
mccormack2016children}
\APACinsertmetastar {%
mccormack2016children}%
\begin{APACrefauthors}%
McCormack, T.%
, Bramley, N\BPBI R.%
, Frosch, C.%
, Patrick, F.%
\BCBL {}\ \BBA {} Lagnado, D\BPBI A.%
\end{APACrefauthors}%
\unskip\
\newblock
\APACrefYearMonthDay{2016}{}{}.
\newblock
{\BBOQ}\APACrefatitle {Children's Use of Interventions to Learn Causal
  Structure} {Children's use of interventions to learn causal
  structure}.{\BBCQ}
\newblock
\APACjournalVolNumPages{Journal of Experimental Child
  Psychology}{141}{}{1--22}.
\PrintBackRefs{\CurrentBib}

\bibitem [\protect \citeauthoryear {%
Metropolis%
, Rosenbluth%
, Rosenbluth%
, Teller%
\BCBL {}\ \BBA {} Teller%
}{%
Metropolis%
\ \protect \BOthers {.}}{%
{\protect \APACyear {1953}}%
}]{%
metropolis1953equation}
\APACinsertmetastar {%
metropolis1953equation}%
\begin{APACrefauthors}%
Metropolis, N.%
, Rosenbluth, A\BPBI W.%
, Rosenbluth, M\BPBI N.%
, Teller, A\BPBI H.%
\BCBL {}\ \BBA {} Teller, E.%
\end{APACrefauthors}%
\unskip\
\newblock
\APACrefYearMonthDay{1953}{}{}.
\newblock
{\BBOQ}\APACrefatitle {Equation of state calculations by fast computing
  machines} {Equation of state calculations by fast computing machines}.{\BBCQ}
\newblock
\APACjournalVolNumPages{The Journal of Chemical Physics}{21}{6}{1087--1092}.
\PrintBackRefs{\CurrentBib}

\bibitem [\protect \citeauthoryear {%
Miyazaki%
, Nozaki%
\BCBL {}\ \BBA {} Nakajima%
}{%
Miyazaki%
\ \protect \BOthers {.}}{%
{\protect \APACyear {2005}}%
}]{%
miyazaki2005testing}
\APACinsertmetastar {%
miyazaki2005testing}%
\begin{APACrefauthors}%
Miyazaki, M.%
, Nozaki, D.%
\BCBL {}\ \BBA {} Nakajima, Y.%
\end{APACrefauthors}%
\unskip\
\newblock
\APACrefYearMonthDay{2005}{}{}.
\newblock
{\BBOQ}\APACrefatitle {Testing Bayesian models of human coincidence timing}
  {Testing bayesian models of human coincidence timing}.{\BBCQ}
\newblock
\APACjournalVolNumPages{Journal of Neurophysiology}{94}{1}{395--399}.
\PrintBackRefs{\CurrentBib}

\bibitem [\protect \citeauthoryear {%
Murphy%
}{%
Murphy%
}{%
{\protect \APACyear {2001}}%
}]{%
murphy2001active}
\APACinsertmetastar {%
murphy2001active}%
\begin{APACrefauthors}%
Murphy, K\BPBI P.%
\end{APACrefauthors}%
\unskip\
\newblock
\APACrefYearMonthDay{2001}{}{}.
\newblock
\APACrefbtitle {Active learning of causal {Bayes} net structure} {Active
  learning of causal {Bayes} net structure}\ \APACbVolEdTR{}{\BTR{}}.
\newblock
\APACaddressInstitution{}{UC Berkeley}.
\PrintBackRefs{\CurrentBib}

\bibitem [\protect \citeauthoryear {%
Navarro%
\ \BBA {} Perfors%
}{%
Navarro%
\ \BBA {} Perfors%
}{%
{\protect \APACyear {2011}}%
}]{%
navarro2011hypothesis}
\APACinsertmetastar {%
navarro2011hypothesis}%
\begin{APACrefauthors}%
Navarro, D\BPBI J.%
\BCBT {}\ \BBA {} Perfors, A\BPBI F.%
\end{APACrefauthors}%
\unskip\
\newblock
\APACrefYearMonthDay{2011}{}{}.
\newblock
{\BBOQ}\APACrefatitle {Hypothesis generation, sparse categories, and the
  positive test strategy.} {Hypothesis generation, sparse categories, and the
  positive test strategy.}{\BBCQ}
\newblock
\APACjournalVolNumPages{Psychological Review}{118}{1}{120}.
\PrintBackRefs{\CurrentBib}

\bibitem [\protect \citeauthoryear {%
Nelson%
}{%
Nelson%
}{%
{\protect \APACyear {2005}}%
}]{%
nelson2005finding}
\APACinsertmetastar {%
nelson2005finding}%
\begin{APACrefauthors}%
Nelson, J\BPBI D.%
\end{APACrefauthors}%
\unskip\
\newblock
\APACrefYearMonthDay{2005}{}{}.
\newblock
{\BBOQ}\APACrefatitle {{Finding useful questions: on {Bayes}ian diagnosticity,
  probability, impact, and information gain.}} {{Finding useful questions: on
  {Bayes}ian diagnosticity, probability, impact, and information gain.}}{\BBCQ}
\newblock
\APACjournalVolNumPages{Psychological Review}{112}{4}{979--99}.
\PrintBackRefs{\CurrentBib}

\bibitem [\protect \citeauthoryear {%
Nelson%
, Divjak%
, Gudmundsdottir%
, Martignon%
\BCBL {}\ \BBA {} Meder%
}{%
Nelson%
\ \protect \BOthers {.}}{%
{\protect \APACyear {2014}}%
}]{%
nelson2014children}
\APACinsertmetastar {%
nelson2014children}%
\begin{APACrefauthors}%
Nelson, J\BPBI D.%
, Divjak, B.%
, Gudmundsdottir, G.%
, Martignon, L\BPBI F.%
\BCBL {}\ \BBA {} Meder, B.%
\end{APACrefauthors}%
\unskip\
\newblock
\APACrefYearMonthDay{2014}{}{}.
\newblock
{\BBOQ}\APACrefatitle {Children's sequential information search is sensitive to
  environmental probabilities} {Children's sequential information search is
  sensitive to environmental probabilities}.{\BBCQ}
\newblock
\APACjournalVolNumPages{Cognition}{130}{1}{74--80}.
\PrintBackRefs{\CurrentBib}

\bibitem [\protect \citeauthoryear {%
Neurath%
}{%
Neurath%
}{%
{\protect \APACyear {1932}}%
}]{%
neurath1983protocol}
\APACinsertmetastar {%
neurath1983protocol}%
\begin{APACrefauthors}%
Neurath, O.%
\end{APACrefauthors}%
\unskip\
\newblock
\APACrefYearMonthDay{1932}{}{}.
\newblock
{\BBOQ}\APACrefatitle {Protocol statements} {Protocol statements}.{\BBCQ}
\newblock
\BIn{} \APACrefbtitle {Philosophical Papers 1913--1946} {Philosophical papers
  1913--1946}\ (\BPGS\ 91--99).
\newblock
\APACaddressPublisher{}{Springer}.
\PrintBackRefs{\CurrentBib}

\bibitem [\protect \citeauthoryear {%
Newell%
\ \BBA {} Shanks%
}{%
Newell%
\ \BBA {} Shanks%
}{%
{\protect \APACyear {2003}}%
}]{%
newell2003take}
\APACinsertmetastar {%
newell2003take}%
\begin{APACrefauthors}%
Newell, B\BPBI R.%
\BCBT {}\ \BBA {} Shanks, D\BPBI R.%
\end{APACrefauthors}%
\unskip\
\newblock
\APACrefYearMonthDay{2003}{}{}.
\newblock
{\BBOQ}\APACrefatitle {{Take the best or look at the rest? Factors influencing
  ``one-reason'' decision making}} {{Take the best or look at the rest? Factors
  influencing ``one-reason'' decision making}}.{\BBCQ}
\newblock
\APACjournalVolNumPages{Journal of Experimental Psychology: Learning, Memory \&
  Cognition}{29}{1}{53--65}.
\PrintBackRefs{\CurrentBib}

\bibitem [\protect \citeauthoryear {%
Newell%
\ \BBA {} Shanks%
}{%
Newell%
\ \BBA {} Shanks%
}{%
{\protect \APACyear {2014}}%
}]{%
newell2014unconscious}
\APACinsertmetastar {%
newell2014unconscious}%
\begin{APACrefauthors}%
Newell, B\BPBI R.%
\BCBT {}\ \BBA {} Shanks, D\BPBI R.%
\end{APACrefauthors}%
\unskip\
\newblock
\APACrefYearMonthDay{2014}{}{}.
\newblock
{\BBOQ}\APACrefatitle {Unconscious influences on decision making: A critical
  review} {Unconscious influences on decision making: A critical
  review}.{\BBCQ}
\newblock
\APACjournalVolNumPages{Behavioral and Brain Sciences}{37}{01}{1--19}.
\PrintBackRefs{\CurrentBib}

\bibitem [\protect \citeauthoryear {%
Nickerson%
}{%
Nickerson%
}{%
{\protect \APACyear {1998}}%
}]{%
nickerson1998confirmation}
\APACinsertmetastar {%
nickerson1998confirmation}%
\begin{APACrefauthors}%
Nickerson, R\BPBI S.%
\end{APACrefauthors}%
\unskip\
\newblock
\APACrefYearMonthDay{1998}{}{}.
\newblock
{\BBOQ}\APACrefatitle {Confirmation bias: A ubiquitous phenomenon in many
  guises} {Confirmation bias: A ubiquitous phenomenon in many guises}.{\BBCQ}
\newblock
\APACjournalVolNumPages{Review of General Psychology}{2}{2}{175}.
\PrintBackRefs{\CurrentBib}

\bibitem [\protect \citeauthoryear {%
Nielsen%
\ \BBA {} Nock%
}{%
Nielsen%
\ \BBA {} Nock%
}{%
{\protect \APACyear {2011}}%
}]{%
nielsen2011closed}
\APACinsertmetastar {%
nielsen2011closed}%
\begin{APACrefauthors}%
Nielsen, F.%
\BCBT {}\ \BBA {} Nock, R.%
\end{APACrefauthors}%
\unskip\
\newblock
\APACrefYearMonthDay{2011}{}{}.
\newblock
{\BBOQ}\APACrefatitle {A closed-form expression for the Sharma--Mittal entropy
  of exponential families} {A closed-form expression for the sharma--mittal
  entropy of exponential families}.{\BBCQ}
\newblock
\APACjournalVolNumPages{Journal of Physics A: Mathematical and
  Theoretical}{45}{3}{032003}.
\PrintBackRefs{\CurrentBib}

\bibitem [\protect \citeauthoryear {%
Nikolic%
\ \BBA {} Lagnado%
}{%
Nikolic%
\ \BBA {} Lagnado%
}{%
{\protect \APACyear {2015}}%
}]{%
nikolic2015there}
\APACinsertmetastar {%
nikolic2015there}%
\begin{APACrefauthors}%
Nikolic, M.%
\BCBT {}\ \BBA {} Lagnado, D\BPBI A.%
\end{APACrefauthors}%
\unskip\
\newblock
\APACrefYearMonthDay{2015}{}{}.
\newblock
{\BBOQ}\APACrefatitle {There aren't plenty more fish in the sea: A causal
  network approach} {There aren't plenty more fish in the sea: A causal network
  approach}.{\BBCQ}
\newblock
\APACjournalVolNumPages{British Journal of Psychology}{106}{4}{564--582}.
\PrintBackRefs{\CurrentBib}

\bibitem [\protect \citeauthoryear {%
Nodelman%
, Shelton%
\BCBL {}\ \BBA {} Koller%
}{%
Nodelman%
\ \protect \BOthers {.}}{%
{\protect \APACyear {2002}}%
}]{%
nodelman2002continuous}
\APACinsertmetastar {%
nodelman2002continuous}%
\begin{APACrefauthors}%
Nodelman, U.%
, Shelton, C\BPBI R.%
\BCBL {}\ \BBA {} Koller, D.%
\end{APACrefauthors}%
\unskip\
\newblock
\APACrefYearMonthDay{2002}{}{}.
\newblock
{\BBOQ}\APACrefatitle {Continuous time Bayesian networks} {Continuous time
  bayesian networks}.{\BBCQ}
\newblock
\BIn{} \APACrefbtitle {{Proceedings of the 18\textsuperscript{th} Conference on
  Uncertainty in Artificial Intelligence}} {{Proceedings of the
  18\textsuperscript{th} Conference on Uncertainty in Artificial
  Intelligence}}\ (\BPGS\ 378--387).
\PrintBackRefs{\CurrentBib}

\bibitem [\protect \citeauthoryear {%
Nyberg%
\ \BBA {} Korb%
}{%
Nyberg%
\ \BBA {} Korb%
}{%
{\protect \APACyear {2006}}%
}]{%
nyberg2006intervention}
\APACinsertmetastar {%
nyberg2006intervention}%
\begin{APACrefauthors}%
Nyberg, E.%
\BCBT {}\ \BBA {} Korb, K.%
\end{APACrefauthors}%
\unskip\
\newblock
\APACrefYearMonthDay{2006}{}{}.
\newblock
{\BBOQ}\APACrefatitle {Informative Interventions} {Informative
  interventions}.{\BBCQ}
\newblock
\BIn{} F.~Russo\ \BBA {} J.~Williamson\ (\BEDS), \APACrefbtitle {Causality and
  Probability in the Sciences.} {Causality and probability in the sciences.}
\newblock
\APACaddressPublisher{College}{London}.
\PrintBackRefs{\CurrentBib}

\bibitem [\protect \citeauthoryear {%
Pacer%
\ \BBA {} Griffiths%
}{%
Pacer%
\ \BBA {} Griffiths%
}{%
{\protect \APACyear {2012}}%
}]{%
pacer2012elements}
\APACinsertmetastar {%
pacer2012elements}%
\begin{APACrefauthors}%
Pacer, M\BPBI D.%
\BCBT {}\ \BBA {} Griffiths, L.%
\end{APACrefauthors}%
\unskip\
\newblock
\APACrefYearMonthDay{2012}{}{}.
\newblock
{\BBOQ}\APACrefatitle {Elements of a rational framework for continuous-time
  causal induction} {Elements of a rational framework for continuous-time
  causal induction}.{\BBCQ}
\newblock
\BIn{} \APACrefbtitle {{Proceedings of the 34\textsuperscript{th} Annual
  Meeting of the Cognitive Science Society}} {{Proceedings of the
  34\textsuperscript{th} Annual Meeting of the Cognitive Science Society}}\
  (\BVOL~1, \BPGS\ 833--838).
\newblock
\APACaddressPublisher{Austin, TX}{Cognitive Science Society}.
\PrintBackRefs{\CurrentBib}

\bibitem [\protect \citeauthoryear {%
Pacer%
\ \BBA {} Griffiths%
}{%
Pacer%
\ \BBA {} Griffiths%
}{%
{\protect \APACyear {2011}}%
}]{%
pacer2011rational}
\APACinsertmetastar {%
pacer2011rational}%
\begin{APACrefauthors}%
Pacer, M\BPBI D.%
\BCBT {}\ \BBA {} Griffiths, T\BPBI L.%
\end{APACrefauthors}%
\unskip\
\newblock
\APACrefYearMonthDay{2011}{}{}.
\newblock
{\BBOQ}\APACrefatitle {A rational model of causal induction with continuous
  causes} {A rational model of causal induction with continuous causes}.{\BBCQ}
\newblock
\BIn{} \APACrefbtitle {{Advances in Neural Information Processing Systems}}
  {{Advances in Neural Information Processing Systems}}\ (\BPGS\ 2384--2392).
\PrintBackRefs{\CurrentBib}

\bibitem [\protect \citeauthoryear {%
Parpart%
, Jones%
\BCBL {}\ \BBA {} Love%
}{%
Parpart%
\ \protect \BOthers {.}}{%
{\protect \APACyear {in revision}}%
}]{%
parpartinpressridge}
\APACinsertmetastar {%
parpartinpressridge}%
\begin{APACrefauthors}%
Parpart, P.%
, Jones, M.%
\BCBL {}\ \BBA {} Love, B.%
\end{APACrefauthors}%
\unskip\
\newblock
\APACrefYearMonthDay{in revision}{}{}.
\newblock
{\BBOQ}\APACrefatitle {Heuristics as {Bayes}ian inference} {Heuristics as
  {Bayes}ian inference}.{\BBCQ}
\newblock

\PrintBackRefs{\CurrentBib}

\bibitem [\protect \citeauthoryear {%
Patil%
, Zhu%
, Kope{\'c}%
\BCBL {}\ \BBA {} Love%
}{%
Patil%
\ \protect \BOthers {.}}{%
{\protect \APACyear {2014}}%
}]{%
patil2014optimal}
\APACinsertmetastar {%
patil2014optimal}%
\begin{APACrefauthors}%
Patil, K\BPBI R.%
, Zhu, X.%
, Kope{\'c}, {\L}.%
\BCBL {}\ \BBA {} Love, B\BPBI C.%
\end{APACrefauthors}%
\unskip\
\newblock
\APACrefYearMonthDay{2014}{}{}.
\newblock
{\BBOQ}\APACrefatitle {Optimal teaching for limited-capacity human learners}
  {Optimal teaching for limited-capacity human learners}.{\BBCQ}
\newblock
\BIn{} \APACrefbtitle {{Advances in Neural Information Processing Systems}}
  {{Advances in Neural Information Processing Systems}}\ (\BPGS\ 2465--2473).
\PrintBackRefs{\CurrentBib}

\bibitem [\protect \citeauthoryear {%
Pearl%
}{%
Pearl%
}{%
{\protect \APACyear {2000}}%
}]{%
pearl2000causality}
\APACinsertmetastar {%
pearl2000causality}%
\begin{APACrefauthors}%
Pearl, J.%
\end{APACrefauthors}%
\unskip\
\newblock
\APACrefYear{2000}.
\newblock
\APACrefbtitle {Causality} {Causality}.
\newblock
\APACaddressPublisher{New York}{Cambridge University Press (2nd edition)}.
\PrintBackRefs{\CurrentBib}

\bibitem [\protect \citeauthoryear {%
Petrov%
\ \BBA {} Anderson%
}{%
Petrov%
\ \BBA {} Anderson%
}{%
{\protect \APACyear {2005}}%
}]{%
petrov2005dynamics}
\APACinsertmetastar {%
petrov2005dynamics}%
\begin{APACrefauthors}%
Petrov, A\BPBI A.%
\BCBT {}\ \BBA {} Anderson, J\BPBI R.%
\end{APACrefauthors}%
\unskip\
\newblock
\APACrefYearMonthDay{2005}{}{}.
\newblock
{\BBOQ}\APACrefatitle {The dynamics of scaling: a memory-based anchor model of
  category rating and absolute identification.} {The dynamics of scaling: a
  memory-based anchor model of category rating and absolute
  identification.}{\BBCQ}
\newblock
\APACjournalVolNumPages{Psychological Review}{112}{2}{383}.
\PrintBackRefs{\CurrentBib}

\bibitem [\protect \citeauthoryear {%
Pfeiffer%
\ \BBA {} Foster%
}{%
Pfeiffer%
\ \BBA {} Foster%
}{%
{\protect \APACyear {2013}}%
}]{%
pfeiffer2013hippocampal}
\APACinsertmetastar {%
pfeiffer2013hippocampal}%
\begin{APACrefauthors}%
Pfeiffer, B\BPBI E.%
\BCBT {}\ \BBA {} Foster, D\BPBI J.%
\end{APACrefauthors}%
\unskip\
\newblock
\APACrefYearMonthDay{2013}{}{}.
\newblock
{\BBOQ}\APACrefatitle {Hippocampal place-cell sequences depict future paths to
  remembered goals} {Hippocampal place-cell sequences depict future paths to
  remembered goals}.{\BBCQ}
\newblock
\APACjournalVolNumPages{Nature}{497}{7447}{74--79}.
\PrintBackRefs{\CurrentBib}

\bibitem [\protect \citeauthoryear {%
Quine%
}{%
Quine%
}{%
{\protect \APACyear {1969}}%
}]{%
quine1969word}
\APACinsertmetastar {%
quine1969word}%
\begin{APACrefauthors}%
Quine, W\BPBI v\BPBI O.%
\end{APACrefauthors}%
\unskip\
\newblock
\APACrefYear{1969}.
\newblock
\APACrefbtitle {Word and object} {Word and object}.
\newblock
\APACaddressPublisher{}{MIT press}.
\PrintBackRefs{\CurrentBib}

\bibitem [\protect \citeauthoryear {%
Rehder%
}{%
Rehder%
}{%
{\protect \APACyear {2014}}%
}]{%
rehder2014independence}
\APACinsertmetastar {%
rehder2014independence}%
\begin{APACrefauthors}%
Rehder, R.%
\end{APACrefauthors}%
\unskip\
\newblock
\APACrefYearMonthDay{2014}{}{}.
\newblock
{\BBOQ}\APACrefatitle {Independence and dependence in human causal reasoning}
  {Independence and dependence in human causal reasoning}.{\BBCQ}
\newblock
\APACjournalVolNumPages{Cognitive Psychology}{72}{}{54--107}.
\PrintBackRefs{\CurrentBib}

\bibitem [\protect \citeauthoryear {%
Rescorla%
\ \BBA {} Wagner%
}{%
Rescorla%
\ \BBA {} Wagner%
}{%
{\protect \APACyear {1972}}%
}]{%
rescorla1972theory}
\APACinsertmetastar {%
rescorla1972theory}%
\begin{APACrefauthors}%
Rescorla, R\BPBI A.%
\BCBT {}\ \BBA {} Wagner, A\BPBI R.%
\end{APACrefauthors}%
\unskip\
\newblock
\APACrefYearMonthDay{1972}{}{}.
\newblock
{\BBOQ}\APACrefatitle {A theory of Pavlovian conditioning: Variations in the
  effectiveness of reinforcement and nonreinforcement} {A theory of pavlovian
  conditioning: Variations in the effectiveness of reinforcement and
  nonreinforcement}.{\BBCQ}
\newblock
\APACjournalVolNumPages{Classical conditioning II: Current research and
  theory}{2}{}{64--99}.
\PrintBackRefs{\CurrentBib}

\bibitem [\protect \citeauthoryear {%
Robinson%
}{%
Robinson%
}{%
{\protect \APACyear {1977}}%
}]{%
robinson1977counting}
\APACinsertmetastar {%
robinson1977counting}%
\begin{APACrefauthors}%
Robinson, R\BPBI W.%
\end{APACrefauthors}%
\unskip\
\newblock
\APACrefYearMonthDay{1977}{}{}.
\newblock
{\BBOQ}\APACrefatitle {Counting unlabeled acyclic digraphs} {Counting unlabeled
  acyclic digraphs}.{\BBCQ}
\newblock
\BIn{} \APACrefbtitle {Combinatorial mathematics V} {Combinatorial mathematics
  v}\ (\BPGS\ 28--43).
\newblock
\APACaddressPublisher{}{Springer}.
\PrintBackRefs{\CurrentBib}

\bibitem [\protect \citeauthoryear {%
Ruggeri%
\ \BBA {} Lombrozo%
}{%
Ruggeri%
\ \BBA {} Lombrozo%
}{%
{\protect \APACyear {2014}}%
}]{%
ruggeri2014learning}
\APACinsertmetastar {%
ruggeri2014learning}%
\begin{APACrefauthors}%
Ruggeri, A.%
\BCBT {}\ \BBA {} Lombrozo, T.%
\end{APACrefauthors}%
\unskip\
\newblock
\APACrefYearMonthDay{2014}{}{}.
\newblock
{\BBOQ}\APACrefatitle {Learning by asking: How children ask questions to
  achieve efficient search} {Learning by asking: How children ask questions to
  achieve efficient search}.{\BBCQ}
\newblock
\BIn{} \APACrefbtitle {36\textsuperscript{th} Annual Meeting of the Cognitive
  Science Society} {36\textsuperscript{th} annual meeting of the cognitive
  science society}\ (\BPGS\ 1335--1340).
\newblock
\APACaddressPublisher{Austin, TX}{Cognitive Science Society}.
\PrintBackRefs{\CurrentBib}

\bibitem [\protect \citeauthoryear {%
Russo%
, Johnson%
\BCBL {}\ \BBA {} Stephens%
}{%
Russo%
\ \protect \BOthers {.}}{%
{\protect \APACyear {1989}}%
}]{%
russo1989validity}
\APACinsertmetastar {%
russo1989validity}%
\begin{APACrefauthors}%
Russo, J\BPBI E.%
, Johnson, E\BPBI J.%
\BCBL {}\ \BBA {} Stephens, D\BPBI L.%
\end{APACrefauthors}%
\unskip\
\newblock
\APACrefYearMonthDay{1989}{}{}.
\newblock
{\BBOQ}\APACrefatitle {The validity of verbal protocols} {The validity of
  verbal protocols}.{\BBCQ}
\newblock
\APACjournalVolNumPages{Memory \& Cognition}{17}{6}{759--769}.
\PrintBackRefs{\CurrentBib}

\bibitem [\protect \citeauthoryear {%
Sanborn%
}{%
Sanborn%
}{%
{\protect \APACyear {2015}}%
}]{%
sanborn2015types}
\APACinsertmetastar {%
sanborn2015types}%
\begin{APACrefauthors}%
Sanborn, A\BPBI N.%
\end{APACrefauthors}%
\unskip\
\newblock
\APACrefYearMonthDay{2015}{}{}.
\newblock
{\BBOQ}\APACrefatitle {Types of approximation for probabilistic cognition:
  Sampling and variational} {Types of approximation for probabilistic
  cognition: Sampling and variational}.{\BBCQ}
\newblock
\APACjournalVolNumPages{Brain and cognition}{to appear}{}{}.
\PrintBackRefs{\CurrentBib}

\bibitem [\protect \citeauthoryear {%
Sanborn%
, Griffiths%
\BCBL {}\ \BBA {} Navarro%
}{%
Sanborn%
\ \protect \BOthers {.}}{%
{\protect \APACyear {2010}}%
}]{%
sanborn2010rational}
\APACinsertmetastar {%
sanborn2010rational}%
\begin{APACrefauthors}%
Sanborn, A\BPBI N.%
, Griffiths, T\BPBI L.%
\BCBL {}\ \BBA {} Navarro, D\BPBI J.%
\end{APACrefauthors}%
\unskip\
\newblock
\APACrefYearMonthDay{2010}{}{}.
\newblock
{\BBOQ}\APACrefatitle {Rational approximations to rational models: alternative
  algorithms for category learning.} {Rational approximations to rational
  models: alternative algorithms for category learning.}{\BBCQ}
\newblock
\APACjournalVolNumPages{Psychological Review}{117}{4}{1144}.
\PrintBackRefs{\CurrentBib}

\bibitem [\protect \citeauthoryear {%
Schwarz%
}{%
Schwarz%
}{%
{\protect \APACyear {1978}}%
}]{%
schwarz1978estimating}
\APACinsertmetastar {%
schwarz1978estimating}%
\begin{APACrefauthors}%
Schwarz, G.%
\end{APACrefauthors}%
\unskip\
\newblock
\APACrefYearMonthDay{1978}{}{}.
\newblock
{\BBOQ}\APACrefatitle {Estimating the dimension of a model} {Estimating the
  dimension of a model}.{\BBCQ}
\newblock
\APACjournalVolNumPages{The Annals of Statistics}{6}{2}{461--464}.
\PrintBackRefs{\CurrentBib}

\bibitem [\protect \citeauthoryear {%
Settles%
}{%
Settles%
}{%
{\protect \APACyear {2012}}%
}]{%
settles2012active}
\APACinsertmetastar {%
settles2012active}%
\begin{APACrefauthors}%
Settles, B.%
\end{APACrefauthors}%
\unskip\
\newblock
\APACrefYearMonthDay{2012}{}{}.
\newblock
{\BBOQ}\APACrefatitle {Active learning} {Active learning}.{\BBCQ}
\newblock
\APACjournalVolNumPages{Synthesis Lectures on Artificial Intelligence and
  Machine Learning}{6}{1}{1--114}.
\PrintBackRefs{\CurrentBib}

\bibitem [\protect \citeauthoryear {%
Shannon%
}{%
Shannon%
}{%
{\protect \APACyear {1951}}%
}]{%
shannon1951prediction}
\APACinsertmetastar {%
shannon1951prediction}%
\begin{APACrefauthors}%
Shannon, C\BPBI E.%
\end{APACrefauthors}%
\unskip\
\newblock
\APACrefYearMonthDay{1951}{}{}.
\newblock
{\BBOQ}\APACrefatitle {Prediction and entropy of printed English} {Prediction
  and entropy of printed english}.{\BBCQ}
\newblock
\APACjournalVolNumPages{The Bell System Technical Journal}{30}{}{50--64}.
\PrintBackRefs{\CurrentBib}

\bibitem [\protect \citeauthoryear {%
Simon%
}{%
Simon%
}{%
{\protect \APACyear {1982}}%
}]{%
simon1982models}
\APACinsertmetastar {%
simon1982models}%
\begin{APACrefauthors}%
Simon, H\BPBI A.%
\end{APACrefauthors}%
\unskip\
\newblock
\APACrefYear{1982}.
\newblock
\APACrefbtitle {Models of bounded rationality: Empirically grounded economic
  reason} {Models of bounded rationality: Empirically grounded economic
  reason}.
\newblock
\APACaddressPublisher{}{MIT press}.
\PrintBackRefs{\CurrentBib}

\bibitem [\protect \citeauthoryear {%
Sloman%
}{%
Sloman%
}{%
{\protect \APACyear {2005}}%
}]{%
sloman2005causal}
\APACinsertmetastar {%
sloman2005causal}%
\begin{APACrefauthors}%
Sloman, S\BPBI A.%
\end{APACrefauthors}%
\unskip\
\newblock
\APACrefYear{2005}.
\newblock
\APACrefbtitle {Causal models: How people think about the world and its
  alternatives} {Causal models: How people think about the world and its
  alternatives}.
\newblock
\APACaddressPublisher{}{Oxford University Press}.
\PrintBackRefs{\CurrentBib}

\bibitem [\protect \citeauthoryear {%
Sobel%
\ \BBA {} Kushnir%
}{%
Sobel%
\ \BBA {} Kushnir%
}{%
{\protect \APACyear {2006}}%
}]{%
sobel2006importance}
\APACinsertmetastar {%
sobel2006importance}%
\begin{APACrefauthors}%
Sobel, D\BPBI M.%
\BCBT {}\ \BBA {} Kushnir, T.%
\end{APACrefauthors}%
\unskip\
\newblock
\APACrefYearMonthDay{2006}{}{}.
\newblock
{\BBOQ}\APACrefatitle {The importance of decision making in causal learning
  from interventions} {The importance of decision making in causal learning
  from interventions}.{\BBCQ}
\newblock
\APACjournalVolNumPages{Memory \& Cognition}{34}{2}{411--419}.
\PrintBackRefs{\CurrentBib}

\bibitem [\protect \citeauthoryear {%
Speekenbrink%
\ \BBA {} Shanks%
}{%
Speekenbrink%
\ \BBA {} Shanks%
}{%
{\protect \APACyear {2010}}%
}]{%
speekenbrink2010learning}
\APACinsertmetastar {%
speekenbrink2010learning}%
\begin{APACrefauthors}%
Speekenbrink, M.%
\BCBT {}\ \BBA {} Shanks, D\BPBI R.%
\end{APACrefauthors}%
\unskip\
\newblock
\APACrefYearMonthDay{2010}{}{}.
\newblock
{\BBOQ}\APACrefatitle {Learning in a changing environment} {Learning in a
  changing environment}.{\BBCQ}
\newblock
\APACjournalVolNumPages{Journal of Experimental Psychology:
  General}{139}{2}{266}.
\PrintBackRefs{\CurrentBib}

\bibitem [\protect \citeauthoryear {%
Steyvers%
, Tenenbaum%
, Wagenmakers%
\BCBL {}\ \BBA {} Blum%
}{%
Steyvers%
\ \protect \BOthers {.}}{%
{\protect \APACyear {2003}}%
}]{%
steyvers2003intervention}
\APACinsertmetastar {%
steyvers2003intervention}%
\begin{APACrefauthors}%
Steyvers, M.%
, Tenenbaum, J\BPBI B.%
, Wagenmakers, E.%
\BCBL {}\ \BBA {} Blum, B.%
\end{APACrefauthors}%
\unskip\
\newblock
\APACrefYearMonthDay{2003}{}{}.
\newblock
{\BBOQ}\APACrefatitle {Inferring causal networks from observations and
  interventions} {Inferring causal networks from observations and
  interventions}.{\BBCQ}
\newblock
\APACjournalVolNumPages{Cognitive Science}{27}{}{453--489}.
\PrintBackRefs{\CurrentBib}

\bibitem [\protect \citeauthoryear {%
Treisman%
\ \BBA {} Williams%
}{%
Treisman%
\ \BBA {} Williams%
}{%
{\protect \APACyear {1984}}%
}]{%
treisman1984theory}
\APACinsertmetastar {%
treisman1984theory}%
\begin{APACrefauthors}%
Treisman, M.%
\BCBT {}\ \BBA {} Williams, T\BPBI C.%
\end{APACrefauthors}%
\unskip\
\newblock
\APACrefYearMonthDay{1984}{}{}.
\newblock
{\BBOQ}\APACrefatitle {A theory of criterion setting with an application to
  sequential dependencies.} {A theory of criterion setting with an application
  to sequential dependencies.}{\BBCQ}
\newblock
\APACjournalVolNumPages{Psychological Review}{91}{1}{68}.
\PrintBackRefs{\CurrentBib}

\bibitem [\protect \citeauthoryear {%
Tsamardinos%
, Brown%
\BCBL {}\ \BBA {} Aliferis%
}{%
Tsamardinos%
\ \protect \BOthers {.}}{%
{\protect \APACyear {2006}}%
}]{%
tsamardinos2006max}
\APACinsertmetastar {%
tsamardinos2006max}%
\begin{APACrefauthors}%
Tsamardinos, I.%
, Brown, L\BPBI E.%
\BCBL {}\ \BBA {} Aliferis, C\BPBI F.%
\end{APACrefauthors}%
\unskip\
\newblock
\APACrefYearMonthDay{2006}{}{}.
\newblock
{\BBOQ}\APACrefatitle {The max-min hill-climbing {Bayes}ian network structure
  learning algorithm} {The max-min hill-climbing {Bayes}ian network structure
  learning algorithm}.{\BBCQ}
\newblock
\APACjournalVolNumPages{Machine Learning}{65}{1}{31--78}.
\PrintBackRefs{\CurrentBib}

\bibitem [\protect \citeauthoryear {%
Ullman%
, Goodman%
\BCBL {}\ \BBA {} Tenenbaum%
}{%
Ullman%
\ \protect \BOthers {.}}{%
{\protect \APACyear {2012}}%
}]{%
ullman2012theory}
\APACinsertmetastar {%
ullman2012theory}%
\begin{APACrefauthors}%
Ullman, T.%
, Goodman, N.%
\BCBL {}\ \BBA {} Tenenbaum, J\BPBI B.%
\end{APACrefauthors}%
\unskip\
\newblock
\APACrefYearMonthDay{2012}{}{}.
\newblock
{\BBOQ}\APACrefatitle {Theory acquisition as stochastic search in a language of
  thought} {Theory acquisition as stochastic search in a language of
  thought}.{\BBCQ}
\newblock
\APACjournalVolNumPages{Cognitive Development}{27}{}{455--480}.
\PrintBackRefs{\CurrentBib}

\bibitem [\protect \citeauthoryear {%
Ullman%
, Stuhlm{\"u}ller%
, Goodman%
\BCBL {}\ \BBA {} Tenenbaum%
}{%
Ullman%
\ \protect \BOthers {.}}{%
{\protect \APACyear {2014}}%
}]{%
ullman2014learning}
\APACinsertmetastar {%
ullman2014learning}%
\begin{APACrefauthors}%
Ullman, T.%
, Stuhlm{\"u}ller, A.%
, Goodman, N.%
\BCBL {}\ \BBA {} Tenenbaum, J\BPBI B.%
\end{APACrefauthors}%
\unskip\
\newblock
\APACrefYearMonthDay{2014}{}{}.
\newblock
{\BBOQ}\APACrefatitle {Learning physics from dynamical scenes} {Learning
  physics from dynamical scenes}.{\BBCQ}
\newblock
\BIn{} \APACrefbtitle {{Proceedings of the 36\textsuperscript{th} Annual
  Meeting of the Cognitive Science Society}.} {{Proceedings of the
  36\textsuperscript{th} Annual Meeting of the Cognitive Science Society}.}
\newblock
\APACaddressPublisher{Austin, TX}{Cognitive Science Society}.
\PrintBackRefs{\CurrentBib}

\bibitem [\protect \citeauthoryear {%
van Rooij%
, Wright%
, Kwisthout%
\BCBL {}\ \BBA {} Wareham%
}{%
van Rooij%
\ \protect \BOthers {.}}{%
{\protect \APACyear {2014}}%
}]{%
van2014rational}
\APACinsertmetastar {%
van2014rational}%
\begin{APACrefauthors}%
van Rooij, I.%
, Wright, C\BPBI D.%
, Kwisthout, J.%
\BCBL {}\ \BBA {} Wareham, T.%
\end{APACrefauthors}%
\unskip\
\newblock
\APACrefYearMonthDay{2014}{}{}.
\newblock
{\BBOQ}\APACrefatitle {Rational analysis, intractability, and the prospects of
  `as if'-explanations} {Rational analysis, intractability, and the prospects
  of `as if'-explanations}.{\BBCQ}
\newblock
\APACjournalVolNumPages{Synthese}{}{}{1--20}.
\PrintBackRefs{\CurrentBib}

\bibitem [\protect \citeauthoryear {%
Vul%
, Goodman%
, Griffiths%
\BCBL {}\ \BBA {} Tenenbaum%
}{%
Vul%
\ \protect \BOthers {.}}{%
{\protect \APACyear {2009}}%
}]{%
vul2009one}
\APACinsertmetastar {%
vul2009one}%
\begin{APACrefauthors}%
Vul, E.%
, Goodman, N\BPBI D.%
, Griffiths, T\BPBI L.%
\BCBL {}\ \BBA {} Tenenbaum, J\BPBI B.%
\end{APACrefauthors}%
\unskip\
\newblock
\APACrefYearMonthDay{2009}{}{}.
\newblock
{\BBOQ}\APACrefatitle {One and done? Optimal decisions from very few samples}
  {One and done? optimal decisions from very few samples}.{\BBCQ}
\newblock
\BIn{} \APACrefbtitle {{Proceedings of the 31\textsuperscript{st} Annual
  Meeting of the Cognitive Science Society}} {{Proceedings of the
  31\textsuperscript{st} Annual Meeting of the Cognitive Science Society}}\
  (\BVOL~1, \BPGS\ 66--72).
\newblock
\APACaddressPublisher{Austin, TX}{Cognitive Science Society}.
\PrintBackRefs{\CurrentBib}

\bibitem [\protect \citeauthoryear {%
Waldmann%
, Cheng%
, Hagmayer%
\BCBL {}\ \BBA {} Blaisdell%
}{%
Waldmann%
\ \protect \BOthers {.}}{%
{\protect \APACyear {2008}}%
}]{%
waldmann2008causal}
\APACinsertmetastar {%
waldmann2008causal}%
\begin{APACrefauthors}%
Waldmann, M\BPBI R.%
, Cheng, P.%
, Hagmayer, Y.%
\BCBL {}\ \BBA {} Blaisdell, A.%
\end{APACrefauthors}%
\unskip\
\newblock
\APACrefYearMonthDay{2008}{}{}.
\newblock
{\BBOQ}\APACrefatitle {Causal learning in rats and humans: A minimal rational
  model} {Causal learning in rats and humans: A minimal rational model}.{\BBCQ}
\newblock
\BIn{} M.~Oaksford\ \BBA {} N.~Chater\ (\BEDS), \APACrefbtitle {The
  Probabilistic Mind} {The probabilistic mind}\ (\BPGS\ 453--484).
\newblock
\APACaddressPublisher{Oxford}{Oxford University Press}.
\PrintBackRefs{\CurrentBib}

\bibitem [\protect \citeauthoryear {%
Weierstrass%
}{%
Weierstrass%
}{%
{\protect \APACyear {1902}}%
}]{%
weierstrass1902mathematische}
\APACinsertmetastar {%
weierstrass1902mathematische}%
\begin{APACrefauthors}%
Weierstrass, K.%
\end{APACrefauthors}%
\unskip\
\newblock
\APACrefYearMonthDay{1902}{}{}.
\newblock
{\BBOQ}\APACrefatitle {Mathematische Werke iv.} {Mathematische werke
  iv.}{\BBCQ}
\newblock
\APACjournalVolNumPages{Georg Olms}{4}{}{}.
\PrintBackRefs{\CurrentBib}

\bibitem [\protect \citeauthoryear {%
Williamson%
\ \BBA {} Gabbay%
}{%
Williamson%
\ \BBA {} Gabbay%
}{%
{\protect \APACyear {2005}}%
}]{%
williamson2005recursive}
\APACinsertmetastar {%
williamson2005recursive}%
\begin{APACrefauthors}%
Williamson, J.%
\BCBT {}\ \BBA {} Gabbay, D.%
\end{APACrefauthors}%
\unskip\
\newblock
\APACrefYearMonthDay{2005}{}{}.
\newblock
{\BBOQ}\APACrefatitle {Recursive causality in {Bayes}ian networks and
  self-fibring networks} {Recursive causality in {Bayes}ian networks and
  self-fibring networks}.{\BBCQ}
\newblock
\APACjournalVolNumPages{Laws and Models in the Sciences}{}{}{173--221}.
\PrintBackRefs{\CurrentBib}

\bibitem [\protect \citeauthoryear {%
Yeung%
\ \BBA {} Griffiths%
}{%
Yeung%
\ \BBA {} Griffiths%
}{%
{\protect \APACyear {2011}}%
}]{%
yeung2011estimating}
\APACinsertmetastar {%
yeung2011estimating}%
\begin{APACrefauthors}%
Yeung, S.%
\BCBT {}\ \BBA {} Griffiths, T\BPBI L.%
\end{APACrefauthors}%
\unskip\
\newblock
\APACrefYearMonthDay{2011}{}{}.
\newblock
{\BBOQ}\APACrefatitle {Estimating human priors on causal strength} {Estimating
  human priors on causal strength}.{\BBCQ}
\newblock
\BIn{} \APACrefbtitle {{Proceedings of the 33\textsuperscript{rd} Annual
  Meeting of the Cognitive Science Society}} {{Proceedings of the
  33\textsuperscript{rd} Annual Meeting of the Cognitive Science Society}}\
  (\BPGS\ 1709--1714).
\newblock
\APACaddressPublisher{Austin, TX}{Cognitive Science Society}.
\PrintBackRefs{\CurrentBib}

\bibitem [\protect \citeauthoryear {%
Yu%
\ \BBA {} Dayan%
}{%
Yu%
\ \BBA {} Dayan%
}{%
{\protect \APACyear {2003}}%
}]{%
yu2003expected}
\APACinsertmetastar {%
yu2003expected}%
\begin{APACrefauthors}%
Yu, A.%
\BCBT {}\ \BBA {} Dayan, P.%
\end{APACrefauthors}%
\unskip\
\newblock
\APACrefYearMonthDay{2003}{}{}.
\newblock
{\BBOQ}\APACrefatitle {Expected and unexpected uncertainty: {AC}h and {NE} in
  the neocortex} {Expected and unexpected uncertainty: {AC}h and {NE} in the
  neocortex}.{\BBCQ}
\newblock
\APACjournalVolNumPages{{Advances in Neural Information Processing
  Systems}}{}{}{173--180}.
\PrintBackRefs{\CurrentBib}

\end{thebibliography}

\clearpage

\begin{appendices}
\section{Formal specification of the models}\label{app:a}

\subsection*{Representation and inference}

A noisy-OR parametrized causal model $m$ over variables $X$, with
strength and background parameters $w_S$ and $w_B$ assign a likelihood
to each datum (a complete observation, or the outcome of an
intervention) ${\dd}$ as the product of the probability of each
variable that was not intervened upon given the states of its parents

\begin{align}
P({\dd}|m,\ww) &= \prod\nolimits_{x\in X}
P(x|{\dd}_{pa(x)}, \ww) \label{likelihood_appendix} \\
P(x|{\dd}_{pa(x)},\ww)&=x+(1-2x)(1-w_B)(1-w_S)^{\sum_{y\in
    pa(x)}y}
\label{noisyor_appendix} 
\end{align}
where $pa(x)$ denotes the parents of variable $x$ in the causal model
(see Figure~\ref{fig:cbn} for an example).  We can thus compute the
posterior probability of model $m\in M$ over a set of models
$\mathcal{M} $ given a prior $P(M)$ and
data $D=\{\dd^i\}$ associated with interventions
$C = \{\eee^i\}$. We can condition on $w_S$ and $w_B$ if known (e.g. in Experiment
1)

\begin{align}
 P(m|D, \ww)&=\frac{P(D|m,\ww;C)P(m)}{\sum_{m' \in M}P(D|m',\ww;C)P(m')}
\intertext{or else marginalize over their possible values (e.g. in Experiment 2)}
 P(m|D)&=\frac{\int_{\ww}P(D|m,\ww;C)p(\ww)P(m)\ \mathrm{d}\ww}{\sum_{m'
     \in M}\int_{\ww}P(D|m',\ww;C)p(\ww)P(m')\ 
   \mathrm{d}\ww}
\end{align}

\subsection*{Intervention choice}%

The value of an intervention can be quantified relative to a notion of uncertainty.  Here we adopt Shannon entropy \citep{shannon1951prediction}, for which the uncertainty in a distribution over causal models $M$ is given by

\begin{equation}
H(M) = -\sum_{m\in M}P(m)\log_2 P(m)
\end{equation}
Assuming $\ww$ is known, let $\Delta H(M|\dd,\ww;\eee)$ refer to the
reduction in uncertainty going from prior $P(M)$ to posterior
$P(M|\dd,\ww;\eee)$ after performing intervention $c$, then seeing data
$\dd$
\begin{equation}
\Delta H(M|d,\ww;\eee)=\left[-\sum_{m\in M}P(m)\log P(m)\right]-
\left[-\sum_{m\in M}P(m|{\dd},\ww;\eee)\log P(m|{\dd},\ww;\eee)\right] 
\label{eq:reduction_entropy}
\end{equation}
Given this objective, we can define the value of an intervention as the
expected reduction in uncertainty after seeing its outcome.  To get the
expectancy, we must average, prospectively, over the different possible
outcomes ${\dd} \in \mathcal{D}_{\eee}$ (where $\mathcal{D}_{\eee}$ is the
space of possible outcomes of intervention $\eee$) weighted by their
marginal likelihoods under the prior, giving 

\begin{equation}
\E_{\dd \in \mathcal{D}_{\eee}} \left[\Delta H(M|{\dd},\ww;\eee)\right]
=\sum_{\dd\in \mathcal{D}_{\eee}}\left(\Delta
H(M|{\dd},\ww;\eee)\sum_{m\in M}P({\dd}|m,\ww;\eee)P(m)\right) 
\label{eq:info_gain_app}
\end{equation}
For a greedily optimal sequence of interventions $\eee^1,\ldots,\eee^t$, we
take $P(M|D^{t-1},\ww;C^{t-1})$ as $P(M)$ and
$P(M|D^{t},\ww;C^{t-1},\eee^{t})$ as  
$P(M|{\dd},\ww;\eee)$ in Equation~\ref{eq:reduction_entropy}.  The most
valuable intervention at a given time point is then  

\begin{equation}
\eee^t=\arg\max_{\eee\in \mathcal{C}}\E_{{\dd} \in \mathcal{D}_{\eee}}
\left[\Delta H(M|{\dd}, D^{t-1},\ww;C^{t-1},\eee)\right] 
\end{equation}
If $\ww$ is unknown, we must use the marginal distribution, replacing Equation~\ref{eq:reduction_entropy} with
\begin{equation}
\Delta H(M|d;\eee)=\left[-\sum_{m\in M}P(m)\log P(m)\right]-
\left[-\sum_{m\in M}\int_{\ww}P(m|{\dd},\ww;\eee)p(\ww)\ \mathrm{d}\ww \log \int_{\ww}P(m|{\dd},\ww;\eee)p(\ww)\ \mathrm{d}\ww\right] 
\end{equation}

\subsection*{An algorithmic-level model of sequential belief change}%

Let $E$ be an adjacency matrix such that the upper triangle entries
where $E_{ij}$ (if $i<j\leq N$) denotes the state of edge $i-j$ in a
causal model $m$.  Any model $m\in M$ corresponds to a setting for all
$E_{ij}$ where $i<j\leq N$, to one of three edge states $e\in\{1:
i\rightarrow j,~0: i \nleftrightarrow j, ~-1: i \leftarrow j\}$. By
starting with any hypothesis and iteratively sampling from the
conditional distributions on edge states $P(E_{ij}|E_{\setminus ij},
\cald_r^t,\ww;\calc_r^t)$ \citep{goudie2011efficient} using the following equation: 

\begin{equation}
P(E_{ij}=e|E_{\setminus ij},\cald_r^t,\ww;\calc_r^t) = \frac{P(E_{ij}=e|E_{\setminus ij},\cald_r^t,\ww;\calc_r^t)}{\sum\limits{_{e'\in E_{ij}}} P(E_{ij}=e'|E_{\setminus ij},\cald_r^t,\ww;\calc_r^t)}
\label{eq:gibbssample}
\end{equation}
we can cheaply generate chains of dependent samples from
$P(M|\cald_r^t,\ww;\calc_r^t)$.  This can be done systematically (cycling
through all edges $\in i<j\leq N$), or randomly selecting the next edge
sample with $P(\frac{1}{|i, j|})$ where $|i, j|$ is the number of edges
in the graph. Here we assume random sampling for simplicity. Thus, on
each step, the selected $E_{ij}$ is updated using the newest values of
$E_{\setminus ij}$.\footnote{Edge changes that would create a cyclic
  graph always have a probability of zero}  Specifically, we assume
that after each new piece of evidence arrives:\\ 

\begin{enumerate}
\item The learner begins sampling with edges $E^{(0)}_{ij}$ for all $i$ and $j$ set as they were in their previous judgment $b^{t-1}$.
\item They then randomly select an edge $E_{ij}$ in $i<j\leq N$ to update.
\item They resample $E^{(1)}_{ij}$ using Equation~\ref{eq:gibbssample}.
\item If the search does not result in a new model they keep collecting evidence $\cald_r^t = \{\cald_r^t,\dd^t\}$, $\eee = \{\calc_r^t,\eee^t\}$.  If it does, the evidence is used up and forgotten, and they begin collecting evidence again (e.g. resetting $\cald_r^t = \{\}$ and $\calc_r^t = \{\}$).
\item The learner repeats steps 1 to 4 $k$ times, with their final edge choices $E^{(k)}$ constituting their new belief $b^t$.
 \end{enumerate}

We assume for simplicity that $b^0$, before any data has been seen is an unconnected graph, but have tested this assumption by fitting the data from t=2 onward only finding better fits overall and a stronger win for \emph{Neurath's ship} over the other models we consider.

\subsubsection*{Resampling, hill climbing or random change}

We also consider generalizations of Equation~\ref{eq:gibbssample}
allowing transitions to be governed by higher powers of $P(E_{ij}=e|
E_{\setminus ij},\cald_r^t,\ww;\calc_r^t)$ 

\begin{equation}
P^\omega(E_{ij}=e|E_{\setminus ij},\cald_r^t,\ww;\calc_r^t) =   \frac{
  P^{\omega}(E_{ij}=e| E_{\setminus ij},\cald_r^t,\ww;\calc_r^t)} {\sum_{e'\in
    E_{ij}} P^{\omega}(E_{ij}=e'| E_{\setminus ij},\cald_r^t,\ww;\calc_r^t)  }  
\end{equation}

 yielding stronger preference for the most likely state of $e_{ij}$ if $\omega>1$ and more random sampling if $\omega<1$.

\subsubsection*{A distribution over search lengths}

We assume that for each update, the learner's length of search $k$ is drawn from a Poisson distribution with average $\lambda\in[0,\infty]$

\begin{equation}
P(k)= \frac{\lambda^k e^{-\lambda}}{k!}
\end{equation}

\subsubsection*{Putting these together}

To calculate the probability distribution of new belief $b^t$ given ${\dd}^{t}$, $b^{t-1}$ search behavior $\omega$ and a chain of length $k$, we first construct the transition matrix $R^\omega_t$ for the Markov search chain by averaging over the conditional distributions associated with the choice of each edge,  weighted by the probability of selecting that edge

\begin{equation}
R^\omega_t = \sum_{i<j \leq N} P^\omega(E_{ij}=e|E_{\setminus
  ij},\cald_r^t,\ww;\calc_r^t)\times\frac{1}{|i, j|} 
\label{eq:create_R}
\end{equation}
for each possible belief $b$.

By raising this transition matrix to the power $k$ (i.e. some search
length) and selecting the row corresponding to starting belief
$[(R^\omega_t)^k]_{b^{t-1}}$, we get the probability of adopting each
$m\in M$ as new belief $b^t$ (see Figure~\ref{fig:neurath_gibbs_example}
for a visualization) at the end of the $k$ length search 
 
\begin{equation}
  P(B^t|\cald_r^t, b^{t-1}, \omega, k;\calc_r^t)=[(R^\omega_t)^k]_{b^{t-1}m}
\end{equation}
Finally, by averaging over different possible chain lengths $k$,
weighted by their probability Poisson($\lambda$) we get the marginal
probability that a learner will move to each possible new belief in $B$
at $t$

\begin{equation}
P(B^t|\cald_r^t, b^{t-1}, \omega, \lambda;\calc_r^t)={\sum\limits_{0}^\infty}
\frac{\lambda^k e^{-\lambda}}{k!} [(R^\omega_t)^k]_{b^{t-1}m} 
\label{eq:generate_new_belief}
\end{equation}

\subsection*{A local uncertainty schema}\label{app:a_local_entropies}
\subsubsection*{Edge focus}

Relative to a focus on an edge $E_{xy}$, intervention values were calculated using expected information as in Appendix~\ref{app:a}, but assuming prior entropy as that of a uniform distribution over the three possible edge states
\begin{equation}
  H(E_{xy}|E_{\setminus xy}) = -3 \left(\frac{1}{3}\log_2\frac{1}{3}\right)
\end{equation}
and calculating posterior entropies for the possible outcomes $\dd\in \mathcal{D}$ using
\begin{equation}
H(E_{xy}|E_{\setminus xy},{\dd},\ww;\eee) = -\sum_{z\in\{-1,0,1\}}P(E_{xy}=z|E_{\setminus xy},{\dd}, \ww;\eee)\log_2 P(E_{xy}=z|E_{\setminus xy},{\dd}, \ww;\eee)
\end{equation}

\subsubsection*{Effect focus entropy}

Relative to a focus on the effects of variable $x$, intervention values were calculated using expected information as in Appendix~\ref{app:a} but using prior entropy, calculated by partitioning a uniform prior over models $M$ into sets of models $\mathrm{Mo}(z)$ corresponding to each descendant set $z\subseteq \mathrm{De}(x)$

\begin{equation}
  H(\mathrm{De}(x)) = -\sum_{z\subseteq\mathrm{De}(x)}\left(\sum_{m\in \mathrm{Mo}(z)} \frac{1}{|M|}\right)\log_2\left(\sum_{m\in \mathrm{Mo}(z)} \frac{1}{|M|}\right)
\end{equation}
Posterior entropies were then calculated by summing over probabilities of the the elements in each $\mathrm{Mo}(z)$ for each $z\subseteq \mathrm{De}(x)$
 \begin{equation}
    H(\mathrm{De}(x)|{\dd},\ww;\eee) = -\sum_{z\subseteq\mathrm{De}(x)}\left(\sum_{m\in \mathrm{Mo}(z)} P(m|{\dd},\ww;\eee)\right)\log_2\left(\sum_{m\in \mathrm{Mo}(z)} P(m|{\dd},\ww;\eee)\right)
 \end{equation}

\subsubsection*{Confirmation focus entropy}

Relative to a focus on distinguishing current hypothesis $b^t$ from null hypothesis $b^0$, intervention values were calculated using expected information as above but prior entropy was always based on a uniform prior over the two hypotheses
\begin{equation}
  H(\{b^t, b^0\}) = -2 \left(\frac{1}{2}\log_2\frac{1}{2}\right)
\end{equation}
and posterior entropies were calculated using
\begin{equation}
  H(\{b^t, b^0\}|{\dd},\ww;\eee) = - \sum_{z \in \{0,t\}} \frac{P(b^z|{\dd},\ww;\eee)}{\sum_{z' \in \{0,t\}} P(b^{z'}|{\dd},\ww;\eee)} \log_2 \frac{P(b^z|{\dd},\ww;\eee)}{\sum_{z' \in \{0,t\}} P(b^{z'}|{\dd},\ww;\eee)}
\end{equation}

\section{Additional modeling details}\label{app:b}

All models were fit using maximum likelihood.  Maximum likelihood estimates were found using Brent (for one parameter) or Nelder-Mead (for several parameters) optimization, as implemented by R's \texttt{optim} function. Convergence to global optima was checked by repeating all optimizations with a range of randomly selected starting parameters.

\paragraph{$k$} 
For averaging across different values of $k$ in the belief models, we capped $k$ at 50 and renormalized the distribution such that $P(k \geq 0 \wedge k\leq 50)=1$.  This made negligible difference to the fits since the probabilities of $P(B^t|\mathbf{d}^r, b^{t-1}, \omega, k;\calc_r^t)$ for values of $k\gg N$ (where $N$ is the number of variables) were very similar.

\paragraph{$\epsilon$} To allow that participants are liable to occasionally lapse concentration or forget the outcome of a test,  we included a lapse parameter $\epsilon$ -- i.e., a parametric amount of decision noise $\epsilon\in[0,1]$ -- so that the probability of a belief would be a mixture of that predicted by the model and uniform noise.  This ensured that occasional random judgments did not have undue effects on the other parameters of each model.

\paragraph{$b^0$} 
We assume for simplicity that people's starting belief, $b^0$, before any data has been seen, is an unconnected graph.  %

\subsection*{Marginalization}

For all modeling in Experiment~3, we had to average over the unknown noise $\ww$.  To do this, we drew 1000 paired uniformly distributed $w_S$ and $w_B$ samples and averaged over these when computing marginal likelihoods and posteriors.  These marginal priors and posteriors were used for computing expected information gain values.

\subsection*{Evaluating fits}

\emph{Baseline} acts as the null model for computing BIC's \citep{schwarz1978estimating} and pseudo-$R^2$'s \citep{dobson2010introduction} for all other models.

\end{appendices}

\end{document}